\documentclass[10pt]{article} % For LaTeX2e
\usepackage[accepted]{tmlr}
\usepackage{amsmath,amsfonts,bm}

\def\eqref#1{equation~\ref{#1}}
\def\1{\bm{1}}

\def\ra{{\textnormal{a}}}

\def\rx{{\textnormal{x}}}

\def\rva{{\mathbf{a}}}

\def\erva{{\textnormal{a}}}

\def\ervx{{\textnormal{x}}}

\def\rmA{{\mathbf{A}}}

\def\vmu{{\bm{\mu}}}
\def\vtheta{{\bm{\theta}}}
\def\va{{\bm{a}}}

\def\ve{{\bm{e}}}

\def\vx{{\bm{x}}}

\def\eva{{a}}

\def\mA{{\bm{A}}}

\def\mH{{\bm{H}}}
\def\mI{{\bm{I}}}
\def\mJ{{\bm{J}}}

\def\mX{{\bm{X}}}

\def\mSigma{{\bm{\Sigma}}}

\DeclareMathAlphabet{\mathsfit}{\encodingdefault}{\sfdefault}{m}{sl}
\SetMathAlphabet{\mathsfit}{bold}{\encodingdefault}{\sfdefault}{bx}{n}
\newcommand{\tens}[1]{\bm{\mathsfit{#1}}}
\def\tA{{\tens{A}}}

\def\tX{{\tens{X}}}

\def\gG{{\mathcal{G}}}

\def\sA{{\mathbb{A}}}
\def\sB{{\mathbb{B}}}

\def\sS{{\mathbb{S}}}

\def\emA{{A}}

\newcommand{\etens}[1]{\mathsfit{#1}}

\def\etA{{\etens{A}}}

\newcommand{\E}{\mathbb{E}}

\newcommand{\R}{\mathbb{R}}

\newcommand{\KL}{D_{\mathrm{KL}}}
\newcommand{\Var}{\mathrm{Var}}

\newcommand{\Cov}{\mathrm{Cov}}
\newcommand{\normltwo}{L^2}
\newcommand{\normlp}{L^p}

\newcommand{\parents}{Pa} % See usage in notation.tex. Chosen to match Daphne's book.

\usepackage{url}

\usepackage{amsthm}
\usepackage{subcaption}
\usepackage{colortbl}

\newtheorem{corollary}{Corollary}%[theorem]

\newtheorem{definition}{Definition}
\newtheorem{proposition}{Proposition}

\usepackage{xspace}
\usepackage{pifont}
\newcommand{\cmark}{\textcolor{teal}{\ding{51}}\xspace}%
\newcommand{\xmark}{\textcolor{purple}{\ding{55}}\xspace}%
\newcommand{\ft}{\textsf{\small FT}\xspace}
\newcommand{\daft}{\textsf{\small DAFT}\xspace}
\newcommand{\dafte}{\textsf{\small DAFT\!-E}\xspace}
\newcommand{\daftsq}{\textsf{\small DA(FT)}${}^2$\xspace}
\newcommand{\daftz}{\daft~\!${}^\text{\!\scriptsize Z}$\xspace}
\newcommand{\daftez}{\dafte~\!${}^\text{\!\scriptsize Z}$\xspace}

\usepackage{comment} 

\usepackage{adjustbox}
\usepackage{multirow}

\usepackage{enumitem}
\setlist[itemize]{noitemsep, topsep=0pt, leftmargin=*}
\setlist[enumerate]{noitemsep, topsep=0pt, leftmargin=*}

\usepackage{amsmath}
\usepackage{booktabs}
\usepackage{amssymb}

\usepackage{wrapfig}

\usepackage{mdframed}
\usepackage{adjustbox}
\newenvironment{conc}{\begin{mdframed}[linecolor=black!0,backgroundcolor=orange!10]\noindent%
\ignorespaces
}{\end{mdframed}}

\newcommand{\iia}[1]{\textcolor{black} {#1}}
\usepackage{cancel}
\usepackage[normalem]{ulem}

\title{On the Utility of Existing Fine-Tuned Models on Data-Scarce Domains}

\author{\name Md Ibrahim Ibne Alam \email alamm4@rpi.edu \\
      \addr Department of Electrical, Computer, and Systems Engineering\\
      Rensselaer Polytechnic Institute
      \AND
      \name Parikshit Ram \email parikshit.Ram@ibm.com \\
      \addr IBM
      \AND
      \name Soham Dan \email sdan021@gmail.com\\
      \addr Microsoft
      \AND
      \name Horst Samulowitz \email samulowitz@us.ibm.com \\
      \addr IBM
      \AND
      \name Koushik Kar \email koushik@ecse.rpi.edu \\
      \addr Department of Electrical, Computer, and Systems Engineering\\
      Rensselaer Polytechnic Institute}

\def\month{05}  % Insert correct month for camera-ready version
\def\year{2025} % Insert correct year for camera-ready version
\def\openreview{\url{https://openreview.net/forum?id=kY2fKLOGkI}} % Insert correct link to OpenReview for camera-ready version

\begin{document}

\maketitle

\begin{abstract}
Large Language Models (LLMs) have been observed to perform well on a wide range of downstream tasks when fine-tuned on domain-specific data. However, such data may not be readily available in many applications, motivating zero-shot or few-shot approaches using \iia{existing \emph{domain or task adjacent (fine-tuned) models}, which we call \daft}. While several fine-tuned models for various tasks are available, finding \iia{one} appropriate \daft
%domain-adjacent 
model for a given task is often not straight forward. In this paper, we \iia{explore different utilization techniques of these existing \daft models for data-scarce problems, i.e., tasks for which data is not available or limited. We observe}
that for zero-shot problems, ensembling of \daft models provides an accuracy performance close to that of the single best model. With few-shot problems (few data from target domain available), this performance can be improved further \iia{by picking or putting more weights to the \daft models that are expected to perform better on the target task.}
\end{abstract}

\section{Introduction}
\label{sec:intro}

% Options: (0) Fine-tune with data
% (1) Pick one of the available ones and use them zero-shot -- no (training) compute cost no data
% (2) Pick one of the available ones and fine-tune with some data -- medium (training) compute cost (need FM backprop) and  low data
% (3) Instead of one, pick all and ensemble and use them zero-shot -- no (training) cost no data
% (4) few-shot learning of ensemble weights

% \todo{after the first line, best to use just FM instead of both LLM and FM. We have FM in our title so best to just go with it.}

Pre-trained Large Language Models (LLMs) are used for different downstream tasks. Usually, for LLMs that are not directly suitable for downstream tasks, a task-specific header layer, \iia{e.g., for classification,} is needed at the output. \textit{We define the \emph{base model} as the \iia{pre-trained} LLM with a task-specific \iia{(un-tuned)} header layer.} As an example, BERT models \citep{kenton2019bert} were trained for sentence completion tasks. In order to perform sentiment classification, we need to add a header layer with output dimension matching the number of classes. Usually, \emph{base models} do not need much training to perform well, because the pre-trained LLMs are already trained on a large corpus of data.

To perform a task with base models, we can fine-tune a base model with task-specific training data, and then use the fine-tuned model. This fine-tuning of model is computationally much less extensive compared to the initial training phase of the LLMs. However, it is often the case that appropriate training data for fine-tuning is not readily 
or abundantly available, \iia{i.e., the domain is data-scarce}. Moreover, even if one does have the data to fine-tune, one may not have the computational resources, or the time required, to fine-tune.
%these base models. 
It is worth noting that fine-tuning can be computation, memory and time intensive, even if very few iterations are needed to attain good performance. Hence, we \iia{compare different fine-tuning solutions with some} alternatives that bypass the challenges of fine-tuning and leverage the large number of fine-tuned LLMs already available (in curated repositories like Hugging-face \citep{huggingface_models, open_source_FMs}).
It is important to acknowledge the scale of fine-tuned models being created; for instance, {\em just three days after LLaMA-3 was released, more than 1,000 fine-tuned LLMs of LLaMA-3} were publicly available on Hugging-face~\citep{huggingface_llama3}.

\begin{conc}
\iia{Research Question: Can we use these plethora of fine-tuned LLMs for a data-scarce task?}
\end{conc}

\begin{wrapfigure}{r}{0.55\textwidth}
\centering
\vspace{-10pt}
\includegraphics[width=0.95\linewidth]{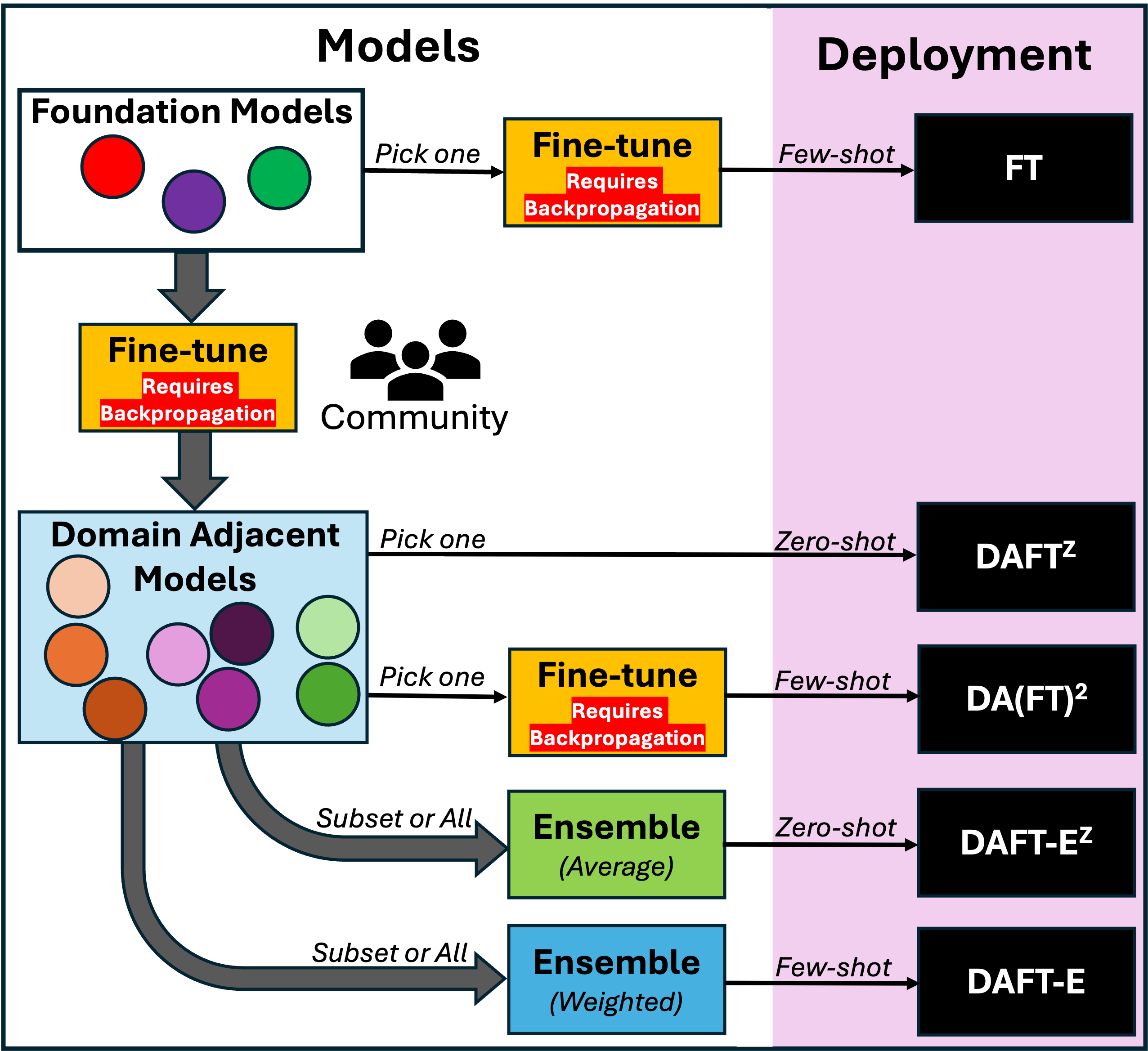}
\caption{{\em Different options for few-shot tasks given LLMs and community created \daft models}: (1)~Directly fine-tune the LLM (\ft); (2)~Select one \daft model and use it zero-shot (\daftz); (3)~Use a \daft model but with task-specific fine-tuning (\daftsq); (4) Ensemble multiple ~\daft models and use it zero-shot (\daftez); (5) Use ensemble of \daft models with extremely lightweight few-shot adaptation of the ensemble weights (\dafte).} 
\label{fig:paper_structure}
\vspace{-10pt}
\end{wrapfigure}

Here, we \iia{investigate} the potential of existing and publicly available fine-tuned models (e.g., including those fine-tuned via  LoRA~\citep{sheng2023slora} and PEFT~\citep{li2021prefixtuning})
% \todo{I think we should say ``including those fine-tuned via LORA and PEFT''} 
by leveraging them to perform different tasks. To use a fine-tuned model on a task, e.g., sentiment classification, we need that model to be \iia{\textit{domain or task adjacent}} (discussed in Section \ref{Sec:DAM}).
% \todo[color=blue!10]{``formally defined and discussed'' $\to$ ``discussed''. Maybe add ``(discussed in Sec 3; at a high level, they have output spaces we can map to our data-scarce task)''.}
%Informally, a fine-tuned LLM is domain-adjacent if (i)~it has been fine-tuned on a similar general task, such as sentiment analysis, and (ii)~its output space can be appropriately mapped to the current (few-shot) task.
% dataset that the model was fine-tuned with (say $D_A$) needs to be similar to the test data (say $D_T$) that we want to use the model on.
% \todo{Instead of test data connection, we at least need (i) the fine-tuning to be done on a similar general task, such as sentiment analysis, and (ii) we are able to appropriately map the output space of the \daft model to our current task.}
We denote these {\bf D}omain {\bf A}djacent {\bf F}ine-{\bf T}uned models as \daft. 
% In the generation process of \daft, 
% \todo[color=blue!10]{$\to$``Intuitively, the closer''}
% \iia{`Intuitively, the closer the fine-tuning dataset of \daft is to the target domain data}, the better will be the performance of \daft. 
One main property of \daft is that it can be used without any further \iia{fine-tuning} 
%training 
(no extra computation) in a zero-shot manner (no data adaptation). Also, inference with \daft does not require large computational resources when compared to performing fine-tuning. 

The availability of \daft models combined with their ease of use, make them very appealing to use for different tasks. \iia{However, the performance of \daft models on the target domain data (denoted as $D_T$) can be poor depending on the fine-tuning dataset of the respective \daft models. In this work, we {\em systematically study four different ways (Fig. \ref{fig:paper_structure}) of leveraging the \daft models}, and demonstrate how it is possible to get strong performance with low compute adaptation, demonstrating the utility of these community generated \daft models. These four options along with fine-tuning the base model using data from target domain (\ft in Fig. \ref{fig:paper_structure}) are defined as follows:}

% . Then to analyze the efficiency of using these \daft models 
%in comparison to other existing options, 
% we first define several comparable 
% alternatives
% \iia{options as follows :}. 
% \todo[color=blue!10]{we not only define, but we also systematically study their performance. we need to highlight that as our contribution}

% $D_A$ as one of the domain adjacent datasets of $D_T$, where 
% $D_T$ as the testing dataset on which we need to perform a given task. 
% Then we have the following options (Figure \ref{fig:paper_structure}):

\begin{enumerate}
\item 
\textbf{\ft} -- Fine-tune \emph{base models} with training data obtained from $D_T$. 
\item
\textbf{\daftz} -- Pick one available \daft model and use it for zero-shot inference (no training cost).
\item
\textbf{\daftsq} -- Pick one available \daft model and fine-tune with training data from $D_T$.
\item
\textbf{\daftez} -- Use a subset or all \daft models and ensemble them for zero-shot inference.
\item
\textbf{\dafte} -- Use a subset or all \daft models for few-shot learning of ensemble weights using training data from $D_T$ and perform inference.
\end{enumerate}

Section \ref{sec:related-work} discusses related work, followed by Section \ref{Sec:DAM} \iia{discussing} \daft models and \iia{their potential for} various NLP tasks.
Section \ref{sec:empirical_eval} discusses zero-shot and few-shot performance, while Section \ref{sec:theory} \iia{theoretically characterizes conditions that result in strong performance from the \daft ensemble}.

\section{Related Work}
\label{sec:related-work}
%
% \todo[ color=blue!10]{we might need a bit intro here since jumping into ensembling reads a little odd. Consider:
% ``''.
% Then maybe we can bring up transfer learning first, then the rest. We can also slightly shrink the detailed explanations here if needed and move detailed explanations to the appendix.
% }

The two main aspects of our work are: (i)~trying to adapt to data-scarce tasks or transfer learning, (ii)~leveraging multiple pre-fine-tuned models via ensembling. Here we will briefly discuss related work on transfer learning, ensembling and the closely related blending and mixture-of-experts.

Transfer learning~\citep{pan2009survey,zhuang2020comprehensive} is closely related to the goals of our work, as we wish to adapt a model trained on a data-rich source distribution to a data-scarce target problem. The success on the target task often relies heavily on the level of \textit{positive transfer} one can achieve, and various ways of training the model on the source and target data have been developed to maximize the positive transfer \iia{(e.g., by extending the already expensive pre-training phase \citep{gururangan-etal-2020-dont})}.
A thorough empirical study on the factors for good transfer learning across diverse domains and tasks has been done in \citet{mensink2021factors}. In \citet{zamir2018taskonomy}, a parameterized readout function represented by a shallow network was introduced to check (after necessary training) the affinity between any two tasks. Given some budget constraints, using the affinity matrix, the performance on $T$ tasks can be optimized by only training models on $S$ tasks, where $T \gg S$.
% \todo[color=blue!10]{We have this ref since it came up, but not sure it fits well here. What is ``domain adjacent data''? and I think the paper refers to pretraining, whereas domain-adjacent data for us comes in at FT.
% Maybe instead say ``(for example, by extending the already expensive pretraining phase (citation))''}
% In the context of LLMs, fine-tuning a (base) LLM with task-specific data is an instance of transfer learning.
\iia{Here}, we consider the \daft models as the source models for knowledge transfer. However, not every \daft model may transfer positively, since the \daft models are obtained from public non-curated pools. \iia{Also, for few shot problems, fine-tuning \daft with target domain data (denoted as \daftsq) is a \emph{transfer learning} option that is considered here (\daftsq in Fig. \ref{fig:paper_structure}).}%; we adapt to the target problem without modifying the source models, taking advantage of the numerous openly available source models.

\emph{Ensembling} is a technique that combines the predictions of multiple classifiers to generate a single decision and has been extensively investigated \iia{over the past few decades}~\citep{breiman1996bagging, clemen1989combining, maclin1997empirical, dietterich2000experimental}. The pre-requisites for effective ensembling are: (i) models with decent performance, and (ii) models that make independent errors, i.e., training diversity in models \citep{ovadia2019can, lakshminarayanan2017simple}. The ensemble of models can be the weighted or unweighted average of the model outputs. For weighted averaging, some training is needed to learn the weights~\citep{caruana2004ensemble}. As another approach, \citet{dvornik2020selecting} proposed a method to select a weighted concatenation of the output features of models corresponding to multiple domains (models fine-tuned with different datasets) in a few shot environment. The optimized weights of the features are calculated by minimizing the negative log-likelihood classification loss on the few-shot target data. Isotropic merging~\citep{polyak1992acceleration} is a technique similar to ensembling that has been studied alongside ensemble methods over the last few decades and has shown promising performance with LLMs, e.g., Model Soup and its variants \citep{wortsman2022model}. Recently, a variation of Isotropic merging has been introduced as AdaMerging~\citep{yang2023adamerging}, where multiple LLMs, individually expert on different tasks, are merged to generate a single model that can perform well on all tasks. A forward pass on some unlabeled data is needed to find the merging coefficients using entropy minimization.
% \todo[color=blue!10]{maybe mention ``... promising performance with LLMs''}
%Recent works have established the potential of using an ensemble to achieve better performance compared to any single model \cite{gontijo2021no, abburi2023generative, ovadia2019can}. 
%Isotropic merging \cite{polyak1992acceleration} is another technique similar to ensembling, 
%and has been studied alongside ensemble methods over the last few decades . The main idea of isotropic merging is that 
%where the parameters of different models of the same structure are averaged to get a single model and that model usually performs well on the target tasks \cite{izmailov2018averaging}. In recent years, isotropic merging on LLM has shown some potential
%a lot of focus and some of these work has shown potential results 
%\cite{wortsman2022model, matena2022merging}. 
\iia{One of the main contributions} of our work is \iia{to focus on understanding the efficacy of ensembling in the context of \daft models for data-scarce tasks}.
% the use of ensembling in the context of \daft models fine-tuned from pretrained LLMs, to maximize the efficacy of using these models.
% \todo[color=blue!10]{how about ``We focus on understanding the efficacy of ensembling in the context of daft models for data-scarce tasks.''}
%While the notion of domain adjacent models and use of ensembling in that context is novel, we briefly survey prior work on broad topics that are related to our work. 

The main differences between Model Soup~\citep{wortsman2022model}, AdaMerging~\citep{yang2023adamerging}, and the proposed ensemble methods (\daftez and \dafte), are highlighted in Table \ref{tab:methods_comparison}. 
In the table the following abbreviations are used;  SLI: Single LLM inference, NMWA: No (LLM model) weight access, MAE: multi-architecture ensemble, SFPA: single forward pass (for few-shot) adaptation.
The table highlights that \daftez and \dafte provide lightweight LLM backprop-free model combination schemes (similar to Uniform Soup, Greedy Soup and AdaMerging), without requiring access to the model weights, and having the ability to combine LLMs with different architectures. Furthermore, the few-shot adaptation in \dafte requires only a single forward-pass with the few-shot samples compared to Greedy Soup, which requires multiple forward-passes. However, both Uniform and Greedy Soup finally need to perform inference with a single LLM while \daftez and \dafte have to perform inference with multiple LLMs.

Another approach related to ensembling is \textit{blending} of models. 
For instance, LLM-Blender~\citep{jiang2023llmblender} performs ensemble of $n$ LLMs by pairwise ranking and generative fusion. 
%\todo{from here ...}
The pairwise ranking approach requires creating a custom dataset, training a BERT model, and it incurs substantial computational overhead to perform inference. The proposed generative fusion combines the ranked list from the pairwise ranking with a fine-tuned LLM 
% for this task 
to generate a response.
%For the former, a data set is created to train BERT model~\cite{deberta} on each input query and a randomly selected subset of all possible pairs of responses from the $n$ models. 
% The learning uses ranks based on standard natural language generation metrics as well as ChatGPT-based scores. 
%During inference on a given input, this trained model uses $n(n-1)$ iterations to return a rank for all pairs of responses from all models in the ensemble. Then the best response candidate is identified by using an aggregation function requiring another $n^2$ iterations.
%For generative fusion the top-k responses are combined with the input and using an LLM fine-tuned for this task, a new response is generated. 
%\todo{... to here. Can we summarize here and move detailed explanation to an appendix section on related work}
In contrast, our approaches with \daft focus on a completely task agnostic (e.g., not just generative tasks) and computationally low-cost solution to leverage multiple fine-tuned LLMs (Please check Table \ref{tab:methods_comparison} for comparison). %For instance, our approach does not require training a pairwise ranker and creating the appropriate data for it nor finetuning an LLM to fuse multiple outputs for any given task. 
Another type of blending is introduced in \citet{lu2024blending}, and is orthogonal to \iia{ensemble}. The method selects base models at random and combines them by adding the response of an already evaluated model to the input of the subsequent model; a sequential approach that can be readily combined with \iia{ensembling}.

\begin{wraptable}{r}{0.52\textwidth}
\footnotesize
\centering
\vspace{-15pt}
\caption{Comparison of different methods 
to combine pretrained models 
in terms of the features.
% requirements on the available (\daft) models that are being combined and the kind of computation necessary. 
{\bf BPF}: Method is LLM backpropagation free. {\bf SLI}: Method needs to perform inference on the target task with a single LLM. {\bf NMWA}: No LLM model weight access is required by the method. {\bf MAE}: Method can handle multi-architecture ensemble. {\bf SFPA}: Method requires a single forward pass for few-shot adaptation. \\
}
\label{tab:methods_comparison}
\vspace{-15pt}
{\footnotesize\begin{tabular}{lcccccc}
%{p{2 cm}| p{1.5cm}|p{2cm}|p{1.5cm}|p{2cm}| p{2.5cm}| p{2.5cm}}
\toprule
Method & 0-shot & BPF & SLI & NMWA & MAE & SFPA \\
\midrule
\daft & \cmark & \cmark & \cmark & \cmark & N/A & N/A \\
% DAFT	& Yes	 & No	& Yes	& No	& N/A & N/A\\
Uniform Soup & 	\cmark & \cmark & \cmark & \xmark & \xmark & N/A \\
AdaMerging & 	\cmark & \cmark & \cmark & \xmark & \xmark & N/A \\
\daftez & \cmark & \cmark & \xmark & \cmark & \cmark & N/A \\
%DAFT-Ez	& Yes	& No	& No	& No	& Yes & N/A\\
\midrule
\ft & \xmark & \xmark & \cmark & \xmark & N/A & \xmark \\
% FT	& No	& Yes	& Yes	& Yes	& N/A & N/A\\
\daftsq & \xmark & \xmark & \cmark & \xmark & N/A & \xmark \\
% DAFT2 & 	No	& Yes	& Yes	& Yes	& N/A & N/A\\
Greedy Soup & \xmark & \cmark & \cmark & \xmark & \xmark & \xmark \\
LLM-Blender & 	\xmark & \xmark & \xmark & \cmark & \cmark & \xmark \\
\dafte & \xmark & \cmark & \xmark & \cmark & \cmark & \cmark \\
%DAFT-E	& No	& No &	No & 	No & 	Yes & No\\
\bottomrule
\end{tabular}}
\vspace{-20pt}
\end{wraptable}

\emph{Mixture of Experts} (MoE)~\citep{jacobs1991adaptive} 
%the idea evolved over time and was used in different 
has been utilized by different ML models for both regression and classification tasks~\citep{yuksel2012twenty}.
%i.e., support vector machines (SVMs), Gaussian processes (GPs), and hidden Markov models (HMMs) 
Before recent developments in LLMs, MoE was mainly focused on dense mixture of experts, which has been replaced by a sparse mixture of experts~\citep{fedus2022review}.
% \todo[color=blue!10]{Is ``dense expert models'' a standard term or are we referring to a ``dense mixture of experts'' as opposed to sparse mixture used with LLMs}
% However, that idea has been transformed into
% sparse  mixture of expert models with the introduction of MoE in LLMs. 
% In current practice, MoE is a sparse expert model, where a set of parameters of a neural network are partitioned into \textit{experts} having their own weights \cite{fedus2022review}. 
%In recent times, most of the work focused on MoE in the field of language modeling is inspired by the work of \cite{shazeer2017outrageously}, and is mainly an extension of their idea. 
%Building upon the work in \cite{shazeer2017outrageously}, more recent works such as \cite{lepikhin2020gshard, fedus2022switch, zhou2022mixture}, focus mostly on scaling the model using different techniques, i.e.,  sparse algorithm, routing, or gating mechanism. 
%In their work, they mainly replaced the feed-forward layers of a Transformer with expert layers. 
%The work in \cite{gururangan2021demix, li2022branch} handles MoE in a different way, where the experts are independent models, which are trained on specific domains. All these MoE approaches need the training of the experts along with the training of the routing or gating mechanism and is usually computationally expensive. 
One recent work on MoE, Mixtral~\citep{jiang2024mixtral} uses a sparse LLM MoE where a routing network (e.g., \citet{rosenbaum2017routing}) is pre-trained to map input tokens to a subset of experts. 
%\todo{from here ...}
%In this particular instance, there are 8 experts, and single tokens will be routed to exactly 2 experts and their output is weighted and summed to obtain a single output. 
%\todo{... can remove from here and move to appendix if we want to discuss it}
Compared to all these MoE models 
\iia{(which requires multiple back-propagation through all the involved LLMs to train the routing network), the ensemble of \daft models that we consider require at most a single forward pass through the LLMs for few-shot target task adaptation, and hence are computationally significantly more lightweight.}

\vspace{-2pt}
%\section{Domain Adjacent Fine-Tuned (\daft) Models}
\section{\textsf{DAFT} Models}
\label{Sec:DAM}
% - Sec 2: What is a Domain Adjacent Fine-tuned Model and how we can use do?
% -- Define notion of domain adjacency & corresponding fine-tuned model
% -- Where are they found?
% -- How can we use them?
\vspace{-1pt}
With the rapid expansion of open platforms like Colab and Kaggle to perform model training, and having access to different public datasets, researchers can access LLMs and datasets that can be used to fine-tune these LLMs. This has lead to a large repository of publicly available \emph{fine-tuned base models} (as defined in Section~\ref{sec:intro})
%becoming available for public use 
\citep{huggingface_models, open_source_FMs}. 
We formally define \textit{base model} as;
\begin{definition}
A \textbf{base model} has a pre-trained LLM and a task-specific (un-tuned) header layer added on top of the LLM.
\end{definition}
%Inspired by the abundance of these base models fine-tuned for different tasks, we analyze the performance of these fine-tuned base models and propose different methods to use them to attain better performance. 
In Section~\ref{sec:intro} we argued that the performance of any such fine-tuned base model on some target dataset depends on the source dataset on which it was fine-tuned on. 
A base model’s header layer can vary depending on the task, and it can become a domain adjacent
fine-tuned model (DAFT) for a task when the base model is fine-tuned (either the header layer or the full
model) with a dataset that has the same task criterion as the target data. So, we define the DAFT model
as follows:
\begin{definition}
A domain adjacent fine-tuned model (DAFT) is a fine-tuned base model, if the model can
be used for the same (general) task as the target task.
\end{definition}

The task similarity between two tasks can be calculated in two ways: (i) by the source and target dataset similarity and (ii) by task description similarity. Since in most cases the source dataset, with which the DAFT model was fine-tuned, is not available, method (ii) is more general. For method (ii), the task similarity can be identified manually or measured using an LLM fine-tuned for `textual similarity’. In our definition of DAFT, we are proposing to follow method (ii) to find DAFT models, i.e., when a model is fine-tuned with a dataset that has the same task criterion as the target data.

\vspace{-2pt}
\subsection{Formation of \daft models}
% \todo{With a section name like ``DAFT framework'', the first subsection of ``Datasets'' seems awkward. We should think about a different name for this section like ``A case for DAFT models''}
\vspace{-2pt}
\subsubsection{Tasks and Datasets}
\label{subsec:dataset}
% Authentic datasets that can be used for specific tasks (i.e., sentiment analysis) are hard to find. 
We consider two broad NLP tasks, (1) Sentiment Analysis (positive or negative sentiment) and 2) Textual Similarity (if two sentences are similar). \iia{Although there are (currently) $1667$ and $936$ \daft models available for sentiment analysis and textual similarity task in Hugging Face platform, to mitigate the risk of benchmark data leakage, we chose to create our own \daft models by fine-tuning the base models with specific datasets}\footnote{The performance of \daft models downloaded directly from Huggingface are analyzed further in Appendix \ref{sec:from_internet}.}~\citep{huggingface_data, cornell_movie_data, kaggle_tweet_data}.
% \todo[color=blue!10]{Instead say ``to mitigate the risk of benchmark data leakage''}
% with datasets from different sources , f
The datasets for Sentiment Analysis are: Amazon polarity, Cornell Movie, IMDB, SST2, Tweet sentiment, and Yelp Polarity; and for Textual Similarity: MRPC, QQP, STS-B (details in Appendix \ref{appndx_subsec_dataset}).
\vspace{-2pt}
\subsubsection{Models}
\label{subsec:models}
For both tasks, we chose three LLMs (with fine-tuning) for different performance analyses: (1) Roberta-base, (2) BERT-based-uncased, and (3) xlnet-base-cased. These models with a header layer form base models for our experiments.\footnote{These three models were chosen from a few other models due to their usually better performance (on the chosen datasets) compared to all the other models of the same size.  We also used {\sc Bart-large-mnli}, {\sc Roberta-large-mnli}, and {\sc Opt-1.3b} models for sentiment analysis (zero-shot classification). These models are much larger in size compared to the above-mentioned models, and therefore needs more computational power for inference.}
% ; let us denote the set of these base models as $\mathcal{B}$.
% For better readability and conciseness, we denote these pre-trained Foundation Models as $F1$, $F2$, and $F3$ respectively. 

%For conciseness, we use $\mathcal{L}$ to denote the set of these larger LLM models. 

%Table \ref{tab:common_notation} shows some of the most commonly used notations for the ease of reading.

% and in general $L_k$.
%\begin{table}
%\small
%  \caption{List of Commonly Used Notations}
%  \vspace{-8pt}
%  \label{tab:common_notation}
%  \begin{tabular}{p{2cm}p{1.4cm}p{1.4cm}}
%    \toprule
%    Term  & General Notation & Notation for Set \\ 
%\midrule
    %\multicolumn{1}{l}{Base-Model} & $B$ & %$\mathcal{B}$ \\
    %\multicolumn{1}{l}{Large FM (pre-tuned)} %&$L$ & $\mathcal{L}$\\
    %\multicolumn{1}{l}{Dataset (domain adjacent)}  & $D$ & $\mathcal{DA}$ \\
     %\multicolumn{1}{l}{\daft Model }  & $M$ %$\big(\Phi(B,D)\big)$ 
     %& $\mathcal{M}$ \\
  %\bottomrule
   % \end{tabular}
   % \vspace{-15pt}
%\end{table}

\begin{figure}[t]
\centering
\includegraphics[width=\linewidth]{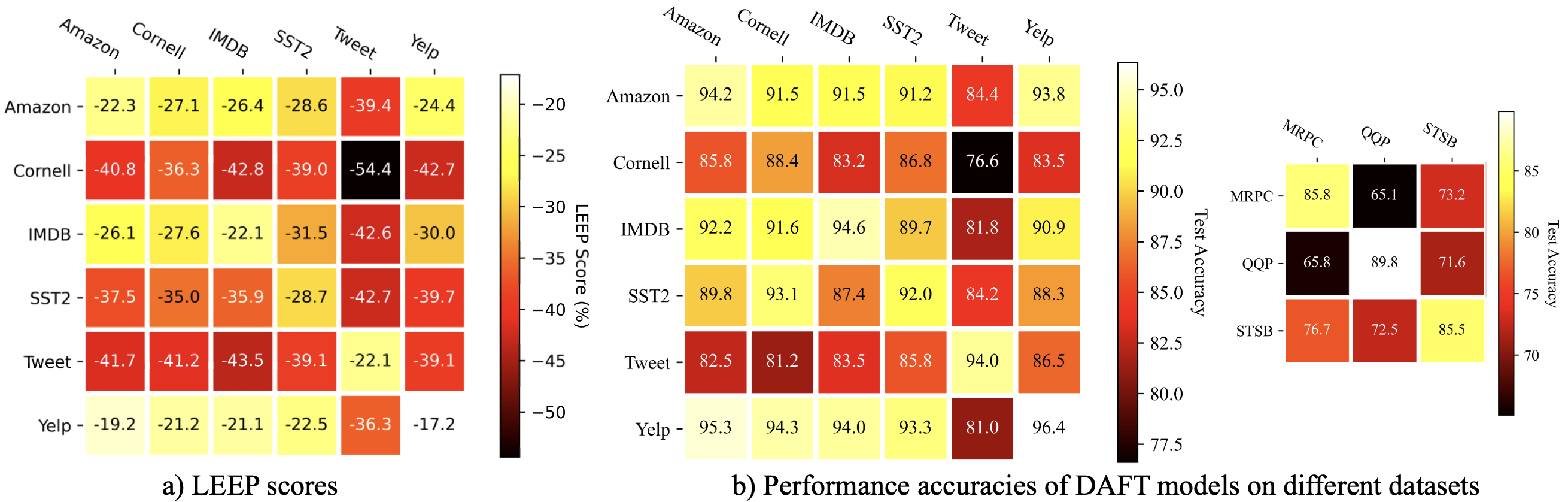}
\caption{a) LEEP score heatmap: LEEP scores signifying the transferability of knowledge from one dataset to another for sentiment analysis datasets. b) {\em Efficacy of \daft models}: The vertical axis corresponds to the data used to fine-tune a base LLM (here we show Roberta-base; the heatmaps for the other models show a similar trend and are provided in appendix), and the horizontal axis corresponds to the unseen test data. \iia{All datasets, for each of the two tasks, have matching output spaces respectively and pertain to \emph{sentiment classification} and  \emph{textual similarity} problem respectively (left and right).} The diagonal is the performance of \ft -- the LLM tested with data from the distribution it was fine-tuned with%(using a sizeable number of training examples)
, denoting the upper watermark for that test data (only exception being SST2 -- discussed in text). The off-diagonal entries correspond to the performance of the \daft -- LLM trained on one dataset, and tested on another, highlighting the potential computationally cheap (free!) zero-shot benefit that \daft models can provide. Note that, for any given test data, the performance of a \daft LLM can vary significantly. Also, LEEP score heatmap and DAFT models performance shows very similar trend.
}
\label{fig:heatmap}
\end{figure}

\vspace{-2pt}
\subsubsection{Base Models}
\label{subsec:baseline_model}
A \emph{base model} 
% (as defined in Section~\ref{sec:intro}) 
is created by adding a header layer to an LLM.
% now, let us denote the \emph{base model} with a pre-trained Foundation Model of Roberta-base ($F1$) added with a header layer (say $H1$) to perform sentiment analysis as $M1$. Similarly, we denote the base models for BERT-based-uncased and xlnet-base-cased as $M2$ and $M3$ respectively. Also, for general purpose, we use $Mk$ to denote any of the three models with $k \in \{1,2,3\}$. 
Since the base models are not trained to perform \iia{any specific task} and the header layer has not been fine-tuned, the performance of all the base models is similar to random guessing. 
% For a specific dataset we can fine-tune the base models, and depending on the extent of fine-tuning, we obtain a partial (few-shot) or fully fine-tuned model.
For some base model $B$ and dataset $D$, let us denote $\Phi(B,D,n)$ as the fine-tuned version of $B$ on $n$ amount of data from dataset $D$. If the parameter $n$ is missing as the argument, the base model is assumed to be fully fine-tuned (\emph{FFT}) on $D$. In our experiments, with few shot fine-tuning, we vary $n$ in the range of $2-256$ samples. For the (\emph{FFT}) models, we fine-tune until loss stabilization\footnote{Fine-tune until the fluctuation of loss is less than $\pm 1\%$.}. %Also, from our experiments we found that for all the six datasets being considered here, if we fine-tune any of the base models with more than $7000$ samples ($3500$ samples from each sentiment class), then the loss becomes stable and does not improve much with more data or iterations. Hence, for the rest of the paper, anytime we talk about $\Phi(B,D)$, that means the model $B$ is fine-tuned on $7000$ samples from the training dataset of $D$ (unless otherwise specified).
% \todo[inline]{We define lot of notation such as $D_T, D_a, D_k \mathcal{DA}, B, \Phi$ etc, but we do not use them often in this section. Do we need to define such notation then?}

% define the following two models. We denote the fine-tuned model as ($FT(M,D,s)$) when the base model $M$ is fine-tuned with $n$ samples (usually in the range of 2-256 samples for the experiments we did) of data from the training dataset of dataset $D$. Whereas, if the fine-tuning is done on the whole training dataset of $D$ or a large enough portion of $D$ such that the loss performance reaches a stable value (and does not improve much with more training), we define that model as a Full Fine Tuned (FFT) model and denote as $FFT(M,D)$. From our observation, we saw that for all the six datasets being considered here, if we fine-tune any of the base models with more than $7000$ samples ($3500$ samples from each sentiment class), then the loss becomes stable and does not improve much with more data or iteration. Hence, for the rest of the paper, anytime we talk about $FFT(M,D)$ models, that means they are fine-tuned on $7000$ samples of data from the training dataset of $D$ (unless otherwise specified).

\subsection{Performance of \daft Models}
\label{subsec:baseline_performance}

\begin{wrapfigure}{r}{0.5\textwidth}
\centering
% \vspace{-10pt}
\includegraphics[width=0.9\linewidth]{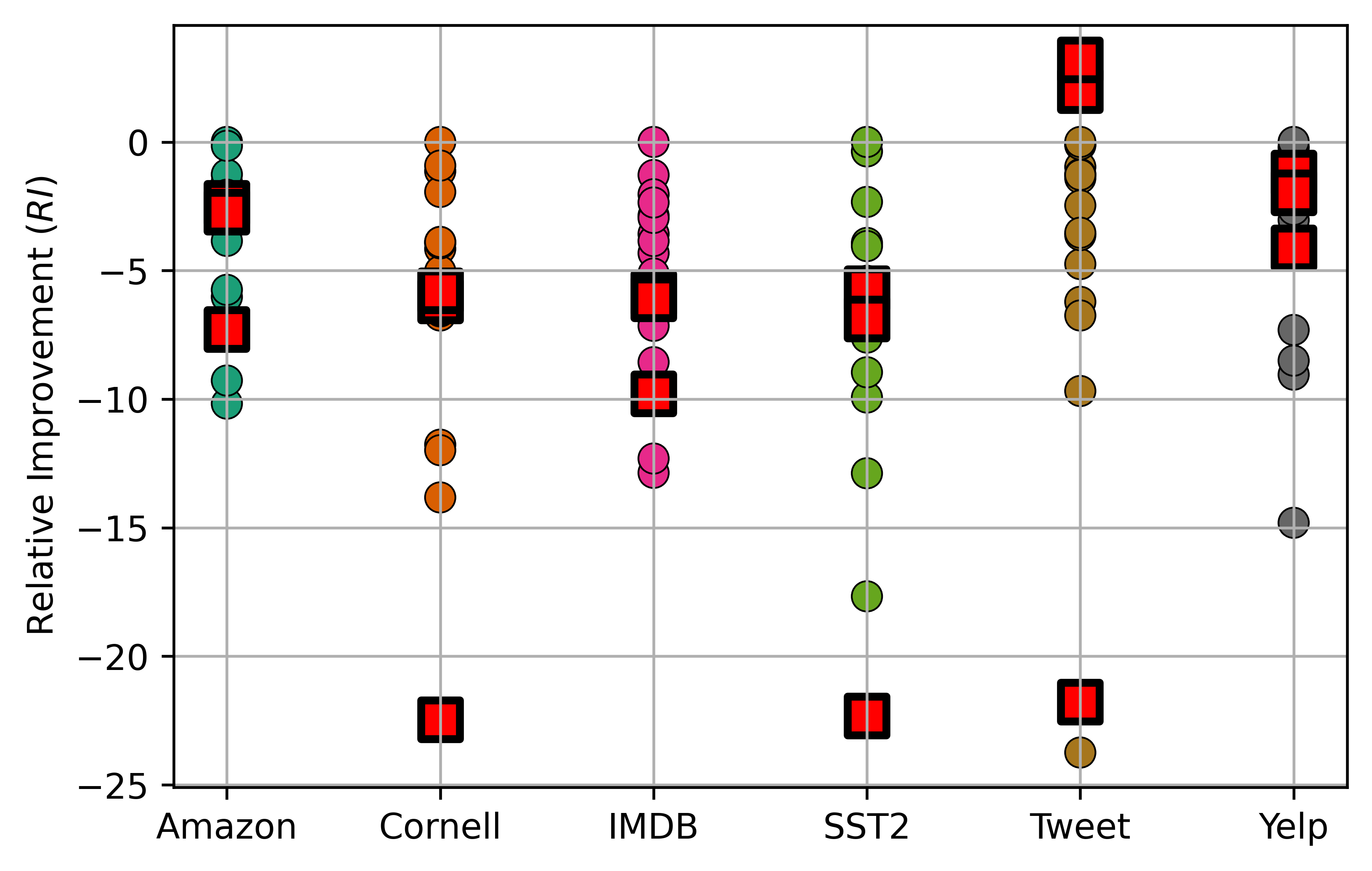}
\caption{{\em $RI$ of {\daft}s compared to the Single Best \daft} for \emph{sentiment analysis}: For each test dataset, we consider 15 \daft LLMs (fine-tuned on data different from the test data), and show the RI of \daftz over the single-best \daft LLM (out of the 15) (the colored circles $\bullet$). 
% found post-hoc (see \eqref{eq:daft-ri}). 
Values less than 0 indicate performance degradation. The red squares \textcolor{red}{$\blacksquare$} denote the performance of zero-shot classification with larger LLMs, i.e., \textsc{Bart-large-mnli}, \textsc{Roberta-large-mnli}, and {prompting with \textsc{Opt-1.3b}. }}
% As in Figure~\ref{fig:heatmap}, the results show that the performance of \daft models can vary significantly.
\vspace{-10pt}
\label{fig:RI_single_DAMs}
\end{wrapfigure}

Let us denote the train and test splits of the target dataset 
%(for the performance analysis) 
as $D_T'$ and $D_T''$ respectively. Since the standard train and test splits are identically distributed (iid), we expect a model trained on the target data ($D_T'$) to perform the best on the test data $D_T''$. This performance value is the \textit{ceiling benchmark} that we aim to match or surpass. For that purpose, we performed full fine-tuning (\emph{FFT}) on all three base models 
% ($B \in \mathcal{B}$) 
using training data of all datasets (sentiment analysis and textual similarity), to come up with %six datasets separately and coming up with 
18 \emph{FFT} models for sentiment analysis and 9 \emph{FFT} models for textual similarity. Hence, for any target dataset%, say $D_T$ 
%(among the six datasets considered)
, we can have 15 \daft models for sentiment analysis by excluding the three models that were fully fine-tuned on $D_T'$, and 6 \daft models for textual similarity task.
% More specifically, if we trained $M1$ with the training data of $D1$, then we get a model of $FFT(M1,D1)$ and for simplification let us denote the model as $M1_{D1}$ (one of the {\daft}s from the 18 {\daft}s). 
% We denote domain adjacent model (\textit{DAM}) for some dataset $Dk$ as the FFT models that are trained on $\tilde{D} \in \mathcal{DAk}$. 
% Hence, for any model $Mk$,  $FFT(M,\tilde{D})$ with $\tilde{D} \in \mathcal{DA}$ is a \textit{DAM} of $Dk$, and as an example $M1_{D2}$ is a \textit{DAM} of $D1$. 

% \todo{from here ...}

We plot the zero-shot performance of model fine-tuned on one dataset and tested on another for both sentiment analysis and textual similarity tasks in Fig.~\ref{fig:heatmap} (b). \iia{It is important to note from the figure that most of the zero-shot performances are significantly better ($\gg 50\%$) than the performance of a model with untrained header layer ($50\%$ for binary classification)}.
% \todo[color=blue!10]{Caption of fig 2 needs updating to incorporate NLI figure
% ``individual heat maps'' $\to$ ``zero-shot performance of model fine-tuned on one dataset and tested on another''}
%The caption summarizes the construction process of the heat maps.}
%The heat map has six rows and six columns, which correspond to the six datasets mentioned in Section \ref{subsec:dataset}. For some specific row and column, say row IMDB and column SST2, the value in that position, i.e., (3,4), is the model performance (accuracy in \%) when the model is trained with SST2 (training) data (i.e., \daft-SST2) and tested on IMDB (test) data.
% \todo{... to here: There is some overlap here with the caption of fig~\ref{fig:heatmap} so we can remove that explanation from the text here and keep it in the caption.}
%For the heat maps given here, we have assumed the model to be \emph{Roberta base} fine-tuned fully on the respective datasets. 
%The heat maps (for both tasks) with the other two base models showed similar patterns, and are not included here. 
% \iia{Also, the heat map for textual similarity task is provided in the appendix.\todo{if pointing to appendix, point to specific figure and section when ready}} 
% It is evident from the figure 
We also note that the performance of the Cornell and Tweet datasets do not match the other four datasets' performance on the sentiment analysis task. Our empirical study suggests that the test data of Cornell is distributionally different from the other datasets, and for all FFT models the performance was poor. 
% \todo[color=blue!10]{before getting into the negative results, it is important to highlight how most zero-shot performances are significantly $>50\%$; 
% remember that zero-shot perf with an untrained head would be around 50\%
% }
% \todo{Given Cornell's performance on model fine-tuned with Cornell is worse than its performance with \daft models, we need to update the caption of Fig~\ref{fig:heatmap}}
The Tweet dataset seems to have a different distribution in both its training and testing data (compared to the other 5 datasets). From the Tweet row and column, we observe that apart from the diagonal element, all the other accuracies are small, and a high $94\%$ accuracy is attained when the model is trained and tested on Tweet data. 
%\textit{This observation is very important in the performance analysis done in the later sections.} 
% Lastly, 
Upon inspecting the row of SST2, another interesting observation can be made, the performance of {\daft}-Cornell beats that of {\daft}-SST2. This is caused by the test dataset of SST2 being more aligned with the training data of Cornell (further details in Appendix \ref{subsec:sst2_ano}). For textual similarity datasets, the performance of \daft models was not as good as the sentiment analysis, but still show promise.
Lastly, to validate the transferability of the \daft models to a new target domain or task, we looked at the LEEP scores \citep{nguyen2020leep} for all the 6 datasets of the sentiment analysis task (Fig. \ref{fig:heatmap} (a)). Interestingly, the heat map with the LEEP scores closely follows the heat map that we generated using the performance of the \daft models (Fig. \ref{fig:heatmap} (b)). Thus, the transferability of \daft models (corresponding LEEP scores) to a target domain is 
%This bolsters the idea of \daft models and their transferability being 
proportional to their (inference) performance on that target domain. These LEEP scores also tells us that a LEEP scored based ensemble method is possible and needs further investigation. 
% \todo[color=blue!10]{Is this a 2-class or 3-class problem? Remember, again zero-shot performance is significantly better than random guessing, and we should not just focus on the negative.}

\begin{wrapfigure}{r}{0.55\textwidth}
\centering
\vspace{-28pt}
\includegraphics[width=0.95\linewidth]{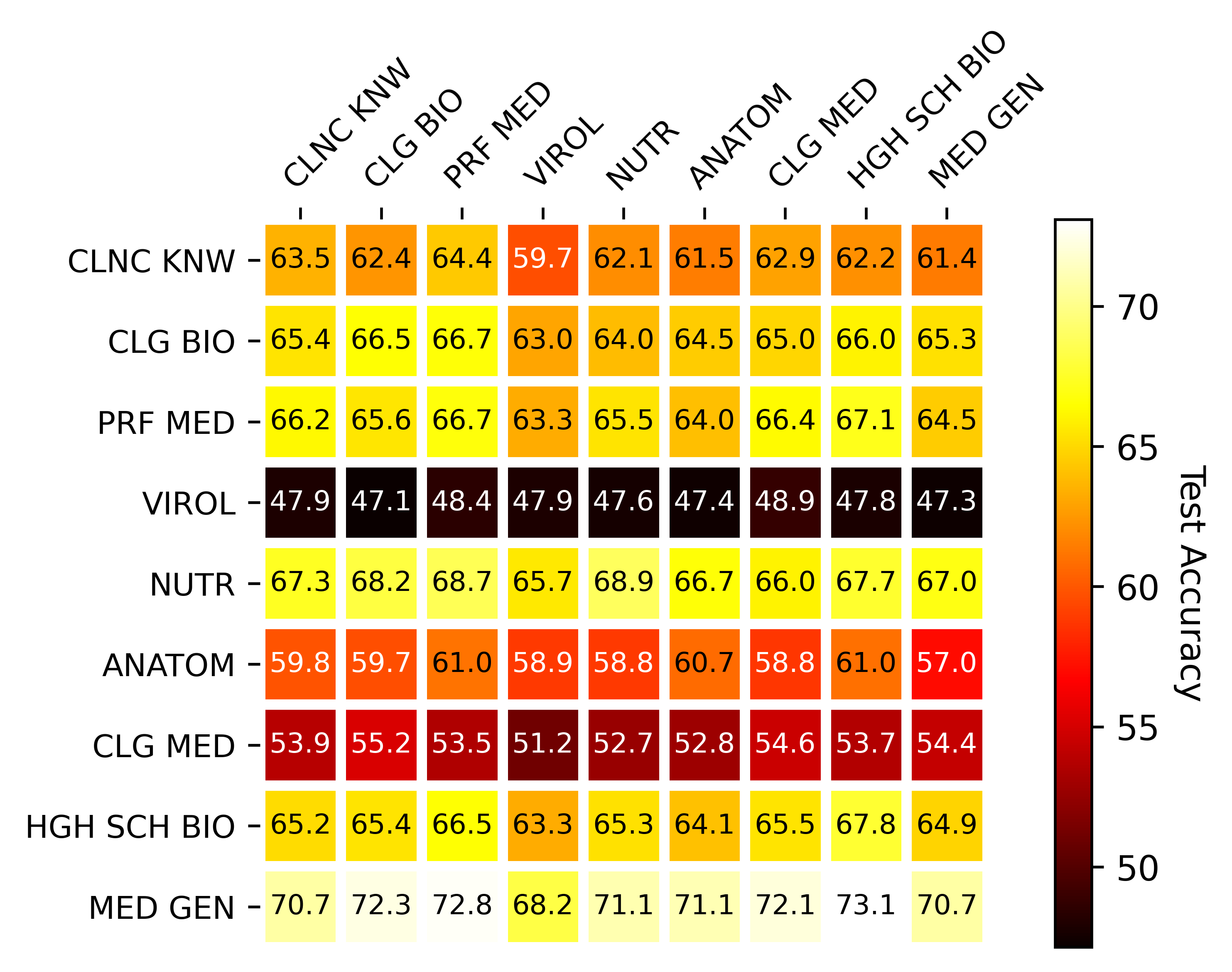}
\caption{Performance of {\daft}s on MMLU medical tasks: The 9 datasets are related to medical tasks, and the corresponding 9 {\daft} models were obtained from Hugging Face.} 
\label{fig:MMLU_heatmap}
\vspace{-10pt}
\end{wrapfigure}

To demonstrate practical feasibility of \daft models, we explored \textit{readily available} \daft models from public repositories. For target task we chose it to be question/answer in the field of {\em Medical Knowledge} and selected 9 sub-datasets from the MMLU dataset to be domain-adjacent of the target task (judging from the dataset names). These 9 sub-datasets are: Clinical Knowledge (CLNC KNW), College Biology (CLG BIO), Professional Medicine (PRF MED), Virology (VIROL), Nutrition (NUTR), Anatomy (ANATOM), College Medicine (CLG MED), High School Biology (HGH SCH BIO), and Medical Genetics (MED GEN). For each of these sub-datasets, we searched for corresponding fine-tuned models in public repositories. We found that \textit{LoRA}-tuned \daft models were available for all of them, with Llama 3.1 serving as the base model. After evaluating the performance of these \daft models on all 9 sub-datasets, we created a heatmap following a similar method used for Fig.~\ref{fig:heatmap}(b), and the result is shown in Fig.~\ref{fig:MMLU_heatmap}. In the heatmap, { \em columns} represent the publicly available (LoRA-tuned) \daft models, and {\em rows} correspond to the individual sub-datasets. It is noticeable that these \daft models exhibit highly variable performance across different sub-datasets; however, for any specific sub-dataset, the performance across different \daft models does not vary significantly.

% To check the domain adjacency between different datasets and compare the performance of different models fine-tuned fully on a domain adjacent dataset we  $D1 \neq D'$ and denoted as $M1_{D1}$(which is equivalent as ). 

\begin{conc}
% These results show that 
\daft models {\em can potentially} provide 
% very 
competitive zero-shot performance on new tasks.
\end{conc}

\subsubsection{Performance of \daftz \iia{\emph{(Zero Shot)}}}
%

%
% Here we 
% focus on the performance of the best \daft model  
% along with the Larger models (i.e., `BART',`Roberta-large' $\in \mathcal{L}$) discussed in Section \ref{subsec:models}.

% comparison of domain adjacent models \emph{DAM}. The main motivation behind using the \emph{DAM}s is that these models are ready for deployment and needs no further training (zero-shot). Since we are using them just for inference, these models can be run on low end computers as well. 
% From the discussion above, we have 15 \daft for any of the target dataset chosen for this study. 
% For a given dataset of a target task, i.e., sentiment analysis, let us denote the $i^{th}$ \daft model of that task as $M_i$ and the full set of \daft models as $\mathcal{M}$. (Note that the set $\mathcal{M}$ for one target dataset can be different from $\mathcal{M}$ for another target dataset). 
%To compare the performance of these models, we use a \emph{Relative Improvement} metric.
% \textit{the Relative Improvement ($RI$) of any \daft model $M_i$ compared to some benchmark model $M_*$ is given by;}
%     \begin{align} \label{eq:daft-ri}
%         RI_{M_i, M_*}  = \frac{P_{M_i} - P_{M_*}}{P_{M_*}},
%     \end{align}
% \todo{not sure if we need to define relative improvement}
% where $P_{M_i}$ and $P_{M_*}$ is the performance (accuracy) of the $M_i$ and $M_*$ models, respectively. 
% Let us denote the single best performing \daft for the target dataset as $M_*$. 

Fig.~\ref{fig:RI_single_DAMs} shows the relative improvement $(RI)$ of all the \daft models for sentiment analysis task when compared to the 
% $M_*$ 
single best performing \daft model.\footnote{Thus, 
%in circular markers and
% . Since, we are performing $RI$ compared to $M_*$, so 
% We expect 
all the $RI$ values (corresponding to the other {\daft}s) will be non-positive (zero for the best single \daft). The relative improvement $(RI)$ of all the \daft models for textual similarity task is provided in Appendix (Fig. \ref{fig:RI_single_DAMs_text}).}
%Moreover, we extend the definition of $RI$ for the larger models (discussed in \ref{subsec:models}) as well, 
% and perform comparison of their performance to $M_*$, 
%which are shown as red squares along the other $RI$ values (in Figure \ref{fig:RI_single_DAMs}). 
From Fig.~\ref{fig:RI_single_DAMs} it is evident that there are some \daft models that are not as good as the single best 
% ($M_*$) 
and for some specific choices (as for Tweet), the performance can be poor.
% \todo[color=blue!10]{Do we define $\mathcal{L}$ anywhere yet? Also, might be good to add how much larger these models are. }
On the other hand, for zero shot classification with \textsc{Bart-large-mnli}, \textsc{Roberta-large-mnli}, the performance were quite well. However, the zero-shot prompting using 
%, among the three larger models , and 
\textsc{Opt-1.3b} showed poor performance on most of the datasets. However, these (8 times) larger models (with zero shot classification or prompting)
% from $\mathcal{L}$ 
still fail to beat the single best \daft in all cases, except Tweet. The low performance of \daft models on Tweet is potentially because of the difference in the nature of this dataset (discussed previously in Section \ref{subsec:baseline_performance} and is supported by Fig. \ref{fig:heatmap}).

\vspace{4pt}
%\iia{A similar figure for the textual similarity task showing $RI$ of 6 \daft compared to the single best \daft is given in the appendix.\todo{as before, point to specific figure/section when ready}}
% \adjustbox{varwidth=\linewidth}{%
\begin{mdframed}[
 % userdefinedwidth= 17cm, 
align=left, linecolor=black!0,backgroundcolor=orange!10]\noindent%
% \ignorespaces
% These results show that 
Zero-shot performance of \daft models can vary significantly, with some DAFT models even outperforming larger LLMs, making the choice of the \daft model critical.
\end{mdframed}
% }%
% \todo[color=blue!10]{maybe add here that ``... vary significantly, with some DAFT models even outperforming larger LLMs, highlighting both the potential of these models and criticality of the choice of the DAFT model''  }

% that may  
% for that lies on the Tweet dataset having different characteristics compared to other dataset.
% non-domain adjacency characteristic of the Tweet dataset.

\subsubsection{\iia{\emph{Few Shot}} Comparison of \ft \& \daftsq }
\label{subsec:ft_to_daftsq}
% \todo[inline]{Should this subsection be with \S \ref{Sec:DAM} since this has nothing to do with ensembling. This could be in with the introduction of \daft models, highlighting what else we can do with \daft models, and how/why}

\begin{comment}
\begin{table*}
\small
  \caption{Average Relative Improvement of \daftsq compared to \ft}
  \label{Tab:FT_base_and_pretuned}
  \centering
  \begin{tabular}{crrrrrrrr}
    \toprule
    Dataset  & $\;\; n=2 \;\;$ & $\;\; n=4 \;\; $ & $\;\; n=8 \;\;$& $\; n=16 \;$ & $\; n=32 \;$ & $\; n=64\;$ & $\;n=128$ & $\;n=256$  \\
    \midrule
    Amazon &  75.95	& 75.25	& 79.62	
    & 65.40	& 53.51	& 34.26	& 0.81	& 0.77\\
    \midrule
        Cornell &  67.95	& 65.61	& 63.26	& 60.72	& 57.77	& 35.85 & 	8.85	& 2.09\\
    \midrule
        IMDB &  74.26& 74.64	& 76.19	& 66.01	& 58.20	& 47.64	& 0.10	& -0.16\\
    \midrule
        SST2 &  73.13	& 72.64	& 71.81 & 	71.95 & 	71.88 & 	58.58 &	12.79 &	2.99\\
    \midrule
    Tweet & 60.67& 54.25	& 55.53 &	52.25 & 	41.82 & 	25.92	& 2.34	& -0.97 \\
    \midrule
Yelp &  74.88 &	81.00	& 66.81	& 81.68 &	46.47 &	27.86 &	-0.72	& -0.73 \\
    \bottomrule
  \end{tabular}
  \vspace{-5pt}
\end{table*}
\end{comment}

\begin{table}[t]
\footnotesize
  \caption{{\em Average Relative Improvement of \daftsq compared to \ft} 
  %(aggregated over different \daft models per dataset)
  \iia{(Top 6: sentiment analysis tasks. Bottom 3: textual similarity tasks)}: Positive values show \daftsq improvement over \ft on target (domain); larger values imply higher improvements. Please see Table \ref{Tab:FT_base_and_pretuned_appnd} and \ref{Tab:FT_base_and_pretuned_appnd_text} in Appendix for detailed results. We see that \daftsq is significantly better than \ft in few-shot problems. (The same \textit{RI} values are plotted as line plot on the right to capture the trend with training data size.)}
  \label{Tab:FT_base_and_pretuned}
  \begin{minipage}{0.6\textwidth}
  \centering
  \begin{tabular}{lrrrrrrr}
    \toprule
    Target  & 2-shot & 4-shot & 8-shot & 16-shot & 32-shot & 64-shot & 128-shot \\ % & 256-sh  \\
    \midrule
    Amazon &  75.95	& 75.25	& 79.62   & 65.40	& 53.51	& 34.26	& 0.81	\\ %& 0.77\\
    Cornell &  67.95	& 65.61	& 63.26	& 60.72	& 57.77	& 35.85 & 	8.85	\\% & 2.09\\
    IMDB &  74.26& 74.64	& 76.19	& 66.01	& 58.20	& 47.64	& 0.10	\\ % & -0.16\\
    SST2 &  73.13	& 72.64	& 71.81 & 	71.95 & 	71.88 & 	58.58 &	12.79 \\ % &	2.99\\
    Tweet & 60.67& 54.25	& 55.53 &	52.25 & 	41.82 & 	25.92	& 2.34	\\ % & -0.97 \\
    Yelp &  74.88 &	81.00	& 66.81	& 81.68 &	46.47 &	27.86 &	-0.72	\\ % & -0.73 \\
%    \bottomrule
%  \end{tabular}
%  \medskip
%  \begin{tabular}{crrrrrrr}
%    \toprule
%    Dataset  & 2-shot & 4-shot & 8-shot & 16-shot & 32-shot & 64-shot & 128-shot \\ % & $\;n=256$  \\
    \midrule
    MRPC &  44.88	& 42.92 &	48.08	& 46.74	& 32.15	& 23.47 &	20.90 \\
        QQP & 72.93& 75.17 &	84.44 &	86.97 &	65.36 &	48.50 &	13.86  \\
        STSB &  48.36 &49.31 & 49.52 & 53.08& 49.88 &	29.24 &-0.12  \\
    \bottomrule
  \end{tabular}
  \vspace{-5pt}
    \end{minipage}%
  \hfill
  \begin{minipage}{0.38\textwidth}
   \vspace{-10pt}
  \centering
    \includegraphics[width=\linewidth]{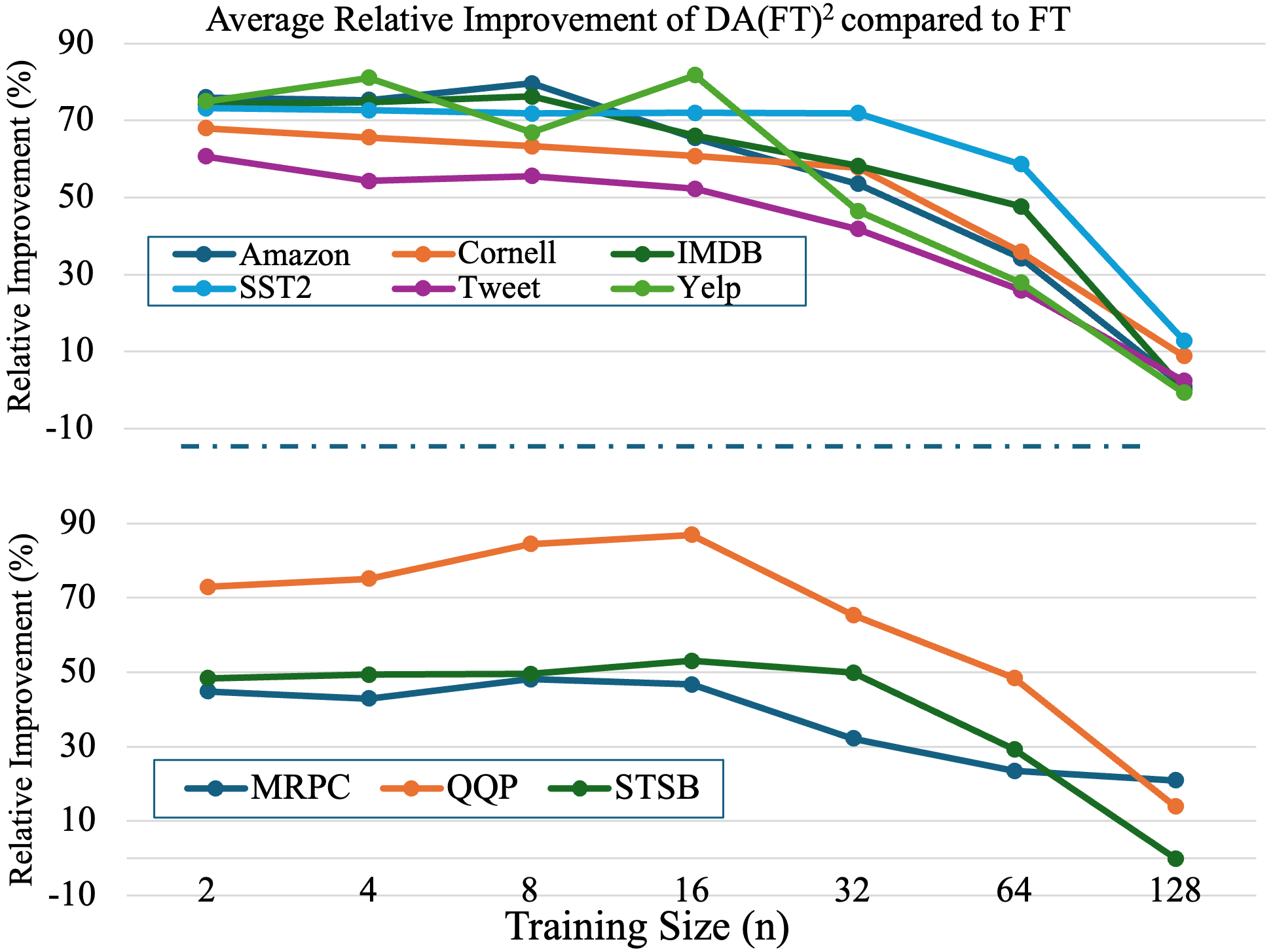}
    % No separate caption here
  \end{minipage}%
\end{table}

% \begin{figure*}
%     \centering
%     \includegraphics[width=\linewidth]{figures/Fine_tuned_base_and_pre_tuned.png}
%     \caption{Relative Improvement in performance when fine-tuning pre-tuned model}
%     \label{Tab:FT_base_and_pretuned}
% \end{figure*}

% \todo{from here ...}
% The use of \daft models when training data is unavailable (from the target domain) is discussed in the previous section. Now, let us explore the case when some

% Here we explore the few-shot case when some training data is available from the target domain. 
% \todo{... to here: We can drop the first line and just start with `` ...''}

\iia{When (some) training data is available from the target domain, the general approach}
% \todo[color=blue!10]{generalized $\to$ general}
% with any training data 
is to fine-tune a model with that data 
% so that the model can 
to perform the domain specific task. The key research question is: should we fine-tune a base model (\ft in Fig.~\ref{fig:paper_structure}), or fine-tune a \daft (\daftsq in Fig.~\ref{fig:paper_structure})? We explored two scenarios, with weight updates of i) only the header layer connection and ii) the whole model, and decided to go with the weight update of the whole model \footnote{
% In our fine-tuning experiments, w
We found that updating the weights of the header layer only does not guarantee good performance and can be surpassed by models designed to perform similar tasks without any fine-tuning (zero-shot). So, we chose to do all performance analysis %in this paper 
with fully fine-tuned models. Performance comparison of updating the weights of the entire model versus only the header layer is provided in the Appendix (Table \ref{Tab:FT_base_and_pretuned_appnd}).}.
Table \ref{Tab:FT_base_and_pretuned} provides the average performance comparison of the \ft and \daftsq models 
% for sentiment analysis task 
using the \emph{Relative Improvement} ($RI$) metric when the number of samples to fine-tune 
% \ft and \daftsq 
is varied from $n = 2$ to $128$ (Full results 
% \iia{along with statistical significance comparison} 
% for both tasks 
are in Appendix, Table \ref{Tab:FT_base_and_pretuned_appnd} and \ref{Tab:FT_base_and_pretuned_appnd_text}) \footnote{The $RI$ values in the table are generated with \emph{Roberta} as the base models for both \daftsq and \ft.}. 
A positive (negative) value of $RI$ means \daftsq is performing better (worse) compared to \ft, and the larger (smaller) the value the better (worse) \daftsq is performing. We observe that \daftsq outperforms \ft by a large margin when $n$ is small ($2$ to $64$), and see only one negative value in the table ($n = 128$). This indicates that \daftsq is outperforming \ft in most \iia{(few shot)} cases \iia{on the target task}. {Please note that if there are a moderate amount of data-samples
(for the current tasks, $n > 128$) available from the target domain (not a data-scarce environment) \ft is
expected to perform similarly as \daftsq.
% \todo[color=blue!10]{``most cases for this target task.''?}
% however those values are also very small. 
% and mostly happens when we are using \daft - Tweet. 
% The performance trend \daft - Tweet remains similar with the increase of fine-tuning data to $n = 256$ and gives larger negative $RI$ values in three cases. 
% On the other hand fine-tuning of \daft - IMDB always resulted in 
% \sout{The presence of almost all positive values of $RI$ showcases the importance of using a fine-tuned \daft instead of \ft when the sample size is small. }
%It is important to note that \daftsq usually outperforms \ft when fine-tuning is done with small to medium data samples. When the size of $n$ grows larger, at around a value of $n = 1024$, we observed that in almost all cases the \ft model performed very similarly to the corresponding \daftsq model.  
% \todo[inline]{Things are a bit confusing here -- if we pick a \daft model, and fine-tuned it with task specific data, we would get \daftsq. If I read table~\ref{Tab:FT_base_and_pretuned}, it seemed to be that \daft-IMDB for Amazon dataset meant that I chose the \daft that was fine-tuned with IMDB, and then fine-tuned further on top of it with Amazon. Is that correct? If so, then entries for (IMDB, \daft-IMDB) and (Tweet, \daft-Tweet) seem a bit weird. We are only considering \daft models that were originally fine-tuned with a different dataset}

\begin{conc}
% These results indicate that 
For few-shot tasks, fine-tuning \daft models, that is \daftsq, is generally more beneficial than fine-tuning base models (\ft).
\end{conc}

% On the other hand, We can use this $FT(M1,D',k)$ model to perform our target task, i.e., sentiment analysis, or we can use $FFT(M,D)$, a model that is fully fine-tuned on a domain adjacent dataset $D1 \neq D'$, which denoted as $M1_{D1}$(which is equivalent as $FFT(M1,D1)$). 

% \sout{The performance of \daft models showcases the significance of using them for domain adjacent tasks.} 

\begin{wrapfigure}{r}{0.5\textwidth}
\centering
\vspace{-10pt}
\includegraphics[width=0.9\linewidth]
{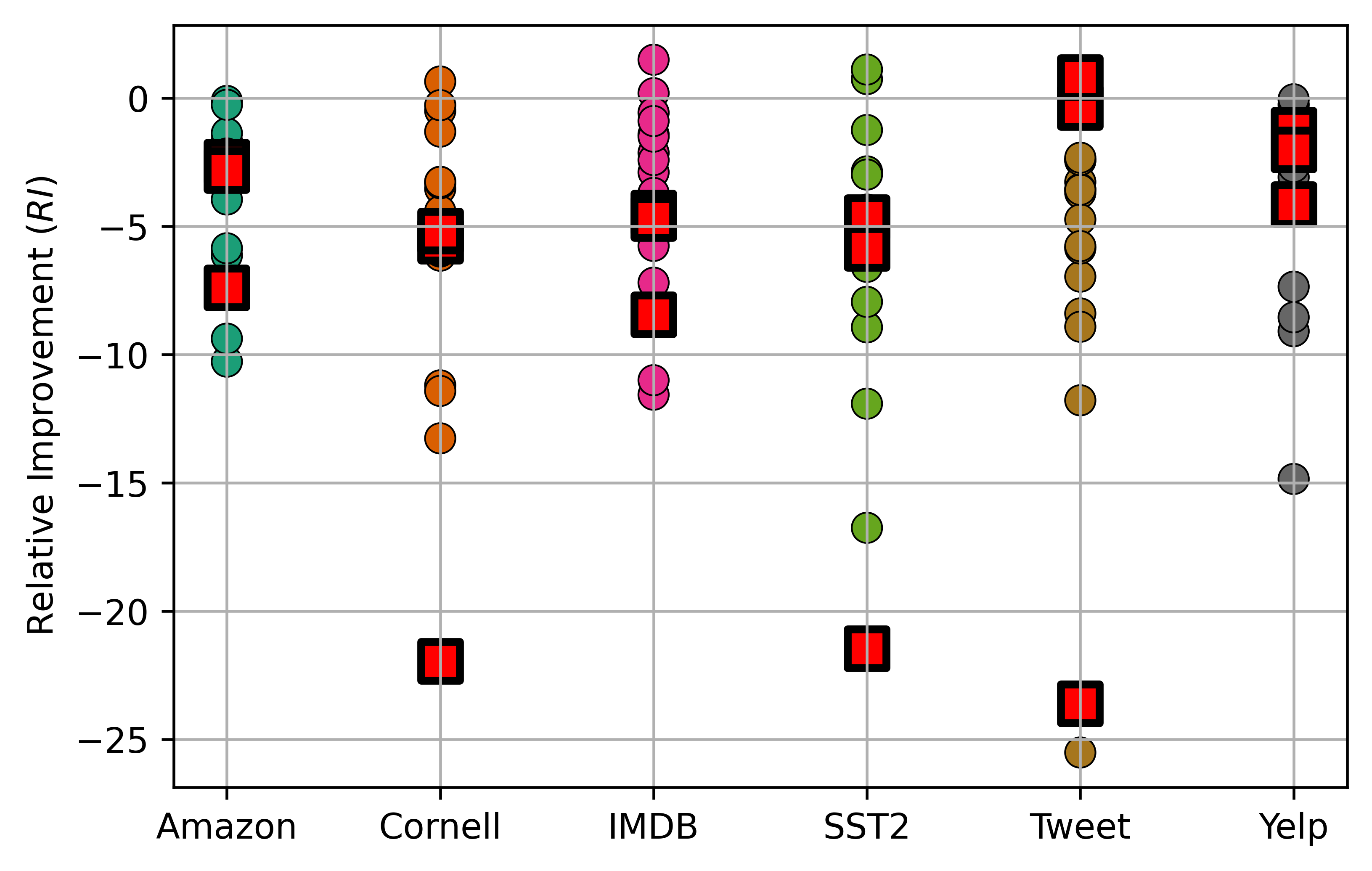}
\vspace{-8pt}
\caption{{\em RI of {\daft}s over \daftez for sentiment analysis}: For each test dataset, we consider 15 \daft LLMs (fine-tuned on data different from the test data), and consider the RI of \daftz over 
%the average ensemble 
\daftez of 15 \daft LLMs. Values less than 0 indicate performance degradation. The red squares \textcolor{red}{$\blacksquare$} denote the performance of zero-shot classification on larger LLMs (\textsc{Bart-large-mnli}, \textsc{Roberta-large-mnli}), and prompting with \textsc{Opt-1.3b}. The results show that \daftez usually has significantly better zero-shot performance than \daftz. 
However, there are some cases where some specific \daft models can outperform the average ensemble (leading to $\text{RI}>0$).}
\label{fig:RI_Ens_DAMs}
\vspace{-45pt}
\end{wrapfigure}

From a practical standpoint, a major advantage of using \daft models is that we get them for \emph{free}. However, as our results show, there is considerable uncertainty about what is the right \daft model to choose (especially in the zero-shot setting), as the performance can significantly vary between different \daft models (Fig. \ref{fig:RI_single_DAMs}). \iia{Also, note that while \daftz is zero shot, \daftsq is computationally expensive, needing back-propagation. Hence, 
in the next section, we explore the use of ensemble methods, which is a cheaper alternative and addresses the model selection problem.}
% \todo[color=blue!10]{maybe refer to Fig 3 here

% also note that while daftz is zero shot, daftsq is computationally expensive, needing backprop. so we consider cheaper alternatives
% }

% \todo[inline]{need to conclude section by saying something of the effect that ``\daft models can be useful, but the performance depends heavily on which one we choose, and that choice is hard to make ahead of time. Thus, we address this selection problem with ensembling.''}
\section{Ensembling of \textsf{DAFT} Models}
\label{sec:empirical_eval}

In this section, we evaluate the performance of the \daft models using two distinct \textit{Ensemble} methods and benchmark their performances. \iia{Advantages of using ensemble methods are discussed in Appendix \ref{subsec:adv_ensmbl}.}

\subsection{\daftez Performance \emph{(Zero Shot)} % \daft and Average Ensemble
}

The average Ensemble (\daftez) method is a zero-shot method that can be used as an alternative to randomly picking one of the \daft models. In \daftez, the \emph{output probabilities} from all the \daft models are averaged and the decision is taken on that value (max-voting is another feasible approach). 
To perform \daftez, we run inference on all the \daft models. However, since these \daft models do not need any further training or fine-tuning and inference can  be run in parallel, it is feasible to utilize them efficiently with the abundance of computational devices (mostly low end) available today.

Fig. \ref{fig:RI_Ens_DAMs} shows the \emph{Relative Improvement} 
%(as defined in Section \ref{subsec:baseline_performance}) 
of all the 15 \daft models and the three larger models (as previously discussed in \ref{subsec:models}) for sentiment analysis, compared to the average ensemble of the \daft models, i.e., \daftez. \iia{There are  very few cases ($5/90$ for \daftz, and $1/18$ for larger LLMs) in Fig. \ref{fig:RI_Ens_DAMs} where a positive $RI$ value is seen}, implying that the average ensemble method (\daftez) is better than most of the \daft models.
% \todo[color=blue!10]{``... very few points have ...'' $\to$ ``.. in  ...''}
Moreover, the performance of the larger LLMs (red squares)
% ($\in \mathcal{L}$) 
are not as good as \daftez.
% \todo{I do not see $\mathcal L$ defined anywhere earlier in the text.} 
Furthermore, for the Tweet dataset, for which none of the \daft models performed well and was beaten in performance by two of the larger models (\textit{Bart-large-mnli}, \textit{Roberta-large-mnli}), we now have $RI$ values close to zero. \iia{Furthermore, for the zero-shot Tweet task, while the larger models outperformed all the \daftz models (Fig \ref{fig:RI_Ens_DAMs}), \daftez is able to match their performance, highlighting the utility of ensembles.}
Hence, \daftez has a performance very close to those larger models and shows the importance of using an ensemble even when individual \daft models cannot perform well.
For text similarity task we see a similar trend in the $RI$ performance (Fig. \ref{fig:RI_Ens_text}).

\begin{wrapfigure}{r}{0.46\textwidth}
\centering
\vspace{-15pt}
\includegraphics[width=0.8\linewidth]{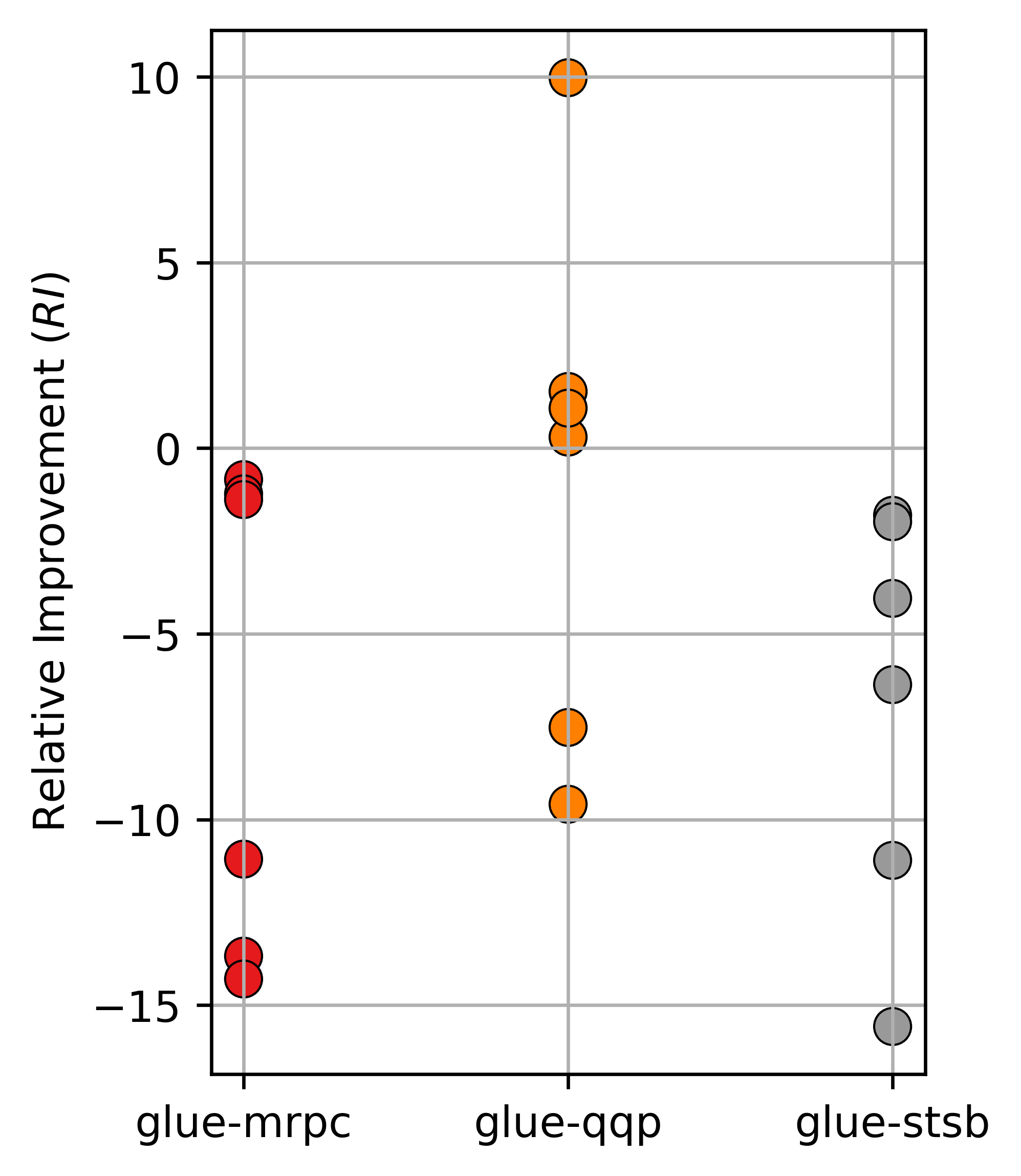}
\vspace{-7pt}
\caption{{\em Relative Improvement of {\daft}s compared to \daftez} for Text similarity task: For each test dataset, we consider 6 \daft FMs (fine-tuned on data different from the test data), and consider the RI of \daftz over the \daftez. 
Values less than 0 indicate better performance of \daftez. }
\label{fig:RI_Ens_text}
\vspace{-45pt}
\end{wrapfigure}

Lastly, to evaluate the performance of \daftez on practical and modern LLM benchmark datasets, we performed an ensemble of 8 \daft models across the 9 datasets selected for the target task of {\em Medical Knowledge} (see Section~\ref{subsec:baseline_performance} for details on the datasets). The results are shown is Fig. \ref{fig:RI_Ens_MMLU}. We observe that \daftez outperforms the base model (Llama 3.1) on all instances (red squares with black border) and surpasses individual \daft models in 56 out of 72 cases (circles in the figure).
% \todo[color=blue!10]{Consider this for last two lines:
% ``''}
%Overall, the zero shot ensemble may be computationally exhausting, but can be deployed parallely and performs well in different situations compared to the \daft models. 
In Section \ref{sec:theory}, we argue theoretically (Proposition \ref{prop:daftez})
% \todo{wrong proposition number}
that it is better (in an expected sense) to use \daftez over choosing a \daft model randomly from the set of {\daft}s.
% \todo[color=blue!10]{have we defined $\mathcal M$ yet?}

\adjustbox{varwidth=\linewidth}{
\begin{mdframed}[ 
userdefinedwidth= 8.5cm,
align=left,
linecolor=black!0,backgroundcolor=orange!10]\noindent%
% \ignorespaces
For zero-shot tasks, just a simple ensemble of \daft models (\daftez) is on average a better choice than any individual \daft model.
\end{mdframed}
}

The average ensemble is therefore a great data-agnostic solution;
% as a zero-shot method for inference, and the performance of average ensemble with respect to the individual \daft models supports that. 
however, when training data from the target domain is available, we show that the data-informed ensembling can be a stronger choice. We thus pursue a weighted ensemble of the \daft models (\dafte) in the following section.

% \todo[inline]{need to conclude this subsection to motivate the weight learning where we discuss how we are leaving some performance on the table with average ensembling}

\subsection{\iia{\emph{Few-Shot}} \dafte and \daftsq}

\begin{table}[t]
\footnotesize
  \caption{Average $RI$ of \daftsq compared to \dafte \iia{(Top 6: sentiment analysis tasks. Bottom 3: text similarity tasks)}. Negative values indicate that \dafte is better than \daftsq on target (task); larger values imply higher improvements. Table \ref{Tab:FT_pretuned_and_Ens_weight_appnd} and \ref{Tab:FT_pretuned_and_Ens_weight_appnd_text} in the Appendix gives more details. In the very few-shot setting, \dafte outperforms \daftsq in almost all cases.}
  \centering
  \label{Tab:FT_pretuned_and_Ens_weight}
  \centering
  \begin{tabular}{crrrrrrr}
    \toprule
    Target  & 2-shot & 4-shot & 8-shot & 16-shot & 32-shot & 64-shot & 128-shot \\ % & $\;n=256$  \\
    \midrule
    Amazon &  -4.03 &	-2.48	& -2.38	&-2.42	& -2.66 &	-2.95 &	-2.06 \\%&	-1.50\\
        Cornell & -3.94	& -3.60	& -3.37 &	-3.54 &	-4.13	& -3.91 &	-2.53	\\%& -2.65\\
        IMDB &  -2.70	& -3.41 &	-3.35	& -3.31 &	-3.04 &	-2.87	& -2.83 \\% & -1.45\\
        SST2 &  -4.67	& -3.32 &	-3.94 &	-4.87 &	-4.11 &	-4.68 &	-4.21 \\% &	-3.75\\
    Tweet & -4.62	& -4.91 &	-5.14 &	-5.54 &	-4.00 &	-3.82 &	-1.59 \\%&	-0.39 \\
Yelp &  -2.56 &	-2.76 &	-2.10	& -2.09	& -1.68 &	-2.46 &	-2.28 \\%&	-1.29 \\
%    \bottomrule
%  \end{tabular}
%
\midrule
%  \medskip
%  
%  \begin{tabular}{crrrrrrr}
%    \toprule
    % Dataset  & 2-shot & 4-shot & 8-shot & 16-shot & 32-shot & 64-shot & 128-shot \\ % & $\;n=256$  \\
    % \midrule
    MRPC &  -5.19	& -4.34 &	-4.11	& -3.60 &	-3.12 &	1.53 &	4.74 \\
        QQP & -2.85	& -2.64	& -0.46	& -0.06 &	2.01	& 2.41	& 4.42 \\
        STSB &  -5.29	& -5.77	& -4.07 & 	-1.89 &	-1.97 &	0.57	& 1.28 \\
    \bottomrule
  \end{tabular}
  \vspace{-5pt}
\end{table}

We define \dafte as the weighted ensemble method, where the weights of the ensemble layer are learned using the training data from the target dataset. To perform the weighted ensemble in \dafte we utilized two specific regression methods, a) Random forest-based regression ($RF$) and b) SGD-based linear regression ($LR$)\footnote{Parameters of these methods are given in Appendix \ref{appnd_subsec_ens_layer}.}.
% \iia{We observed that, i) With less sample data $LR$ performs much better with less deviation in performance, compared to $RF$; and 
%ii) the average performance of $LR$ is better than  $RF$ for smaller sample data, and i
% ii) $RF$ performs better with higher sample data and specifically when \daft models perform subpar on the target dataset (i.e., \daft models of text similarity). 
Since, $LR$ is more lightweight with comparable or better performance than $RF$, we considered $LR$ to generate the results of \dafte (More discussion on $LR$ and $RF$ performance comparison with the results are in Appendix: Table \ref{Tab:FT_pretuned_and_Ens_weight_appnd} and \ref{Tab:FT_pretuned_and_Ens_weight_appnd_text}).
% , and in Section \ref{limitations}: Limitations.
% \todo[color=blue!10]{Above description is nice -- can move to appendix if space needed and just say that we have tried two diff weight learning schemes, and LR is more lightweight with comparable or better performance than RF, so we just consider it here.}
% with $LR$ is not significant,  %Since, the $LR$ method is very lightweight and the weight of the ensemble layer can be calculated very quickly given the output from the \daft models. Hence, w
% We chose to go with $LR$ method to perform the weighted ensemble for \dafte.
% due to better after close inspection decided to use $LR$ for $n \leq 8$ and $RF$ for $n>8$ (details in 
 %Overall \dafte has weights ($w_i$) for each of the models ($M_i$) and \emph{optimizes the weights such that the output probabilities multiplied by their respective weights results in minimum loss} (maximum performance).

% \textcolor{red}{
% \begin{align}
%     \ell(\text{\dafte}) =  \min_{w_i, i\in \mathcal{N}} \left[ \ell \left(\sum_{i} w_i M_i \right)\right],
% \end{align}
% \noindent where  $w_i$ is the ensemble weight for the $i^{th}$ \daft model ($M_i$)}

% \todo[inline]{We need to describe what we mean by \dafte here. We have not mathematically defined it at all. While average ensemble is understandable, we need to clearly define \dafte here. We also need to discuss how we learn the weights, and how computationally efficient the process is.}
% Also, we need to bring in table~\ref{tab:cc} here in the text in this section 

\begin{wrapfigure}{r}{0.46\textwidth}
\centering
\vspace{-8pt}
\includegraphics[width=\linewidth]{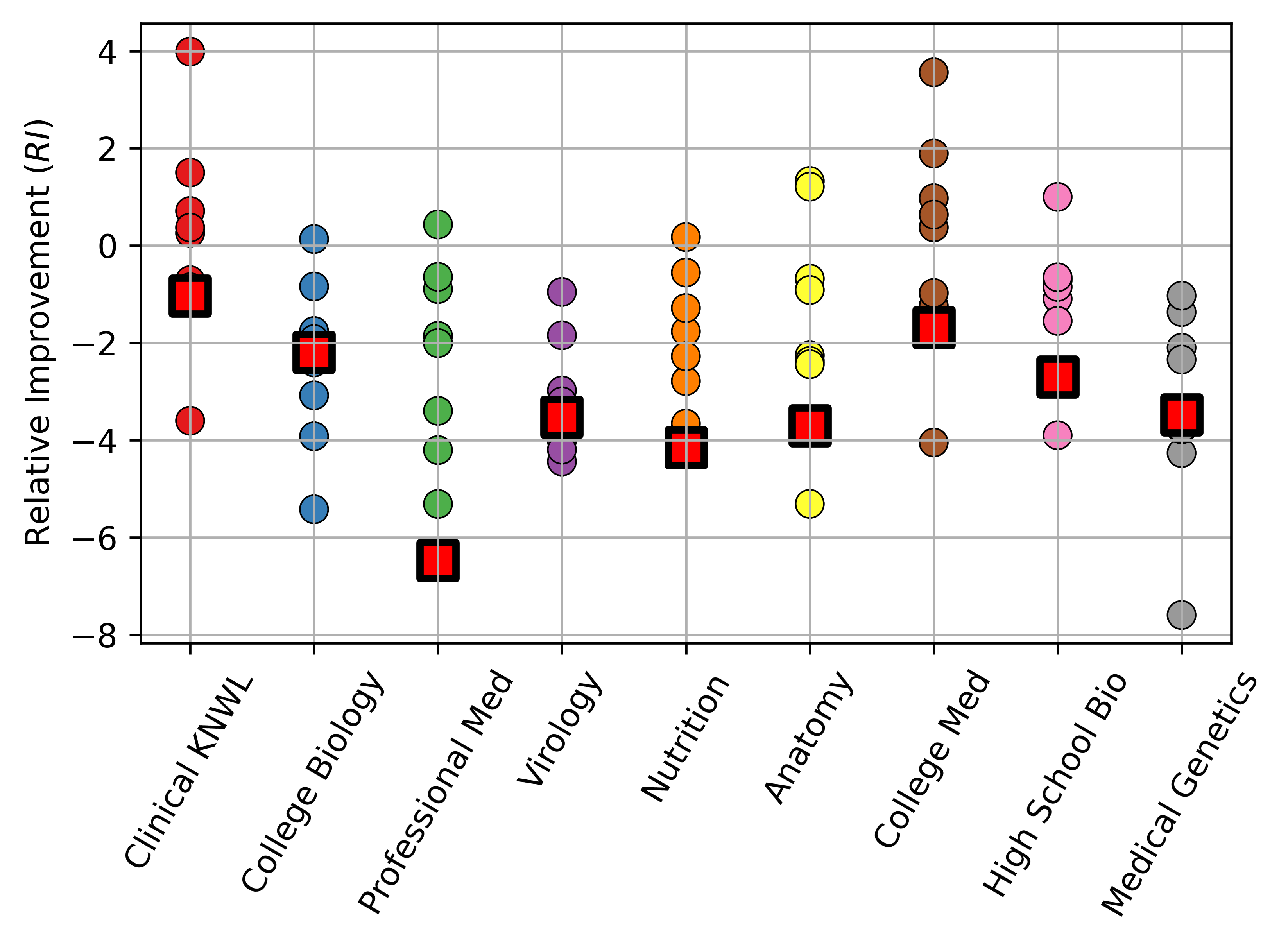}
\vspace{-12pt}
\caption{{\em Relative Improvement of {\daft}s compared to \daftez} for {\em Medical Knowledge} task: For each dataset, we consider 8 {\daft}s, and consider the RI of \daftz over the \daftez. 
Values less than 0 indicate better performance of \daftez. }
\label{fig:RI_Ens_MMLU}
\vspace{-10pt}
\end{wrapfigure}

In Section \ref{subsec:ft_to_daftsq}, we observed that \daftsq usually performs better than \ft for few-shot learning. Hence, we compare the performance of \dafte (weighted ensemble) with \daftsq. For a fair comparison, the same amount of data is used when fine-tuning \daftsq and learning the weights of the ensemble layer.
Table \ref{Tab:FT_pretuned_and_Ens_weight} shows the average RI of \daftsq compared to \dafte. 
%\\
% \textcolor{red}{I will rewrite this paragraph, still not sure which way to go. If we should use LR for sentiment, but RF for text similarity or LR for both. LR for text similarity gives poor results.}\iia{where the \dafte values for sentiment analysis is calculated using $LR$ method, whereas for text similarity we used $RF$ method.}
% performance comparison of \daftsq models with \dafte using the
% , as defined in Section \ref{Sec:DAM}. 
An $RI$ of positive (negative) means that \daftsq performed better (worse, resp.) compared to \dafte. 
% Due to space constraints the average $RI$ values are shown in Table \ref{Tab:FT_pretuned_and_Ens_weight}, and the full result is given in the Appendix (Table \ref{Tab:FT_pretuned_and_Ens_weight_appnd}). 
% \iia{The reason of using  did perform both methods, and detailed results are on  in the Appendix}. 
From Table \ref{Tab:FT_pretuned_and_Ens_weight} it is evident that \dafte outperforms \daftsq for sentiment analysis tasks, i.e., \emph{all} values are negative in the table. On the other hand, for textual similarity  \dafte usually performs well for smaller data samples 
%(except STSB, see Appendix~\ref{subsec:stsb_ano} for discussion),
% \todo{we need to point to the specific appendix section}
%), 
but under-performs on more-data (positive $RI$ values). However,  $RF$ performed well with more samples and actually performs better or similarly to \daftsq for $n>16$ (Table \ref{Tab:FT_pretuned_and_Ens_weight_appnd_text_RF} in the Appendix). This shows that weighted ensembling is effective, with task-specific variations between $RF$ and $LR$.
% does not perform that well and in many case the $RI$ value is in favor of \daftsq by a good margin.
% and 
%with $n$ varying from a small value of $2$ to $256$ samples ($1$ to $128$ pairs). 
% This proves that for few-shot learning, using \dafte is in general a much better option.
Note that
% the 
updating 
% of weights in 
the ensemble layer depends on the predictions of the \daft models, and requires solving either (i)~a simple linear equation to optimize the weights of the ensemble layer, or (ii)~training a $RF$ with shallow trees.
% \todo[color=blue!10]{``running ... RF'' $\to$ ``training a RF with shallow trees''}
Thus, {\em only a single inference on each of the \daft models} with the few-shot samples is necessary \footnote{In our evaluations, we found that \dafte puts much more weight on the top few DAFT models. 
% we can filter out those models as the candidate of \dafte for inference with test data.
Hence if we have some training data from the target domain and have some budget constraints (e.g., the number of models),  one solution could be to pick the models with the highest weights (from \dafte LR). Moreover, other transferability metrics can be used to select the top models within a budget. Some of those transferability metrics are LEEP score, MS-LEEP, E-LEEP, and SoftIoU-EEP \citep{agostinelli2022transferability}. LEEP scores can be calculated for each model individually (as discussed in Section \ref{Sec:DAM}). The other metrics, i.e., MS-LEEP, E-LEEP, and SoftIoU-EEP, are computed for a set of models and can be an alternative method to LR in DAFT-E to ensure better transferability for an ensemble of models.}. In \daftsq, where we fine-tune a \daft model, we need to perform \emph{computationally expensive} back-propagation through the large model to update weights.

% In DAFT-E the model weights (from LR) may differ a lot; given some budget constraints (e.g., number of models), one solution could be to pick the models with the highest weights. Moreover, other transferability metrics can be used to select the top models within a budget. Some of those transferability metrics are LEEP score, MS-LEEP, E-LEEP, and SoftIoU-EEP [B]. LEEP score can be calculated for each model individually (as discussed in Section 2). The other metrics are calculated for a set of models and can be a great solution to ensure the best transferability for an ensemble of models within a computational budget.
% \todo[color=blue!10]{In contrast, to fine-tune ...'' $\to$ `In daftsq, where we fine-tune ...'' }

% \todo[inline]{Same as with \ft vs \daftsq, we should not be considering \daftsq with base models that were fine-tuned with the same data as in (IMDB, \daft-IMDB), and (Tweet, \daft-Tweet) }
%\todo[inline]{Also, we should definitely have a \dafte vs \daftez}

\begin{conc}
For few shot tasks, (cheap) \emph{weighted} ensemble of {\daft}s (\dafte) can outperform (expensive) \daftsq.
% , with a suitable ensembling method.
\end{conc}
% \todo[color=blue!10]{maybe we add ``cheap weighted ensembling'' and ``expensive daftsq'' }

\subsection{Overall Comparison}

\begin{figure*}
    \centering
    \includegraphics[width=\linewidth]{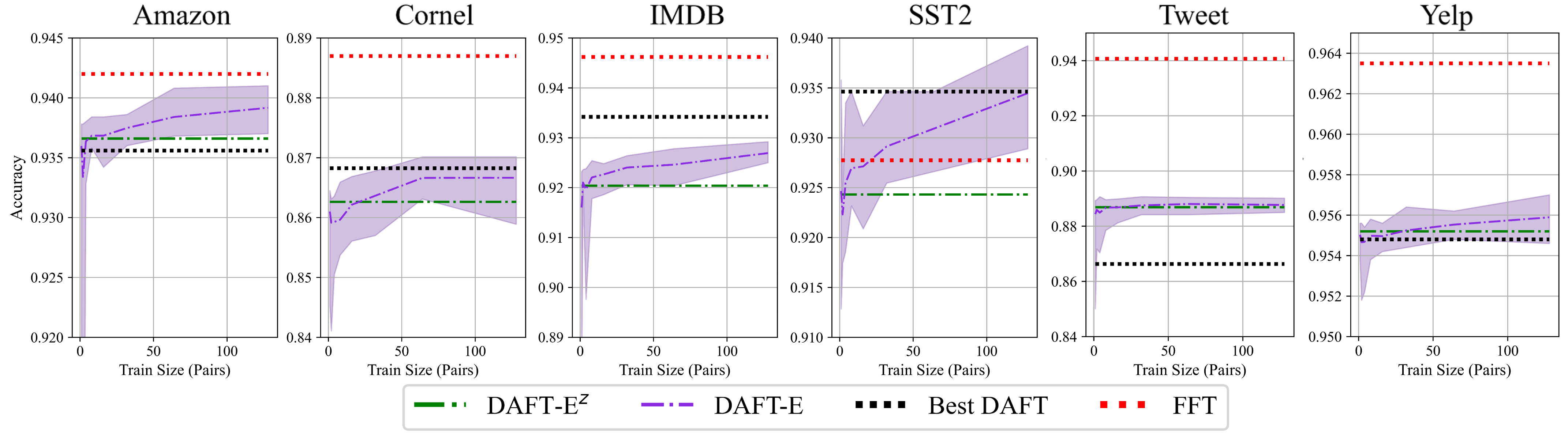}
    \vspace{-23pt}
    \caption{{\em Performance comparison of \daftez, \dafte, single-best \daft and FFT.} Error interval for \dafte is based on the random choice of the few-shot samples used to learn the weights of the ensemble aggregated over 10 trials. 
The single best \daft for different datasets: Amazon (Roberta - Yelp), Cornell (Roberta - SST2), IMDB (xlnet - amazon), SST2 (xlnet - Cornell), Tweet (xlnet - SST2), Yelp (xlnet - amazon).}
    \label{fig:Ensmbl_FFT}
    \vspace{-10pt}
\end{figure*}

\begin{figure*}[t]
\begin{subfigure}{0.32\textwidth}
  \includegraphics[width=\linewidth]{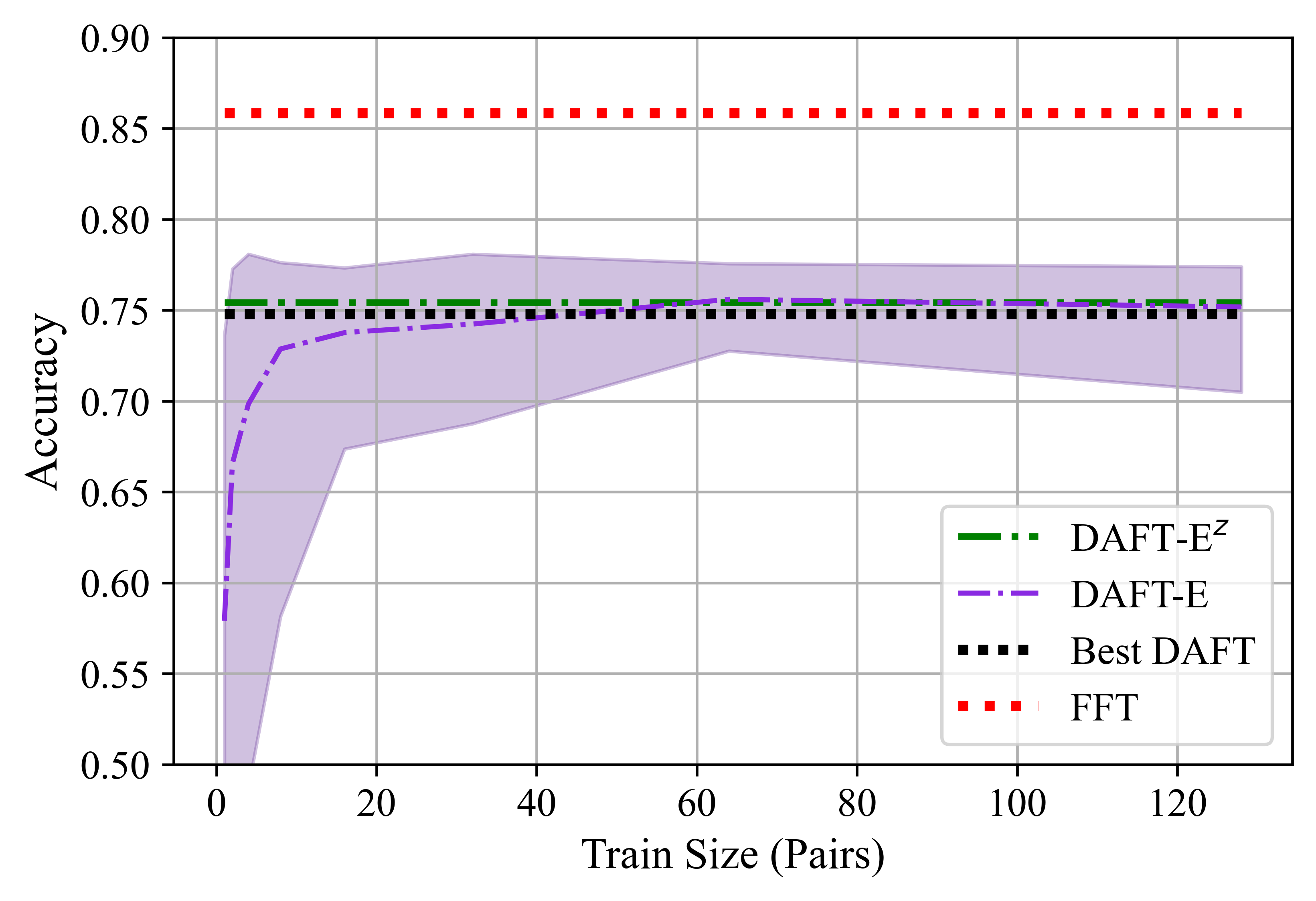}
\vspace{-18pt}
\caption{MRPC}
\label{fig:Ensmbl_FFT1_text}
\end{subfigure}
~
\begin{subfigure}{0.32\textwidth}
\includegraphics[width=\linewidth]{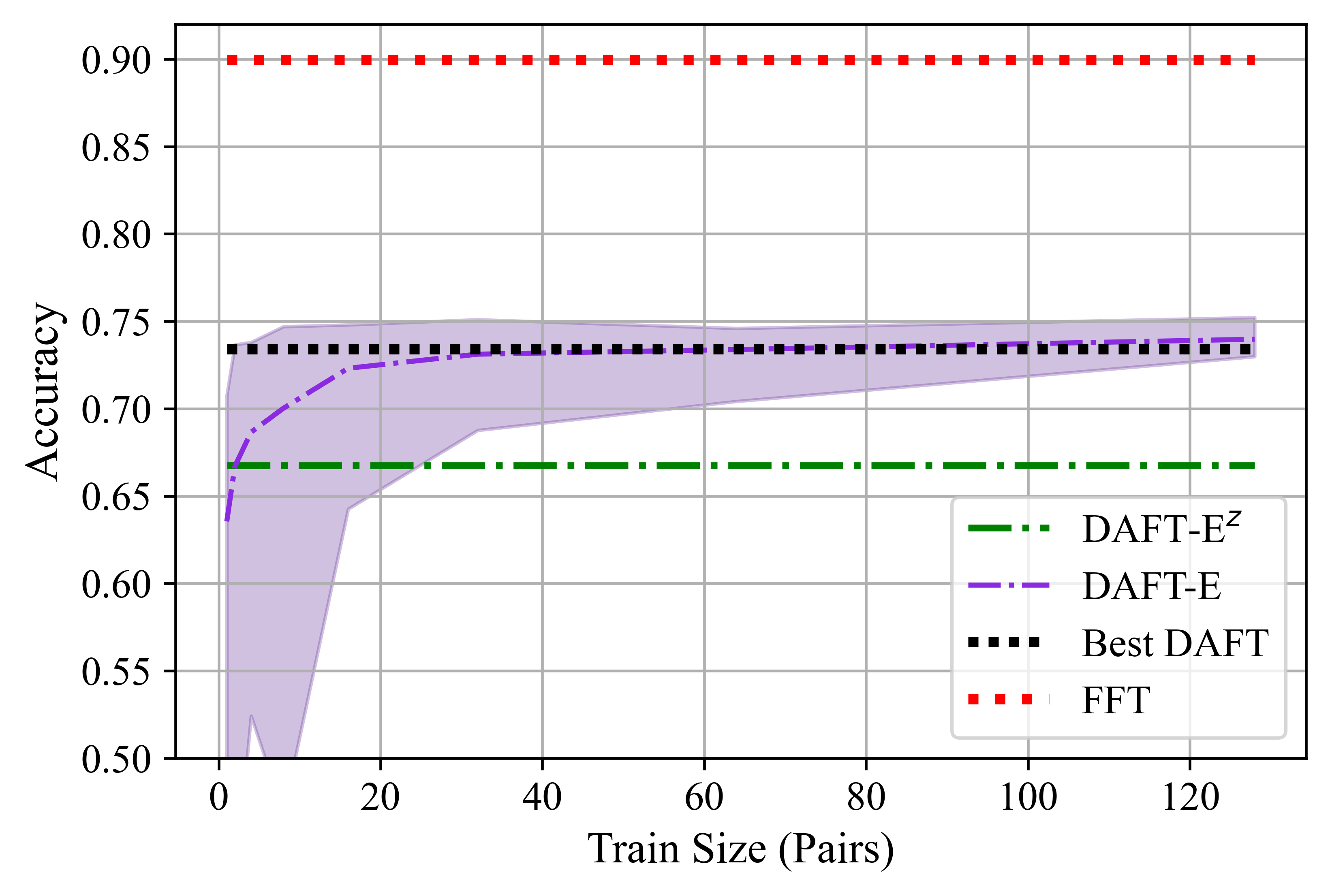}
\vspace{-18pt}
\caption{QQP}
  \label{fig:Ensmbl_FFT2_text}
\end{subfigure}
~
\begin{subfigure}{0.32\textwidth}
 \includegraphics[width=\linewidth]{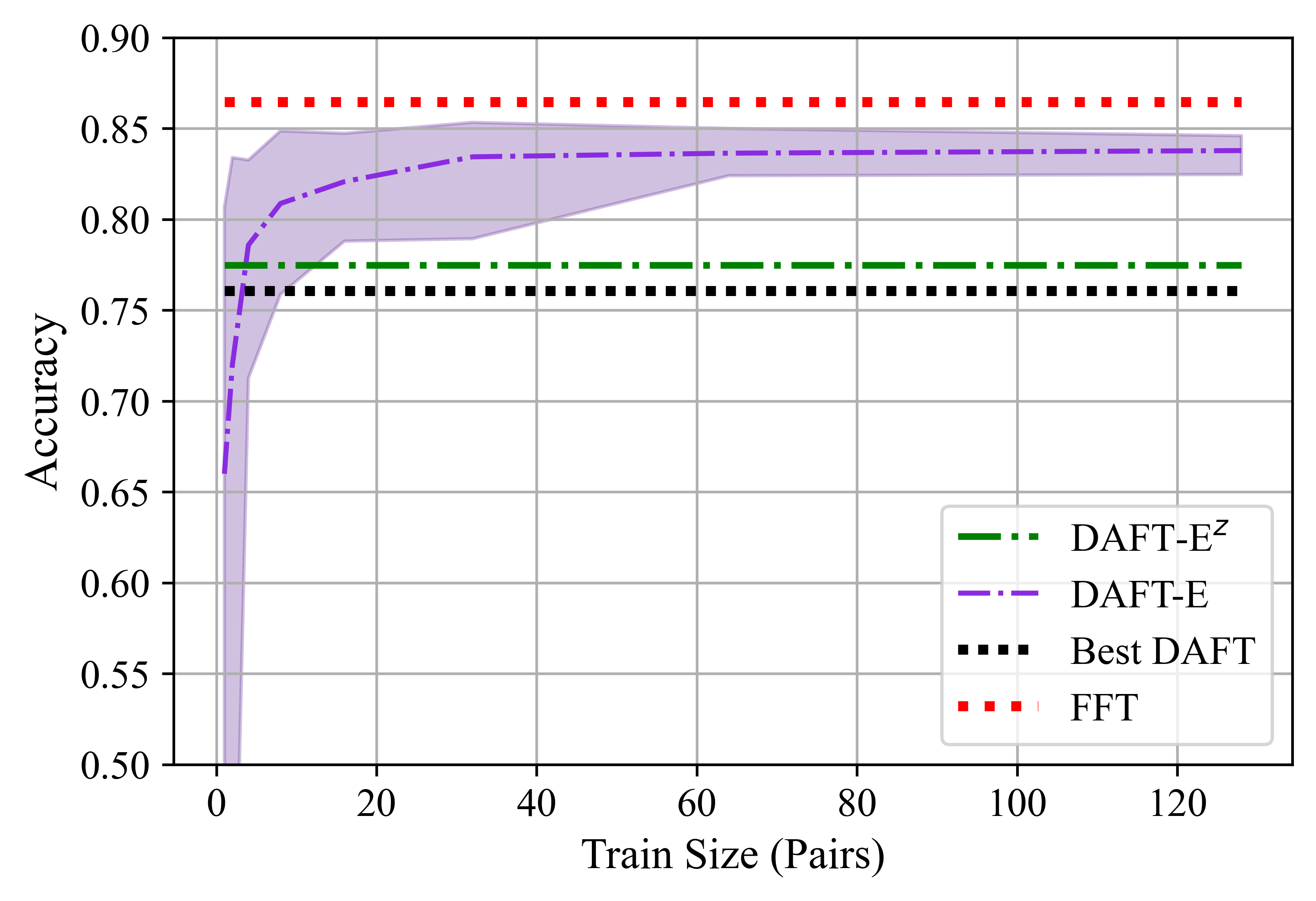}
\vspace{-18pt}
\caption{STSB}
  \label{fig:Ensmbl_FFT3_text}
\end{subfigure}
\vspace{-20pt}
\caption{{\em Performance comparison of \daftez, \dafte, single-best \daft and FFT} for Text Similarity task. The error interval for \dafte is based on the random choice of the few-shot samples used to learn the weights of the ensemble, using RF method, aggregated over 10 trials.  
% (Might be good here to list the best \daft model for each dataset, highlighting how they can vary). \\ 
The single best \daft for different datasets are: MRPC (Roberta - STSB), QQP (Roberta - STSB), STSB (Roberta - MRPC).}
\label{fig:Ensmbl_FFT_text}
\vspace{-12pt}
\end{figure*}

\begin{wrapfigure}{r}{0.5\textwidth}
\centering
\vspace{-2pt}
\includegraphics[width=0.9\linewidth]{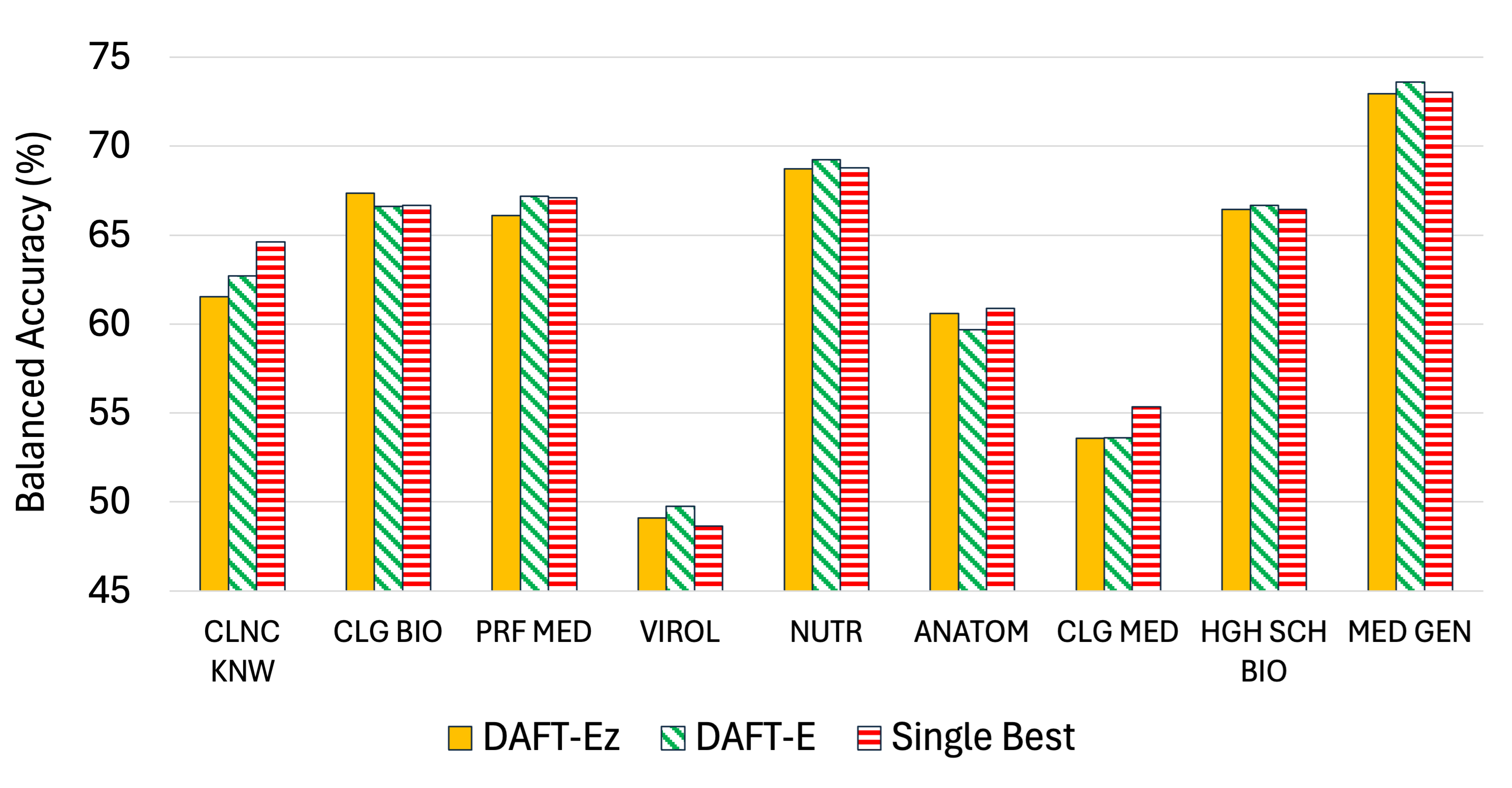}
\caption{Performance of \daftez, \dafte and Single Best \daft for Medical Knowledge task with MMLU datasets. 
% found post-hoc (see \eqref{eq:daft-ri}). 
}
% As in Figure~\ref{fig:heatmap}, the results show that the performance of \daft models can vary significantly.
\vspace{-10pt}
\label{fig:daft_e_MMLU}
\end{wrapfigure}

%As mentioned earlier, the motivation behind using ensemble models is the unavailability of large amounts of training data from the target domain, constraints on computational power. However, 

For benchmarking purposes, let us assume that we have the entire training data ($D_T$) of the target domain, and have a model that is fine-tuned on $D_T$ (\emph{FFT} on $D_T$). Let us compare the performance of the ensemble methods with this FFT on different datasets. Fig. \ref{fig:Ensmbl_FFT} and \ref{fig:Ensmbl_FFT_text} show the performance comparison of \daftez (green dash-dot line), \dafte (purple dash-dot line), single best \daft (black dotted line), and FFT on $D_T$ (red dotted line) for all six datasets of the sentiment analysis and for all three datasets of the text similarity tasks respectively.
% \todo[color=blue!10]{The above line can go in the caption of figure 5 to save space}
% The result for text similarity tasks is provided in the Appendix (Fig.).
%In the figure, the error intervals show the minimum and maximum performance (accuracy) with \dafte when the ensemble weights are calculated using different random data samples from the target dataset (with different random seeds). The error intervals are usually within a range of less than $2\%$ when weights are tuned by using $n \geq 8$. Furthermore, the standard deviation of the performance is much smaller compared to the error interval.
From  Fig.~\ref{fig:Ensmbl_FFT} and \ref{fig:Ensmbl_FFT_text}, we observe the advantage of using \dafte over \daftez when training data is available. It is interesting to note that \dafte has a strong increasing trend in performance (w.r.t. fine-tuning data size) for most cases, and catches up with either the single best \daft or FFT for the target domain with the increase of fine-tuning data. The performance of \dafte compared to the best solution, i.e., FFT on the target domain, can be theoretically bound and is given by Proposition \ref{prop:dafte-vs-opt} in Section \ref{sec:theory}.

\begin{wraptable}{r}{0.48\textwidth}
\footnotesize
\vspace{-12pt}
    \caption{Computational cost of methods shown in Fig.~\ref{fig:paper_structure}. $n$ denotes the number of few-shot samples. $C_F$ and $C_B$ denote the computational costs of a forward and backward pass for a LLM. $E$ is the number of fine-tuning epochs. $N$ is the number of \daft models available to ensemble, $\bar N \leq N$ is the number of nonzero weights in the weighted ensemble.}
    \vspace{-8pt}
    \label{tab:cc}
    \centering
    \begin{tabular}{lccc}
    \toprule
    Method & Zero-shot & Training cost & Inference cost \\
    \midrule
    \ft     & \xmark & $n(C_F+C_B)E$ & $C_F$              \\
    \midrule
    \daftz  & \cmark & $0$       & $C_F$              \\
    \daftez & \cmark & $0$       & $N \cdot C_F$      \\
    \midrule
    \daftsq & \xmark & $n(C_F+C_B)E$ & $C_F$              \\
    \dafte  & \xmark & $Nn(C_F + E)$  & $\bar N \cdot C_F$ \\
    \bottomrule
    \end{tabular}
    % \vspace{-10pt}
\end{wraptable}

Table \ref{tab:cc} shows the overall comparison of the computational cost of all the five options that we have discussed in this paper. It should be noted that none of the ensemble methods suffer from the high computational complexity of back-propagation computation; further, using sparse weighting we can reduce the inference cost in \dafte compared to \daftez. 
It is straightforward to see that \daftz and \daftez are the only ones among the five options discussed here that are zero-shot, and hence incurs zero training cost. In contrast, \ft and \daftsq has fine-tuning cost, i.e., back-propagation cost ($C_B$), along with forward propagation cost ($C_F$). For \dafte, the training cost is the forward propagation cost of $N$ \daft models along with the linear cost for learning the weight of the ensemble layer. Lastly, \ft, \daftz and \daftsq use only a single (\daft) model, and hence incurs only a forward propagation cost ($C_F$), whereas \daftez and \dafte use $N$ and $\Tilde{N}$ \daft models respectively in the inference stage.

To conclude the overall comparison, we present the performance of \daftez, \dafte, and the single best \daft model for the Medical Knowledge task using the MMLU datasets in Fig.~\ref{fig:daft_e_MMLU}. For \dafte, each dataset was split in half: one half was used to tune the linear weights of \dafte, and the other half was used for performance evaluation\footnote{\dafte performance reported here uses the same set of hyper-parameters for ensemble-weight learning across all datasets; dataset-specific tuning could further improve results.}. Since \daftez and \daft models required no tuning, only the second half of the data was used for evaluation. This ensured that all three methods were evaluated on the same data. Due to the small size of each dataset, we repeated the random split 200 times for each of the 9 datasets and report the average performance. From the figure, we observe that \dafte outperformed \daftez in 7 out of the 9 tasks. Most notably, \dafte even outperformed the single best \daft model in 5 of the 9 cases, demonstrating the effectiveness of the \dafte approach.

\begin{conc}
\daftez and \dafte are strong lightweight methods for leveraging already available \daft models, often matching the single best \daft model (only known post-hoc), and at times match the best possible performance of full fine-tuning with a large fine-tuning set.
\end{conc}

% \begin{wrapfigure}{r}{0.5\textwidth}
% \vspace{-30pt}
%     \centering
%     \includegraphics[width=\linewidth]{figures/Heatmap_LEEP.png}
%     \caption{Heatmap using LEEP scores (Sentiment Analysis)}
%     \label{fig:LEEP_score_heatmap}
%     \vspace{-10pt}
% \end{wrapfigure}

\subsection{Comparison with Model Soup}
We compare the performance of \daftez and \dafte to a state-of-the-art zero shot ensemble model, Uniform Soup (Model Soup)~\citep{wortsman2022model} 
% We provide a short comparison between model soup and the ensemble methods discussed here (\daftez and \dafte) 
% in terms of their performances and features
\footnote{\iia{Methods like mixture of experts (MoE) need training with data and our motivation for ensemble was to avoid any training that needs expensive back-propagation. Hence, we refrained from comparing with techniques that require back-propagation unless it is fine-tuning with in-domain data, i.e., \ft and \daftsq.}}. To do performance comparison between the zero-shot version of model soup, i.e., Uniform soup with \daftez, we used the sentiment analysis task. We observed that when there are 5 DAFT models from some specific architecture (we could not use all 15, since there were 3 different architectures)\footnote{\iia{Model Soup only works if all the (\daft) models are of the same architecture, which is not the case for \daftez (or \dafte). Furthermore, Model Soup needs all the weights of the models to work, but for ensembling, we do not need that. Just API access to the \daft models is adequate.}}, \textsf{DAFT-E}$^\text{Z}$  performs better than Model Soup (Table \ref{tab:daft-Ez_Uniform-Soup}). To check the robustness of the \daftez method, we then changed our experiments to check performance of \daftez and Uniform soup on IMDB dataset by adding \daft models to the respective methods. The result is shown in Table \ref{tab:robustness}. Interestingly, \daftez showed more resiliency than Uniform soup. 

\begin{table}[t]
\begin{minipage}{0.45\textwidth}
\footnotesize
    \centering
    \caption{Performance Comparison of \textsf{DAFT-E}$^\text{Z}$ and Uniform Soup. The results show that, for zero-shot sentiment analysis tasks, \daftez performs at par or better than Uniform Soup. Note that Uniform Soup requires all models to have matching architectures so as to be able to uniformly combine the model weights. \daftez (and \dafte) do not have any such requirement.}
    \label{tab:daft-Ez_Uniform-Soup}
   \begin{tabular}{lcc}
    \toprule
    Dataset & Uniform Soup & \daftez  \\
    \midrule
    Amazon  &     93.29\% &  {\bf 93.72}\% \\
    Cornell &     85.09\% &  {\bf 85.37}\%  \\
    IMDB    &     92.09\% &  {\bf 92.28}\%  \\
    SST2    &     88.07\%&  {\bf 89.55}\%  \\
    Tweet   &     86.09\%&  {\bf 87.33}\%   \\
    Yelp    &     94.67\%&  {\bf 95.30}\%  \\
    \bottomrule
    \end{tabular}
    \end{minipage}
% \end{table}
\hfill
% \begin{table}[htb]
\begin{minipage}{0.52\textwidth}
\centering
\caption{Robustness Comparison of \daftez and Uniform Soup in terms of the set of models being combined. The results show that as we increase the number of \daft models, the performance of both Uniform Soup and \daftez improves, with \daftez consistently outperforming Uniform Soup in the zero-shot setting. As discussed in Table~\ref{tab:daft-Ez_Uniform-Soup}, we are able to add many more models into \daftez than in Uniform Soup because Uniform Soup requires all models in the ensemble to have matching architectures.}
\label{tab:robustness}
{\scriptsize\begin{tabular}{lcc}
\toprule
\daft models combined & Uniform Soup &  \daftez \\
\midrule
SST2, Tweet & 88.78\% & {\bf 89.09}\% \\
SST2, Tweet, Yelp & 90.57\% & {\bf 90.86}\%\\
SST2, Tweet, Yelp, Cornell & 91.19\% & {\bf 91.39}\%\\
SST2, Tweet, Yelp, Cornell, Amazon & 92.09\% & {\bf 92.28}\%\\
\bottomrule
\end{tabular}}
\end{minipage}
\end{table}

\section{Theoretical Analysis}
\label{sec:theory}
% Here in this section we theoretically claim a few propositions to relate the performance 
%of fine-tuning \daft (\ft and \daftsq), 
% of the ensemble methods (\daftez and \dafte) to the individual (\daft and FFT) models. First we provide few new notations that will be utilized in expressing the propositions and the discussion that follows.

\noindent\textbf{Notations.} The dataset of the target domain is $D_T$, the input data to the model is $\mathbf{x}$ and the %actual 
output data is $\mathbf{y}$, which the model should predict given the input data. There are $N$ number of \daft models, and a \daft model can be built using any base model $B_j \in \mathcal{B}$, with $j = \{1, 2,... , J\}$ and fine-tuning it on dataset $D_k \in \mathcal{DA}$ with $k = \{1, 2, ... , K\}$, where $\mathcal{DA}$ is the set of domain adjacent datasets. 
The $i^{th}$ \daft model is given by $M_i$ and the corresponding base model and fine-tuning dataset are given by $B_{\kappa(i)}$ and $D_{\nu(i)}$, respectively, i.e., $B_{\kappa(i)}$ was fine-tuned on $D_{\nu(i)}$ dataset to get $M_i$. Here, $\kappa(i)$ and $\nu(i)$ are index mapping functions. 
Also, if we assume that all these \daft models were created by fine-tuning fully on the given datasets (no fractional fine-tuning), then the total number of \daft can not be greater than $J \times K$, or $N \leq JK$. 
% Let $\mathcal{DA}$ be the set of domain adjacent data of the target dataset $D_T$. 
Lastly, let $\ell(M(\mathbf{x}),\mathbf{y})$ be the loss in output when model $M$ is used on target dataset input $\mathbf{x}$ and the output is compared with $\mathbf{y}$. Since $\mathbf{x}$ and $\mathbf{y}$ are common for all loss calculation,
%when comparing different models, 
we %will mostly 
% omit those variables from the loss function for conciseness and 
can just use $\ell(M)$ to represent $\ell(M(\mathbf{x}),\mathbf{y})$.

\subsection{\daftez versus \daftz}
\label{subsec:daftez_daft}

Determining if a \daft model is going to perform well or not is very hard to know a priori if adequate time or data to train or test the model is not available. Hence, instead of choosing one of the \daft models in random (\daftz), one solution could be to use the ensemble of the output results (\daftez) from all the \daft models.

\begin{proposition}
\label{prop:daftez}
    The performance of the average ensemble of \daft models (\daftez) is no worse than the expected performance obtained from choosing the \daft models uniformly at random.
\end{proposition}

Proposition~\ref{prop:daftez} states that in terms of expected performance, using \daftez is no worse (strictly better, if $\ell$ is assumed to be strictly convex) than picking from the \daft models uniformly at random. 
From the performance of \daftez shown in Fig. \ref{fig:RI_Ens_DAMs} and \ref{fig:RI_Ens_text}, we see that Proposition \ref{prop:daftez} holds, i.e., \daftez has better performance than the average performance of the \daft models. Moreover, for these two specific types of tasks, the ensemble of the \daft models beats the performance of individual \daft models in most cases (98 out of 108 in total for both tasks).

% Proposition~\ref{prop:daftez} states that in terms of expected performance, using \daftez is no worse (strictly better, if $\ell$ is assumed to be strictly convex) than picking from the \daft models uniformly at random.

\subsection{\dafte versus Optimum}
\label{subsec:dafte_opt}
% The performance of weighted ensemble depends on the base models on which the \daft models were trained on and also how close the domain adjacent datasets are to the target domain. 
Let us denote $\tilde{M}_i$ as the model that uses the base model $B_{\kappa(i)}$ and is fine-tuned on $D_T$, i.e., $\tilde{M}_i = \Phi(B_{\kappa(i)}, D_T)$. Also, we denote $\tilde{M_*}$ is the best performing model when fine-tuned on $D_T$ (the base model for $\tilde{M_*}$ can be from $\mathcal{B}$ that are used to generate \daft models or some other large model). The following proposition bounds the difference of loss between the optimum solution and \dafte. 
\begin{proposition}
\label{prop:dafte-vs-opt}
    The loss of \dafte is bounded as:
    {%\footnotesize
    \begin{align}
        \ell (\text{\dafte}) \leq & \ell(\tilde{M_*}) 
     + \min_{i \in \mathcal{N}} \left[ \mu (\tilde{M}_i, \tilde{M_*}) + \rho( D_{\nu(i)}, D_T) \right],
        \label{prop:dafte-vs-opt-eqn}
    \end{align}}
where $\mu (\tilde{M}_i, \tilde{M_*})$ is the performance difference between the models $\tilde{M}_i$ and $\tilde{M_*}$ both fine-tuned on $D_T$, and $\rho( D_{\nu(i)}, D_T) $\footnote{Some practical dataset distance measurement metric are the Wasserstein distance metric \citep{panaretos2019statistical}, Jensen-Shannon Divergence (JSD), etc.} is an appropriately defined distance measure between $D_{\nu(i)}$ 
%(on which the model $\tilde{M}_i$, a \daft, was fine-tuned on) 
and $D_T$.
\end{proposition}

% \todo[inline]{The notation is leading to confusion here. There are the \daft $M_i$ and then there are the $\tilde M_i$. I might be missing something but how are they different from just \ft or \daftsq.}

%From  Equation \ref{eq:proof3_4} we see that the equation is true for any $i$, and hence if we can ensure that we have chosen a \daft for which $\mu(\tilde{M}_i ,\tilde{M}_*)$ is very small, then the bound becomes dependent only on the $\rho$ function. 
Usually base models (derived from FMs which are all quite large) perform similarly when fine-tuned on the full target dataset. Under that assumption, we can ignore the $\mu()$ term in the bound,
resulting in the following corollary.

\begin{corollary}
\label{cor:dafte-vs-opt}
    The loss of \dafte is larger than the optimum loss by no greater than $min_i \rho(D_{\nu(i)},D_T)$, with the assumption that the base models of the {\daft}s can perform as well as the optimum when fine-tuned on the target domain. 
\end{corollary}

%\qed

Proposition~\ref{prop:dafte-vs-opt} and Corollary~\ref{cor:dafte-vs-opt} 
% (which follows directly from Proposition~\ref{prop:dafte-vs-opt}) 
imply that the performance of \dafte depends on the base model on which the \daft models were trained and the datasets they were fine-tuned on. If the base models (FMs) in the ensemble are nearly as good as the best possible base model (FM) (when the performance of the \emph{FFT} of the base models are compared), then we can ignore the $\mu()$ term in Equation \ref{prop:dafte-vs-opt-eqn}, and the performance of \dafte depends \textit{only on the dataset that is closest} to the target domain. 
% If any one of the datasets matches with the target domain, \dafte will perform the same as the the best \emph{FFT} model fine-tuned on target data, i.e., achieving the best possible performance. 
If the \daft models are generated from a large number of adjacent datasets from diverse domains, we expect the distance from the target dataset to the closest \daft dataset to be small, resulting in \dafte performing very close to the optimum.

In our experimental setting, the assumption made for the Corollary \ref{cor:dafte-vs-opt} holds, that is, all three base models are comparable in terms of performance when fine-tuned on the same dataset. Hence, the performance of \dafte depends on the adjacency of the datasets used in the \daft models with the target dataset. For the case of sentiment analysis, we have six datasets, whereas for the text similarity task, we only have three. Thus, it is more probable that the \daft models of the text similarity task are more diverse and \dafte will be able to achieve close to optimum performance. From the results (Fig. \ref{fig:Ensmbl_FFT}, and \ref{fig:Ensmbl_FFT_text}) it is evident that \dafte was able to perform much better for the sentiment analysis task, compared to the textual similarity task.

% \begin{figure}
%     \centering
%     \includegraphics[width=\linewidth]{figures/Proof_1.png}
%     % \caption{proof}
%     \label{fig:proof_1}
% \end{figure}
% \begin{figure}
%     \centering
%     \includegraphics[width=\linewidth]{figures/Proof_2.png}
%     % \caption{proof}
%     \label{fig:proof_2}
% \end{figure}
% \begin{figure}
%     \centering
%     \includegraphics[width=\linewidth]{figures/Proof_3.png}
%     \caption{proof 1 :  Relating FT and DAFT$^2$}
%     \label{fig:proof_3}
% \end{figure}

% \begin{figure}
%     \centering
%     \includegraphics[width=\linewidth]{figures/Proof_4.png}
%     \caption{proof 2 : Relating DAFT$^z$ and DAFT-E$^z$}
%     \label{fig:proof_4}
% \end{figure}

% \begin{figure}
%     \centering
%     \includegraphics[width=\linewidth]{figures/Proof_5.png}
%     % \caption{proof}
%     \label{fig:proof_5}
% \end{figure}
% \begin{figure}
%     \centering
%     \includegraphics[width=\linewidth]{figures/Proof_6.png}
%     % \caption{proof}
%     \label{fig:proof_6}
% \end{figure}
% \begin{figure}
%     \centering
%     \includegraphics[width=\linewidth]{figures/Proof_7.png}
%     % \caption{proof}
%     \label{fig:proof_7}
% \end{figure}
% \begin{figure}
%     \centering
%     \includegraphics[width=\linewidth]{figures/Proof_8.png}
%     % \caption{proof}
%     \label{fig:proof_8}
% \end{figure}

% \begin{figure}
%     \centering
%     \includegraphics[width=\linewidth]{figures/Proof_9.png}
%     \caption{proof 3 : Relating DAFT-E and FFT}
%     \label{fig:proof_9}
% \end{figure}

\vspace{-5pt}
\section{Conclusion}
\label{sec:conclusion}
\iia{In this paper, we explore 4 ways to leverage the abundance of publicly available fine-tuned LLMs for data-scarce tasks.}
% \todo[color=blue!10]{Above line $\to$``In this paper, we explore 4 ways to leverage the abundance of publicly available fine-tuned LLMs for data-scarce tasks.''}
We investigate how the \daft models can be utilized for inference under limitations on computation time or data required for training a model on a target task. The \daft models can be used on the target domain (zero-shot) without any fine-tuning. If chosen properly, the performance of \daft models on the target domain can be close to the optimal performance. However, this performance depends strongly on the choice of the \daft model, 
and ensemble methods can be used to address this issue. Two ensemble methods, \daftez and \dafte are explored; their performances are empirically evaluated and compared against individual \daft models and other benchmarks involving larger LLMs and base models trained with the full target domain dataset. Our theoretical results support the conclusions from the empirical findings, and provide insights under what conditions \dafte can provide near-optimal performance.

\subsubsection*{Broader Impact Statement}
The use of \daft models can be a great low-resource and easy-to-use solution in data-scarce domains. Due to the current surge of open source \daft models for different downstream tasks, 
the community can use these models directly without any (computationally expensive) fine-tuning of LLMs. 
The main innovation of the proposed ensemble methods, i.e., \daftez and \dafte, is the utilization of readily available \daft models to perform lightweight solutions for problems where little or no data from the target domain is available.
% , and implement a  with {\daft}s. 
% There can be many solutions for data-scarce domains, but we  we are studying something that has not been studies, simple to implement and lightweight.

On the other hand, there are certain limitations of our method that the users may face and should keep in mind. 
% affect the performance of the proposed methods. 
% our proposed methods can have. 
Firstly, 
the performance of these methods is highly dependent on the availability of DAFT models. \dafte or \daftez will not perform well when none of the available \daft models are adjacent to the target domain.
% In scenarios where scarcity also applies to models, the method may not perform as well as showed in this paper.It is true, and the issue will be solved if the current surge of DAFT models continues. 
Moreover, while many \daft models for sentiment analysis and NLI tasks are currently available, {\daft}s for some other tasks might not be as widely available soon.
% Our goal is to make general observations on the the overall applicability of such DAFT models, rather than focusing on their utility at this very moment. Even if the applicability of our method in not high at the moment, it is likely to increase as more and more DAFT models become publicly available. 
Secondly, \daftez and \dafte require inference with multiple \daft models. While the multi-model inference can be parallelized, thereby reducing inference time in scenarios where multiple smaller servers are available, inference time would increase if memory is limited or the per-model inference needs to be sequential. Further, with DAFT-E, we can filter out some of the candidate DAFT models, thereby reducing $N$, but that depends on having some data from the target domain. 
% While the complexity of our method, especially, DAFT-E$^Z$ is proportional to $N$, note that the inference execution for the $N$ candidate  models can be parallelized, thereby reducing inference time in scenarios where multiple smaller servers are available. 
Thirdly, 
with the proposed methods, the users are supposed to use openly available fine-tuned models; however, the quality or correctness of the fine-tuning datasets and training methods may be questionable. There is also the question of the privacy of the data and the use of the models, which the user might not be aware of.
Lastly, the risk of bias and misuse of using the DAFT models is possible. However, in this work, we are not proposing new models, just some methodologies to use already available models through public repositories. Since we are not changing the DAFT models, if the fine-tuning of the DAFT models were done according to safety and bias alignment, we expect to have minimal risk of bias and misuse from using ensembles of these
models.

% While the complexity of our method, especially, DAFT-E$^Z$ is proportional to $N$, note that the inference execution for the $N$ candidate  models can be parallelized, . Further, with DAFT-E, given a small amount of data from the target domain, we can filter out some of the candidate DAFT models, thus reducing $N$. In our evaluations, we found that DAFT-E puts much more weight on the top 6-7 DAFT models among the 15. One could also envision using LLMs to perform the selection of the $N$ DAFT models from the candidate models, without the need to use target domain data, i.e., using just the description of the task. However, that will be explored in future work.

% In this optional section, TMLR encourages authors to discuss possible repercussions of their work,
% notably any potential negative impact that a user of this research should be aware of. 
% Authors should consult the TMLR Ethics Guidelines available on the TMLR website
% for guidance on how to approach this subject.

% \subsubsection*{Author Contributions}
% If you'd like to, you may include a section for author contributions as is done
% in many journals. This is optional and at the discretion of the authors. Only add
% this information once your submission is accepted and deanonymized. 

\subsubsection*{Acknowledgments}
The work was supported by RPI-IBM Future of Computing Research Collaboration (https://fcrc.rpi.edu).
% Use unnumbered third level headings for the acknowledgments. All
% acknowledgments, including those to funding agencies, go at the end of the paper.
% Only add this information once your submission is accepted and deanonymized. 

\bibliography{main}
\bibliographystyle{tmlr}

% \clearpage
% \pagebreak 

\appendix

\appendix

\section{Appendix}
\label{sec:appendix}

\subsection{Datasets and Models}
\label{appndx_subsec_dataset}

The attributes of the six datasets used for sentiment analysis and the three dataset used for text similarity analysis are given in Table \ref{tab:dataset_info1} and \ref{tab:dataset_info2}.
For `SST2' we had to use validation dataset as test data, because the test data does not have any labels. The direct link to download these datasets are given as follows: https:// huggingface.co /datasets/amazon\_polarity, `https:// www.cs.cornell.edu /people/pabo/movie-review-data/', `https://huggingface.co/datasets/ imdb', `https:// huggingface.co/ datasets/sst2', `https:// huggingface.co/ datasets/ mteb/tweet\_sentiment\_extraction', `https:// huggingface.co/ datasets/ yelp\_polarity', 'https://huggingface.co /datasets/nyu-mll/glue' (when copying the links please remove any spaces from the texts). Figs. \ref{fig:heatmap_v2_bert} and \ref{fig:heatmap_v2_xlnet}
 shows the performance heat-map for sentiment analysis task when the base models are BERT-based-uncased and xlnet-base-cased respectively. The heat maps in general follows a similar pattern as observed in Fig. \ref{fig:heatmap}, i.e., Tweet test data cannot be predicted well by any other \daft models, and Cornell test data is also hard to predict. The observation of \daft trained on tweet data performing poor is true for bert-based-uncased as well, but not always true for xlnet-base-cased. Another interesting observation is that for xlnet, the \daft model trained on IMDB data seems to perform quite poorly in a few instances (i.e., on test data of Cornell and Tweet).

\begin{figure}
    \centering
    \begin{minipage}{0.49\textwidth}
    \includegraphics[width=0.9\linewidth]{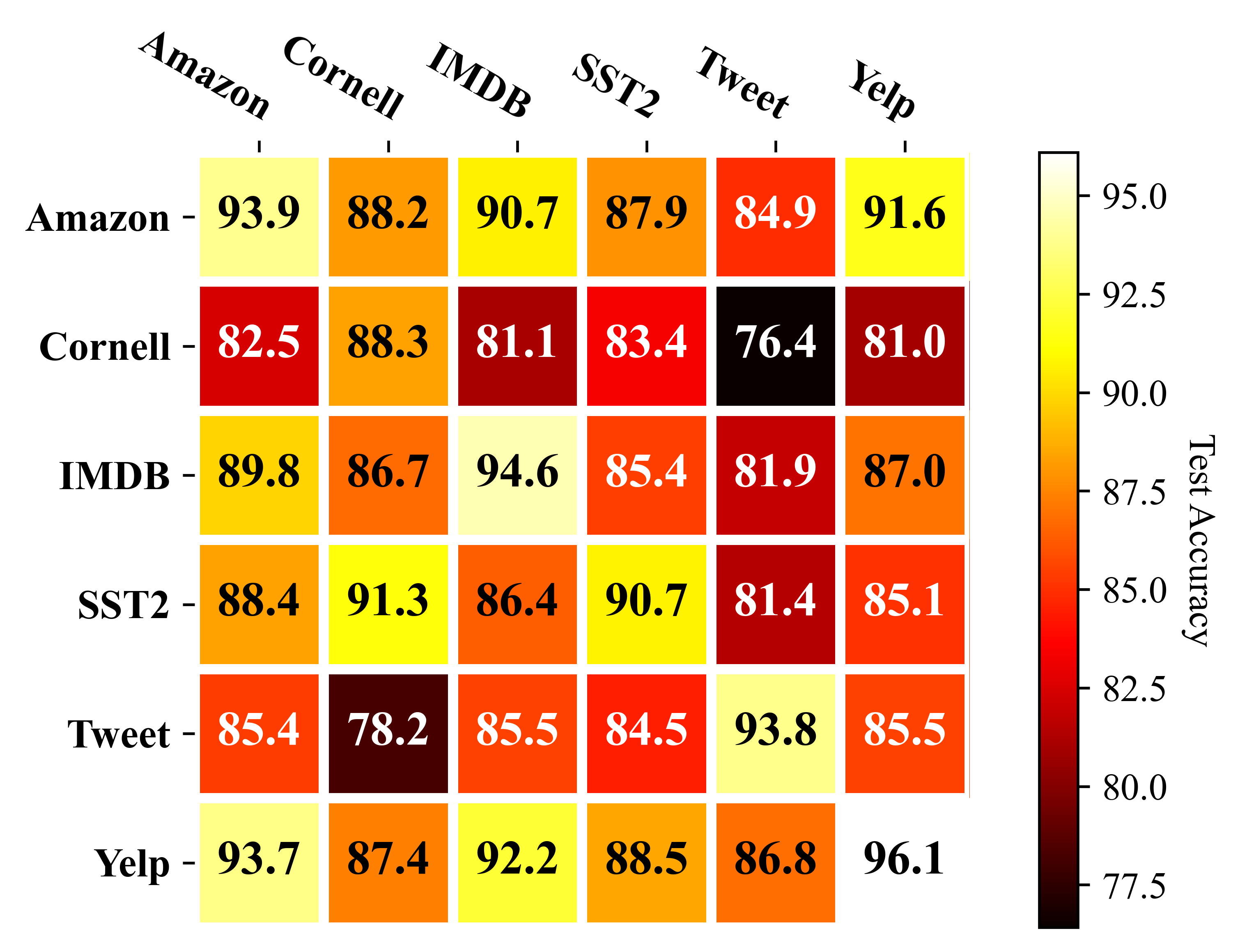}
    \caption{Heatmap showing the performance of \daft models for sentiment analysis task when base model is BERT-based-uncased}
    \label{fig:heatmap_v2_bert}
    \end{minipage}
% \end{figure}
% \begin{figure}
\begin{minipage}{0.49\textwidth}
    \centering
    \includegraphics[width=0.9\linewidth]{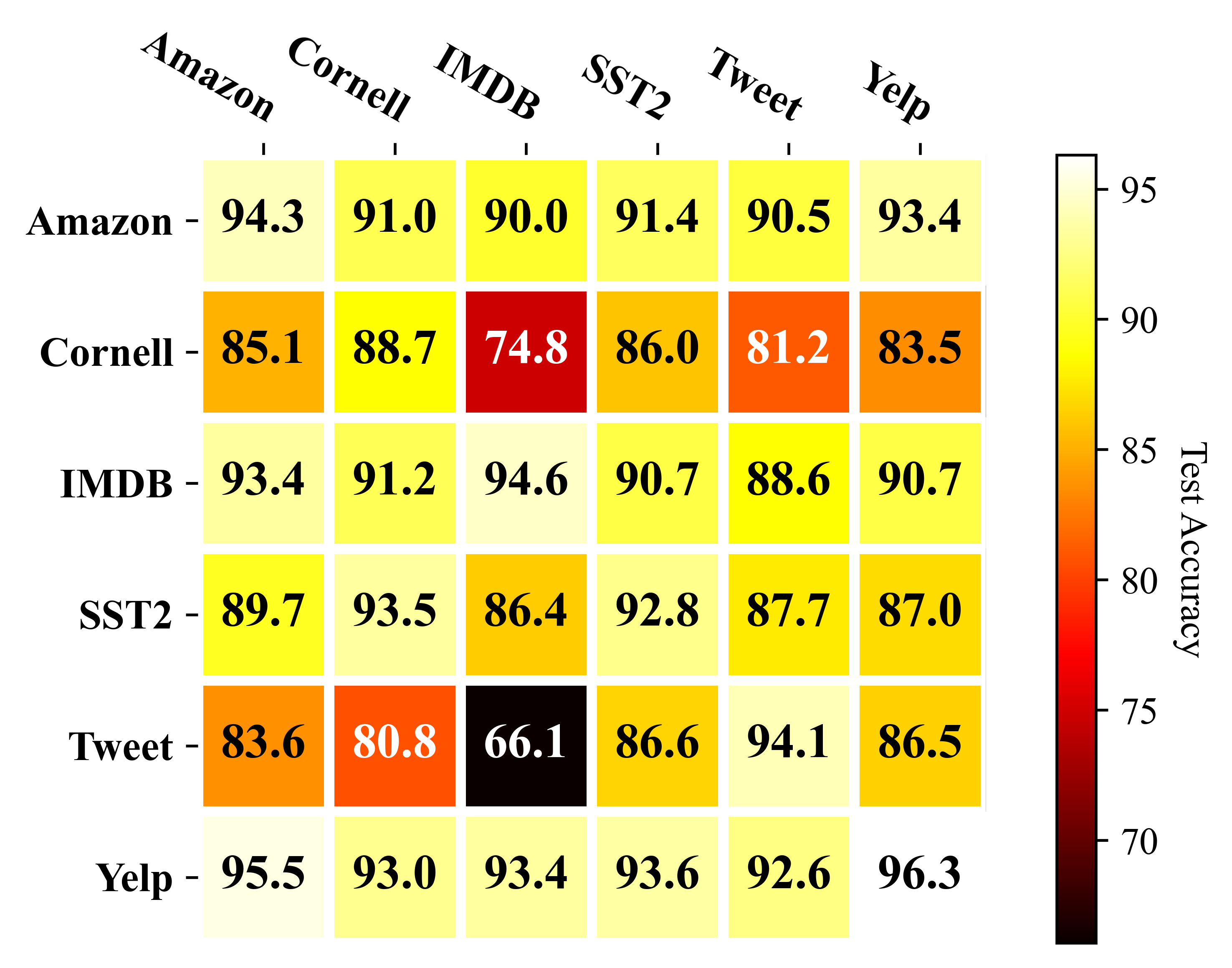}
    \caption{Heatmap showing the performance of \daft models for sentiment analysis task when base model is xlnet-base-cased}
    \label{fig:heatmap_v2_xlnet}
\end{minipage}
\end{figure}

\subsection{Experimental Settings}
The base models were used from huggingface using the 'AutoTokenizer.from\_pretrained', and the webpage with descriptions and examples can be found in 'https://huggingface.co/ transformers/v3.0.2/model\_doc/auto.html'. To fine-tune and train these models we used Google Colab platform with the T4 GPU equipped machine. For \ft we chose each of the three base models (Roberta, BERT, xlnet) and fine-tuned the models with target data until the loss per epoch did not improve more than 1\% or at least a fixed number of data-samples has not been used for training. All these fine-tuned models on the target dataset acted as a suitable candidate for \daft models. For \daftsq we used any of these \daft models to fine-tune on another target dataset for few shot training. In the performance comparison of \daftsq we only showed the performance of \daftsq that was fine-tuned on \daft models having Roberta as its base. For both \ft and \daftsq on few shot training, we performed the all the runs five times with five different seeds. For the case of \dafte, the weight calculations were done using five different random seeds as well. All the results shown here are the mean values of all those runs.

\subsection{Discussion on Performance Anomalies - SST2}
% \subsubsection{SST2 on Heatmap}
\label{subsec:sst2_ano}
The SST2 dataset obtained from Huggingface did not have any label on its test data, and we had to use the validation data of SST2 as the test data instead. Also, the validation data of SST2 only had $872$ samples. Now, 
in the heat map shown in Fig. \ref{fig:heatmap}, the performance of \daft-SST2 achieved $92\%$ accuracy on the SST2 validation data, whereas, \daft-Cornell achieved $93.1\%$ accuracy on that same test dataset. This is quite counter intuitive, because we expect \daft -SST2 to be the best performing one on its own domain (SST2 validation data). We did a thorough simulation analysis to check what might be the reason. Since the \daft -SST2 was generated using randomly chosen $7,000$ samples from the training dataset of SST2, and we initially extracted the rest of the training data and divided it to different batches of $5,000$ samples. Now, with this batches we checked the performance of \daft - SST2 and \daft- Cornell. As expected, we got better performance with \daft - SST2 compared to \daft - Cornell with these new batches. With this experiment, we concluded that SST2 validation dataset is, for some unknown reason, more similar to the Cornell training dataset than the SST2 training data, and thus giving a counter intuitive result.

% \subsubsection{STSB with DAFT-E}
% \label{subsec:stsb_ano}
% The performance of DAFT-E with $LR$ suffers a lot compared to both \daftsq that have been considered here, i.e., \daft-mrpc and \daft-qqp. Both of these \daft models have roberta as their base model, and interestingly these roberta based \daft models seemed to perform well even at zero shot on STSB dataset. Whereas, the \daft models based on BERT and xlnet showed considerable less performance. We found this to be the main reason for which \dafte performed poorly in comparison to the \daftsq models which already had better performance even without any fine-tuning. To check this, we even checked the \dafte performance with \daft-mrpc and \daft-qqp with roberta base only, and found that \dafte outperformed the \daftsq up to $n=64$.

\subsection{Fine-tuning the header or whole model}
Figs. \ref{fig:header_vs_FT1} to \ref{fig:header_vs_FT6} show the performance comparison of fine-tuning only the header layer of the base model and fine-tuning the whole model of the base models. It is straightforward to see the superior performance of fine-tuning the whole model even though the base models (i.e., BERT, Roberta) have already gone through extensive training with language data.

\begin{table}
\begin{minipage}{0.53\textwidth}
\centering
  \caption{Dataset sizes and classes (Sentiment Analysis)}
  \label{tab:dataset_info1}
  \begin{tabular}{lccc}
    \toprule
    Dataset  &  Train data & Test data & Classes\\ 
    \midrule
        Amazon  & 3,600,000 & 400,000 & 2 \\
       Cornell & 7,463 & 2134 & 2 \\
       IMDB & 25,000 & 25,000 & 2\\
       SST2 & 67,349 & 1821  & 2\\
       Tweet & 12927  & 3696 & 3\\
       Yelp & 560,000 & 38,000  & 2 \\
  \bottomrule
    \end{tabular}
    \end{minipage}
% \end{table}
% \begin{table}
\begin{minipage}{0.45\textwidth}
\centering
  \caption{Dataset sizes and classes (Text Similarity)}
  \label{tab:dataset_info2}
  \begin{tabular}{lccc}
    \toprule
    Dataset  &  Train data & Test data & Classes\\ 
    \midrule
        MRPC  & 3,670 & 1,730 & 2 \\
       QQP & 364,000 & 40,000 & 2 \\
       STSB & 7,500 & 1,500 & 2\\
  \bottomrule
  \\
  \\
    \end{tabular}
        \end{minipage}
\end{table}

\begin{figure*}[t]
\minipage{0.32\textwidth}
  \includegraphics[width=\linewidth]{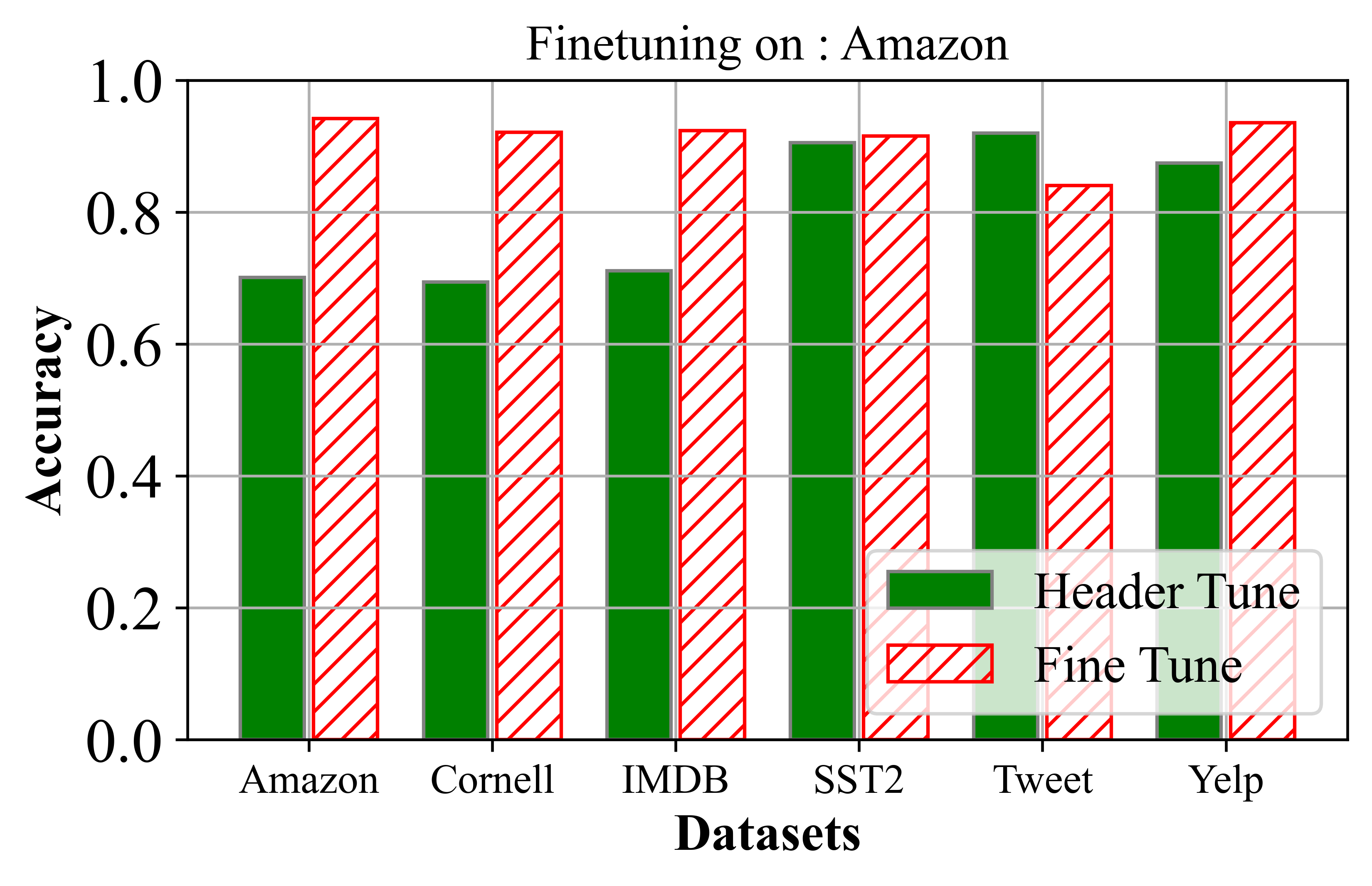}
\caption{Performance comparison of FT Header layer to FT whole Model - Amazon dataset.}
\label{fig:header_vs_FT1}
\endminipage
\hfill
\minipage{0.32\textwidth}
\includegraphics[width=\linewidth]{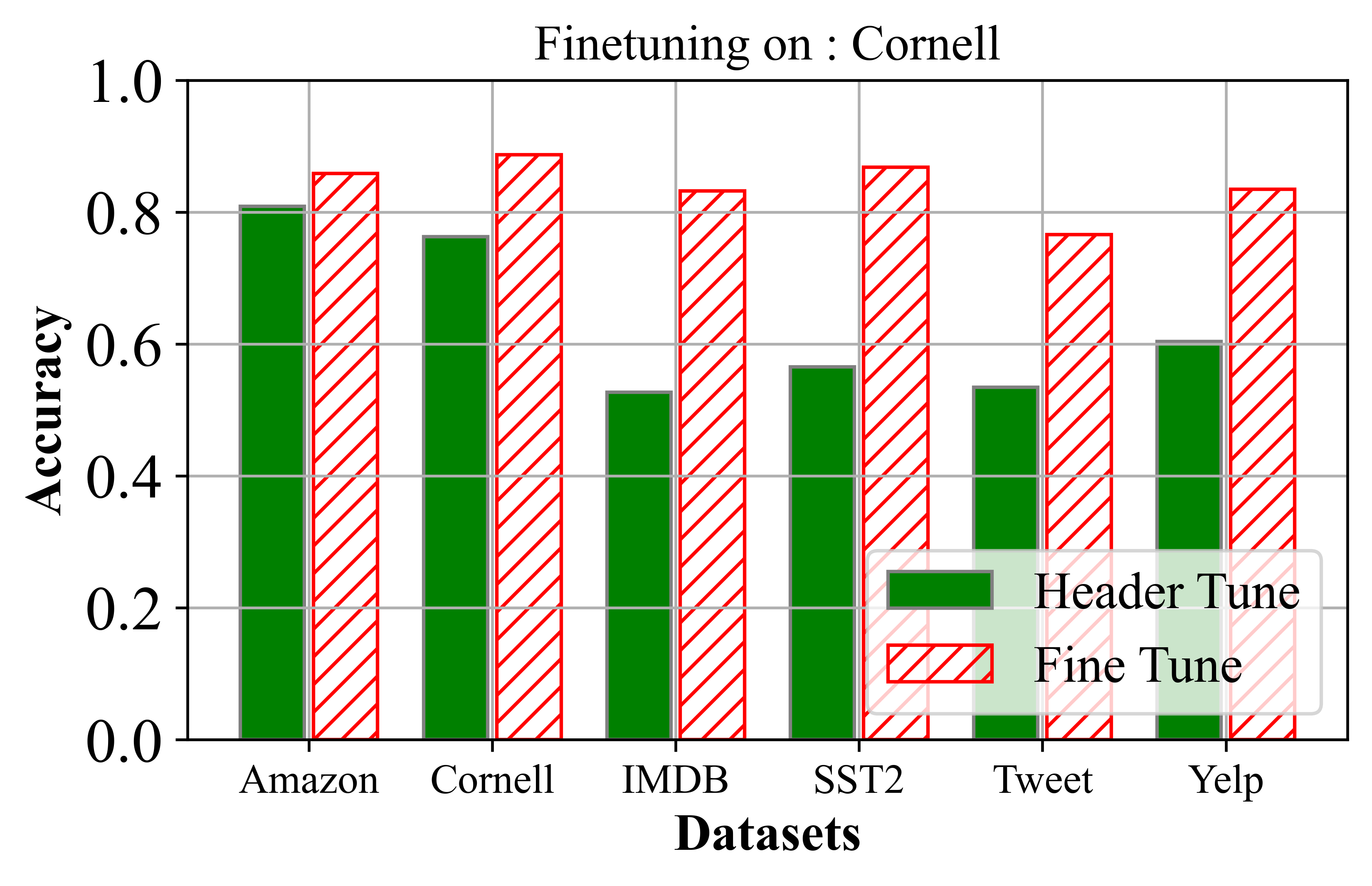}
  \caption{Performance comparison of FT Header layer to FT whole Model - Cornel dataset.}
  \label{fig:header_vs_FT2}
\endminipage
\hfill
\minipage{0.32\textwidth}
 \includegraphics[width=\linewidth]{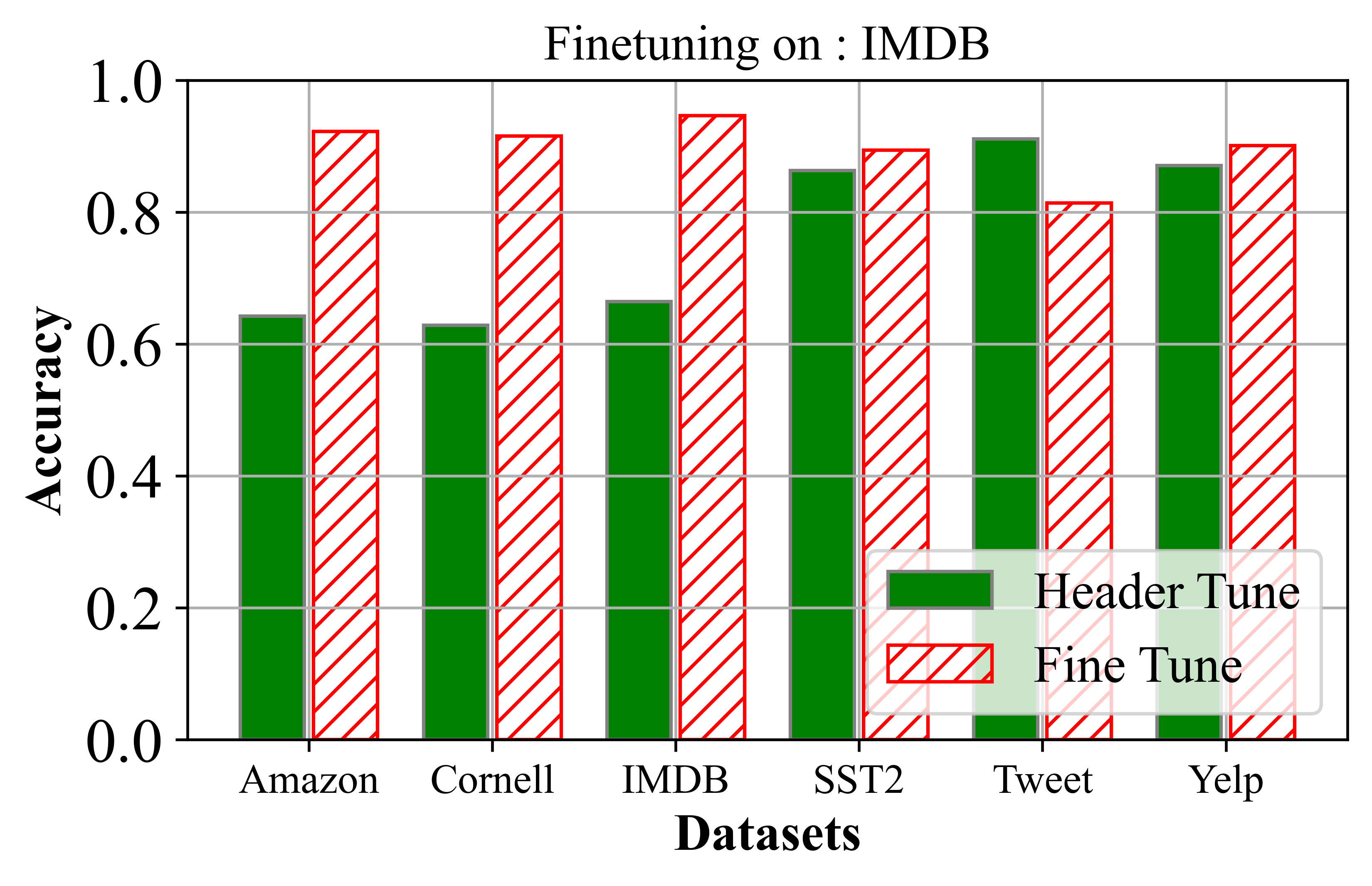}
  \caption{Performance comparison of FT Header layer to FT whole Model - IMDB dataset.}
  \label{fig:header_vs_FT3}
\endminipage
% \end{figure*}

% \begin{figure*}[t]
\minipage{0.32\textwidth}
  \includegraphics[width=\linewidth]{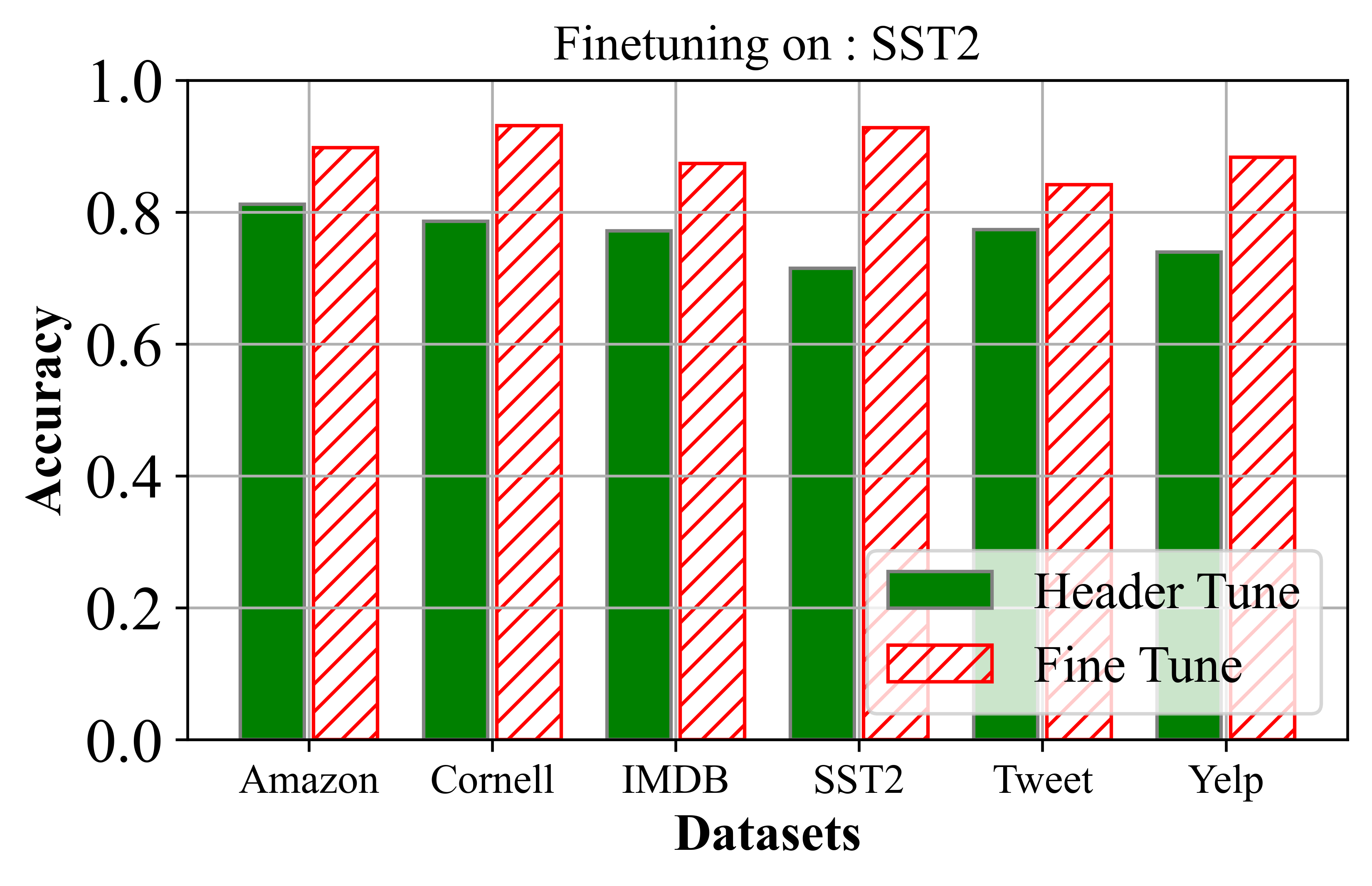}
\caption{Performance comparison of FT Header layer to FT whole Model - SST2 dataset.}
\label{fig:header_vs_FT4}
\endminipage
\hfill
\minipage{0.32\textwidth}
\includegraphics[width=\linewidth]{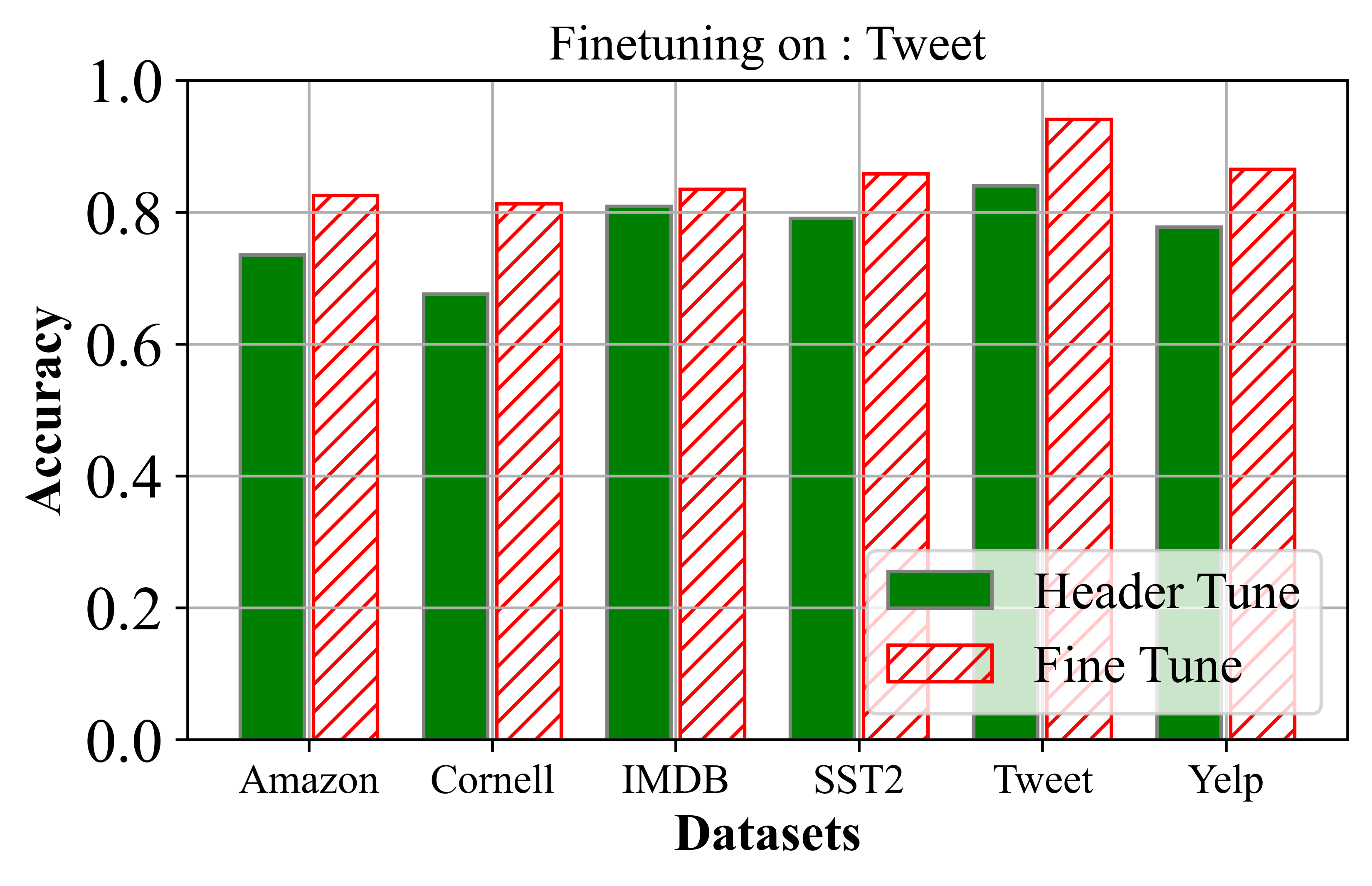}
  \caption{Performance comparison of FT Header layer to FT whole Model - Tweet dataset.}
  \label{fig:header_vs_FT5}
\endminipage
\hfill
\minipage{0.32\textwidth}
 \includegraphics[width=\linewidth]{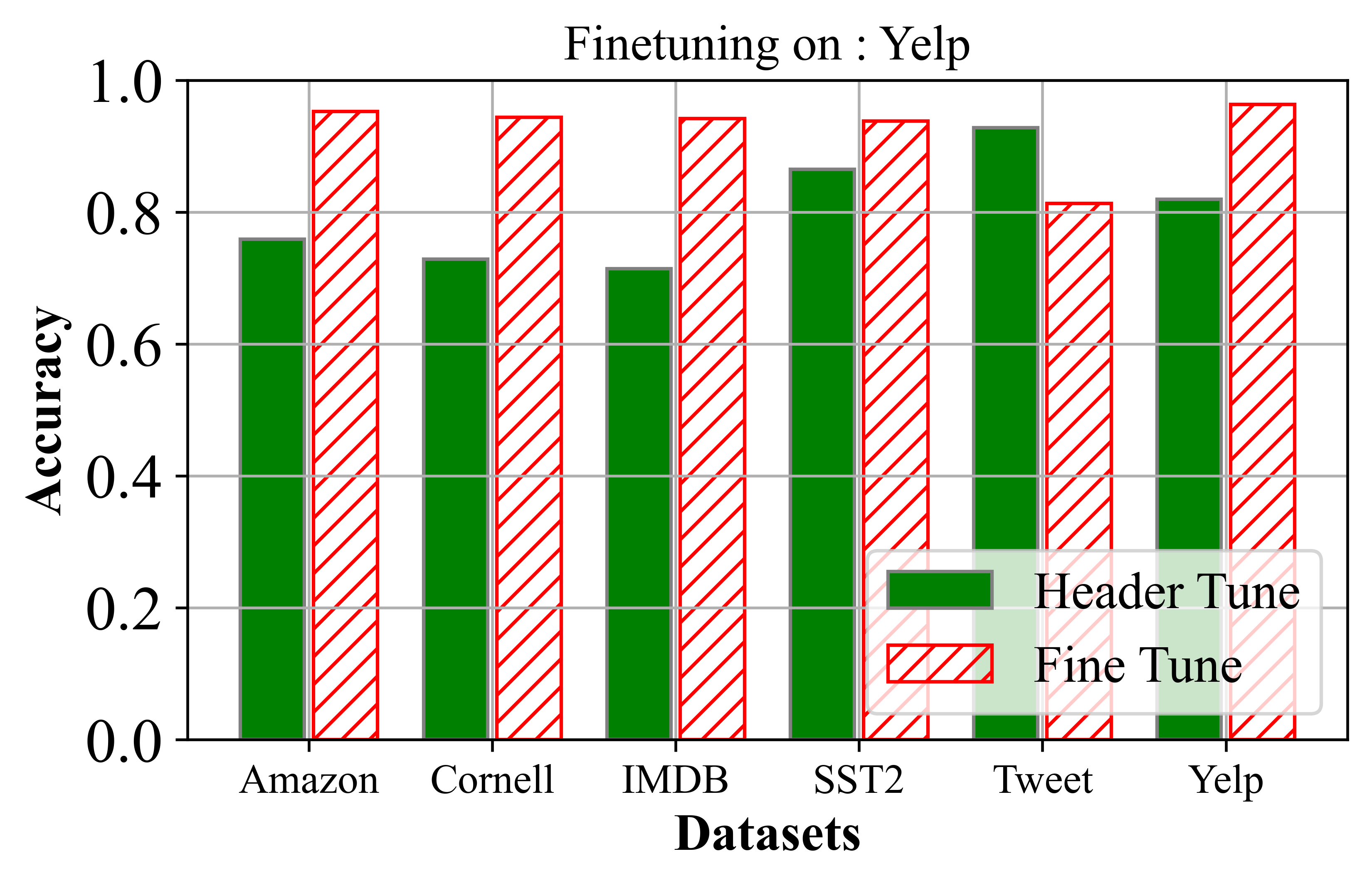}
  \caption{Performance comparison of FT Header layer to FT whole Model - Yelp dataset.}
  \label{fig:header_vs_FT6}
  \vspace{-.1in}
\endminipage
\end{figure*}

\subsection{Advantage of Ensemble}
\label{subsec:adv_ensmbl}
\iia{Our goal in this work is to find a suitable method that ensures better performance when data from the target domain is scarce, and therefore it is not possible to determine conclusively which of the available fine-tuned foundation models are adjacent to the target domain (if \daft or not)\footnote{\iia{There are a few established methods (e.g. LEEP score) to find the adjacency between two datasets, however these methods for determining domain-adjacency generally require a substantial amount of data from the target domain.}}. The advantage of using Ensemble method are: 1) there is no constraint on the \daft model size and architecture, 2) no need of knowing the weights of the \daft models, just an API access to the \daft models is adequate.
In our study, we found that even if we do not know how domain-adjacent the available models are, their aggregate ensemble can still perform well (with zero or very limited training) as long as there are some models in the ensemble that are domain-adjacent (although we may not know which ones). Furthermore, as we have observed in the paper, the weighted ensemble of the DAFT models (\dafte) can perform similarly to or better than the best single (available) \daft model. In summary, without adequate data it may be hard to determine which models are \daft and which are not,  and a very low-data low-resource ensembling method like \daftez or \dafte can give us good results.}

\subsection{Ensemble Layer of \dafte}
\label{appnd_subsec_ens_layer}
\subsubsection{Weight update of $LR$}
Our ensemble technique uses the final layer output of the DAFT models as the input of the  ensemble layer. When performing DAFT-E with $LR$,  If $w_i$ is the weight of model $i$ in the Ensemble weight layer, then the $LR$ performs the following:
\begin{align}
    \min_{w_i \in \mathbf{w}} \sum_{(X,y) \in D_T} \ell \bigg( \Big(\sum_i w_i M_i(X)\Big), y)\bigg)
    + R(\mathbf{w}) \nonumber
\end{align}
where, $\ell$ is the loss calculated using the output from the DAFT models ($M_i(X)$), to the ground truth $y$. Also, the term $R(\mathbf{w})$ can be used to regularize (or control) the number of models ($\bar{N}$) we want to use for our final ensemble. To keep the method simple, we did not use the regularization when implementing \textsf{DAFT-E}.

\begin{wrapfigure}{r}{0.47\textwidth}
% \begin{minipage}{0.47\textwidth}
\centering
\vspace{-30pt}
\includegraphics[width=0.8\linewidth]{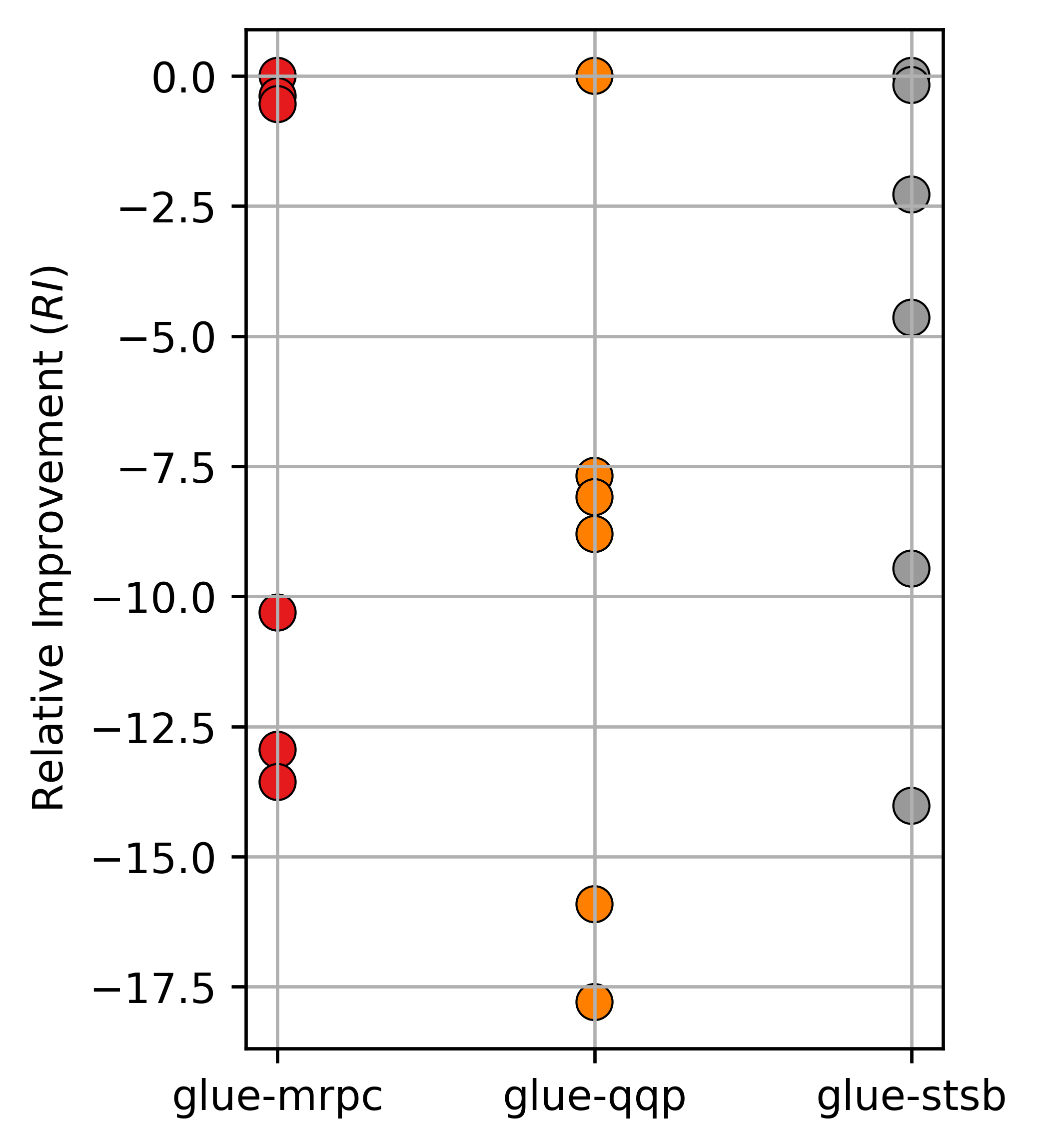}
\caption{{\em Relative Improvement of {\daft}s compared to the Single Best \daft} for Text similarity task: For each test dataset, we consider 6 \daft FMs (fine-tuned on data different from the test data), and consider the RI of \daftz over the single-best \daft FM (out of the 6). 
Values less than 0 indicate performance degradation.}
\label{fig:RI_single_DAMs_text}
\vspace{-30pt}
\end{wrapfigure}

\subsubsection{Regression methods: $LR$ and $RF$}
We observed the following in terms of performance of the two regression methods: (i)~$LR$ has better performance and less variation in performance for smaller sample data, (ii)~for higher sample data, i.e., $n$>32, $RF$ usually performs very similar to $LR$, (iii)~for sentiment analysis tasks, $LR$ and $RF$ both perform well with similar performance at higher sample data, and (iv)~for textual similarity tasks, $LR$ shows minimal performance improvement with more samples, whereas $RF$ shows promising results.

For $LR$ we have used SGDRegressor from sklearn.linear\_model with maximum iteration of 3. The reason behind this small iteration number is because of the few shot regime. If we had used larger bound on iteration, overfitting might have casued more harm than good. We also used coefficient initialization = $1/N$, where $N$ is the number of \daft models used, so that at the start \dafte gives same weights to all the models.
For $RF$ we imported the RandomForestClassifier from sklearn.ensemble, and set the max depth = 2. The smaller max depth was chosen to avoid overfitting.

\subsection{Performance of \daft models 
%and \daftez 
for Textual similarity task}

Fig. \ref{fig:RI_single_DAMs_text} 
% and \ref{fig:RI_Ens_text} 
shows the Relative Improvement of {\daft}s compared to Single Best \daft and \daftez respectively for the textual similarity task with three datasets. Similar to the results with sentiment analysis task, we observe the performance deviation among the \daft models and usual better performance of \daftez.

\subsection{Performance of \daft models from Internet}
\label{sec:from_internet}

To evaluate the performance of the readily available \daft models from public repositories, we have downloaded $100$ sentiment analysis models from Huggingface using `popularity' as the sorting criterion. The performance of these $100$ models compared to the single best \daft and \daftez is shown in Figs. \ref{fig:RI_single_DAMs_100_internet} and \ref{fig:RI_Ens_100_internet} respectively. Similar to what we observed previously, \daftez still outperforms most of the \daft performances ($551$ among $600$ cases). Moreover, it is notable that a lot of the \daft models are very poor performing compared to our controlled \daft models introduced in Section \ref{Sec:DAM}.

\begin{figure*}[t]
\begin{minipage}{0.48\textwidth}
      \centering
\vspace{-20pt}
\includegraphics[width=0.8\linewidth]{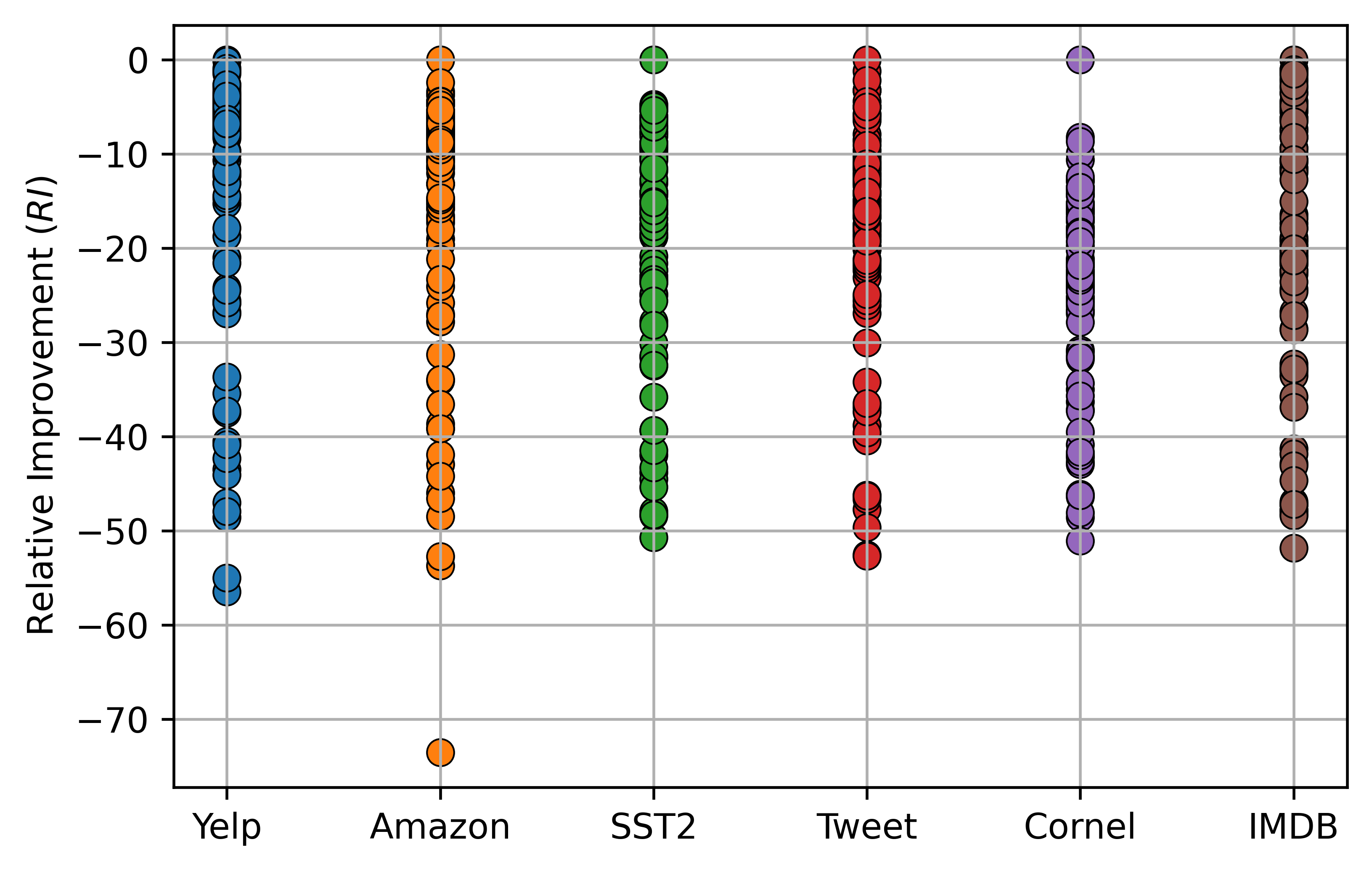}
\caption{{\em Relative Improvement of {\daft}s (from Huggingface) compared to the Single Best \daft} for sentiment analysis task. 
Values less than 0 indicate performance degradation.}
\label{fig:RI_single_DAMs_100_internet} 
\end{minipage}
\hfill
\begin{minipage}{0.48\textwidth}
      \centering
\vspace{-20pt}
\includegraphics[width=0.8\linewidth]{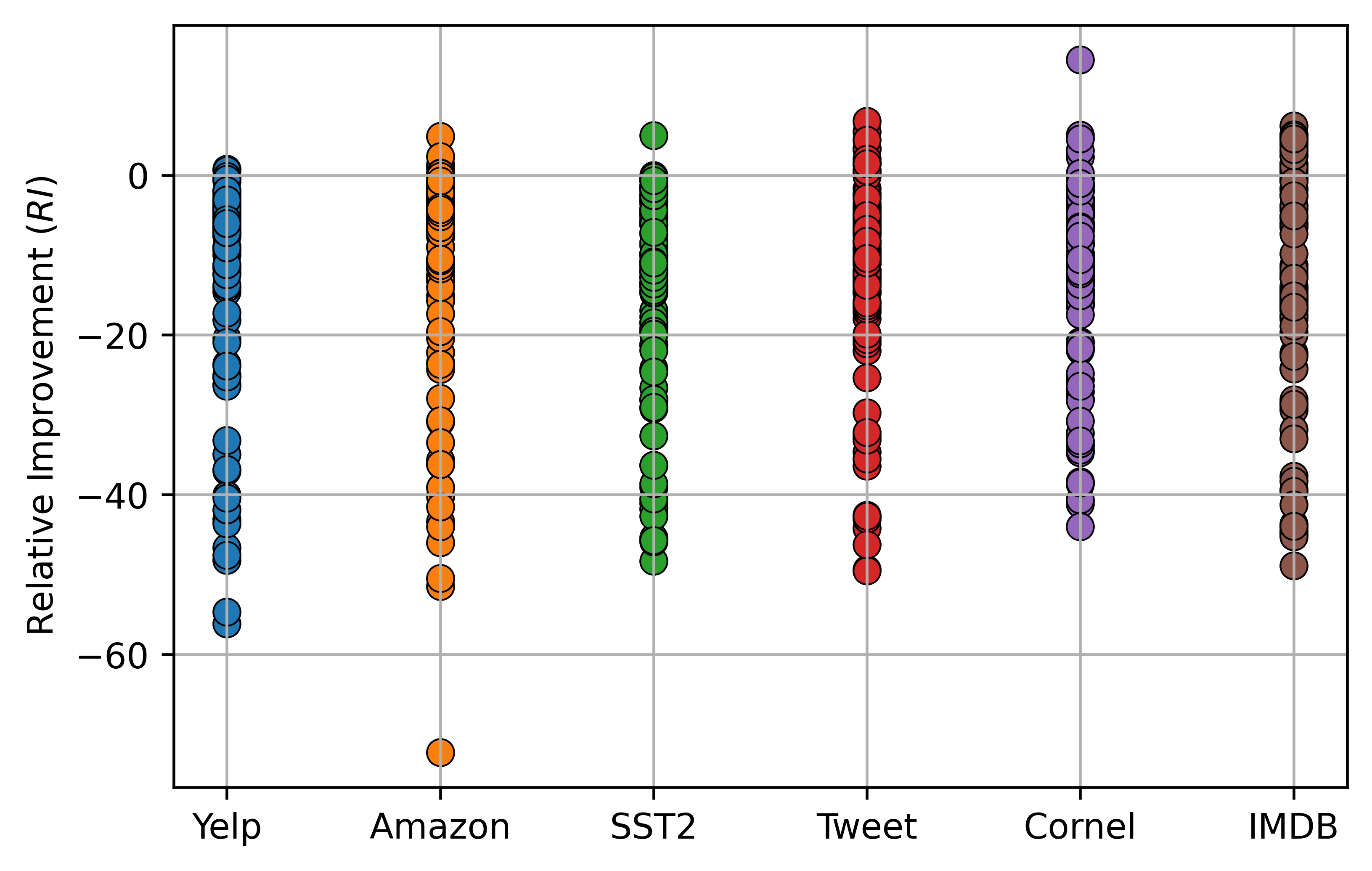}
\caption{{\em Relative Improvement of {\daft}s (from Huggingface) compared to \daftez} for sentiment analysis task. 
Values less than 0 indicate performance degradation.}
\label{fig:RI_Ens_100_internet} 
\end{minipage}
 \vspace{-10pt}
\end{figure*}

The performance of \dafte for the $100$ models from Huggingface is depicted in Fig. \ref{fig:DAFT-E_100_internet}. The results are consistent with the results that we observed in Fig. \ref{fig:Ensmbl_FFT}. In all cases, there is an upward trend in performance w.r.t. training data with the use of \dafte. Also, for $3$ of the $6$ datasets, \dafte catches up (or surpasses) the single best \daft with 128 training pairs. Since we do not have the FFT for this scenario, the FFT comparison is not shown in Fig. \ref{fig:DAFT-E_100_internet}.

\begin{figure*}
    \centering
\vspace{5pt}
\includegraphics[width=0.95\linewidth]{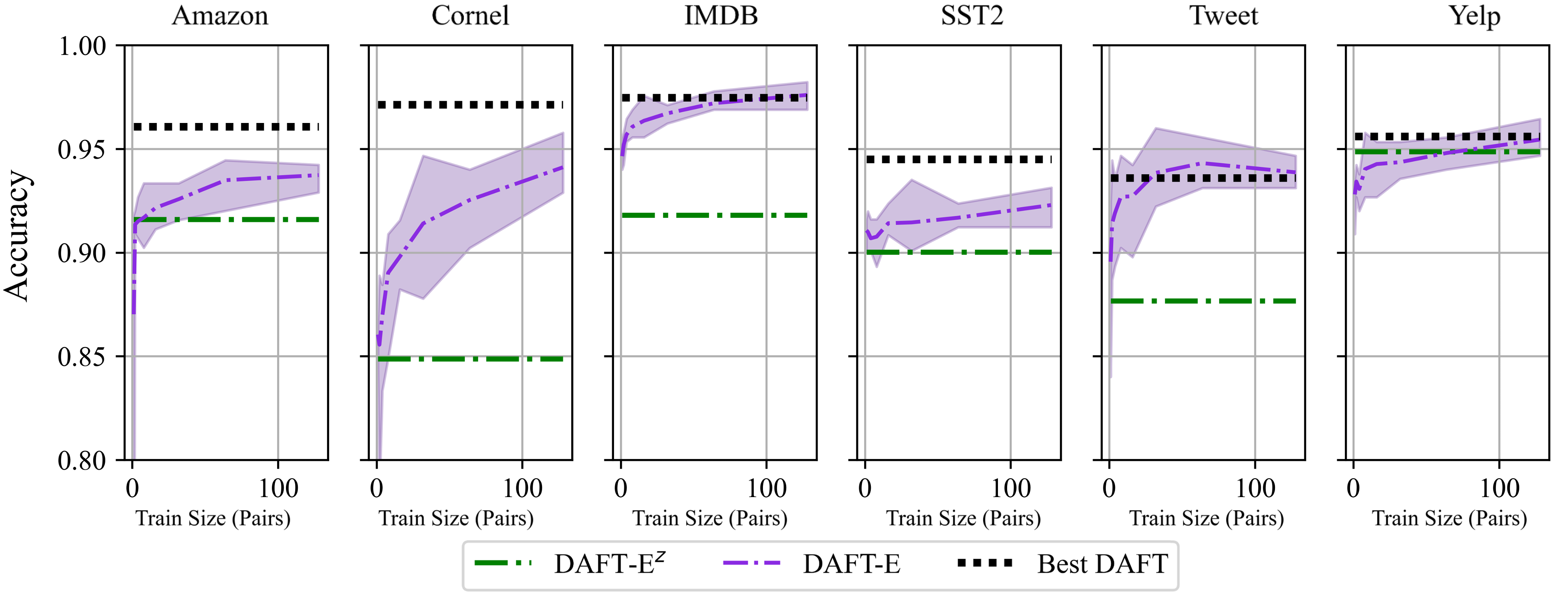}
\caption{{\em Performance comparison of \daftez, \dafte, and single-best.} Error interval for \dafte is based on the random choice of the few-shot samples used to learn the weights of the ensemble aggregated over 10 trials.}
\label{fig:DAFT-E_100_internet} 
\end{figure*}

\color{black}

\subsection{Proof of Propositions}

\paragraph{Proof of Proposition \ref{prop:daftez}}
In the following, we let $\mathcal{N} = \{1, 2, \cdots, N\}$ denote the set of indices of the DAFT models in the ensemble.
\begin{align}
& \mathbb{E} (\ell(\text{\daftz}))
= \;  \mathbb{E} (\ell(M_i(X),y)) \;\;\; \forall i \in \mathcal{N} \nonumber \\
= \;& \frac{1}{N} \sum_{i \in \mathcal{N}} \ell(M_i(X),y) \nonumber \\
= \; & \frac{1}{N} \sum_{i \in \mathcal{N}} \ell(\mu_i,y) \;\; \text{[denoting $M_i(X) $ as $ \mu_i$]} \nonumber \\
\geq \; & \ell\left( \sum_{i \in \mathcal{N}} \frac{1}{N} \mu_i,y\right) \; \text{[assuming $\ell$ is convex in $\mu_i$]} \nonumber \\
= \; & \text{loss of average ensemble} 
= \; \ell(\text{\daftez}).
\end{align}
Note that the derivation assumes $\ell$ to be convex in $\mu_i$, the outputs from the \daft models. 
% , which is a common assumption for loss functions. 
We assume a convex loss function since the loss is calculated using the
output probabilities of each DAFT model, which is a continuous value. Also, the most popular loss functions like MAE, or MSE are directly proportional to the distance between the predicted output $\hat{y}_{\text{pred}}$ and the true output $y$, and are convex functions~\citep{terven2023loss}.
\qed

\paragraph{Proof of Proposition \ref{prop:dafte-vs-opt}}
If the ensemble weights for the $i^{th}$ model is $w_i$, then for the weighted ensemble method we have;
\begin{align}
\label{eq:proof3_1}
& \ell(\text{\dafte}) \nonumber \\
= & \; \text{loss of weighted ensemble} \nonumber \\
    = & \; \min_{w_i, i\in \mathcal{N}} \left[ \ell \left(\sum_{i} w_i M_i \right)\right] \nonumber \\
 \leq & \; \ell (M_i), \;\; \forall i \in \mathcal{N}. 
\end{align}
Note that the above derivation assumes that the set weights $w_i, i\in \mathcal{N}$, have been optimized (trained using the fine-tuning dataset from the target domain) to minimize the loss function $\ell \left(\sum_{i} w_i M_i \right)$. 

Note that the base model that $M_i$ is built from is $B_{\kappa(i)}$. Further, $\tilde{M}_i$ denotes the corresponding FFT, i.e., the model built from $B_{\kappa(i)}$ by fine-tuning on the target dataset $D_T$. Now, let us assume that for two \daft models $M_1$ and $M_2$ developed from the same base model and fine-tuned on different datasets, i.e., $D_{\nu(1)}$ and $D_{\nu(2)}$, to have the following bound on their losses;
    \begin{align}
        |\ell(M_1)-\ell(M_2)| \leq \rho(D_{\nu(1)},D_{\nu(2)}),
    \end{align}
where $\rho$ is an appropriately defined distance measure between the datasets $D_{\nu(1)}$ and $D_{\nu(2)}$. Then from Equation \ref{eq:proof3_1} we have;
\begin{align}
\label{eq:proof3_2}
 \ell(\text{\dafte})
    \leq & \ell(M_i) \leq \ell(\tilde{M}_i) + \rho(D_{\nu(i)},D_T),  \forall i.
\end{align}

% We define the function  that relates the loss values of two models as follows;
% \begin{definition}
% \begin{align}
% \label{eq:proof3_3}
%     \mu(M_1,M_2) = 
% \end{align}
% \end{definition}
Now, let us denote $\tilde{M}_*$ be the model fine-tuned on $D_T$ and performs the best among all the models when fine-tuned on $D_T$ (all possible FFTs). Then from \ref{eq:proof3_2} and by defining  $\mu(M_1,M_2) \overset{\Delta}{=} \ell(M_1)- \ell(M_2)$, we have
\begin{align}
\label{eq:proof3_4}
     \ell(\text{\dafte})
    \leq & \; \ell(\tilde{M}_*) + \rho(D_{\nu(i)},D_T) \nonumber \\
    & \; \;+ \mu(\tilde{M}_i ,\tilde{M}_*); \;\; \forall i.
\end{align}
Since Equation \ref{eq:proof3_4} holds for all $i \in \mathcal{N}$, Equation \ref{prop:dafte-vs-opt-eqn} follows by taking a minimum over all $i$, completing the proof.
\qed

Note that Equation \ref{prop:dafte-vs-opt-eqn} bounds the gap between the mimimum loss possible under any model and the loss of \dafte. The result is intuitive. The term $\mu()$ 
%indicates that the performance gap (sub-optimality of \dafte) depends on the on the performance of any of the base models (from where the \daft models were built) compared to the best possible base model
tells us that the performance gap depends on how good the base models (corresponding to the DAFT models in the ensemble) are, when compared to the best possible model, when they are all trained on the full target dataset. The term $\rho()$ implies that the performance gap also depends on how closely the datasets used to generate the DAFT models represent the target dataset. Note that the minimum over $i$ is taken on the sum of $\mu()$ and $\rho()$, instead of of each of the two terms individually. However, for an ensemble of DAFT models obtained by fine-tuning a ``good'' set of FMs with a large number of datasets, the approximation term $\mu()+ \rho()$ is expected to be small, as argued in Section~\ref{subsec:dafte_opt}.

% \newpage
% \pagebreak
% \newpage
% \pagebreak

%%

\begin{table*}
\small
  \caption{Relative Improvement of \daftsq compared to \ft (Sentiment Analysis). The base model for all the models (\daft and \ft) here is Roberta-base.}
  \label{Tab:FT_base_and_pretuned_appnd}
  \begin{tabular}{clrrrrrrrr}
    \toprule
    $\;$ Dataset  $\;$ &  \multicolumn{1}{c}{$\;\;$ \daft Name $\;\;$}  & $\; n=2 \;$ & $\; n=4 \; $ & $\; n=8 \;$& $ n=16 \;$ & $ n=32 \;$ & $ n=64\;$ & $n=128$ & $n=256$  \\
    \midrule
    \multirow{6}{4em}{Amazon} & \daft -  Cornell & 80.42 & 76.28 & 81.10 & 67.16 & 55.45 & 32.55& 0.50 & 0.48 \\
                              & \daft - IMDB & 78.33 & 77.79 & 83.46 & 68.53 & 57.12 & 37.85 & 2.66 & 1.92 \\
                              & \daft - SST2 & 78.11 & 75.33 & 78.67 & 65.6 & 54.45 & 34.45 & 0.57 & 0.77 \\
                              & \daft - Tweet & 61.33 & 67.70 & 71.34 & 56.76 & 47.07 & 31.19 & -1.11 & -0.85\\
                              & \daft - Yelp & 81.59 & 79.18 & 83.50 & 68.95 & 53.48 & 35.24 & 1.42 & 1.54 \\ %\cmidrule{2-10}
                              \cmidrule{2-10}
                              & Average & 75.95	& 75.25	& 79.62   & 65.40	& 53.51	& 34.26	& 0.81	& 0.77 \\
   \midrule
    \multirow{6}{4em}{Cornell} & \daft - Amazon & 73.60 & 71.74 & 68.88 & 63.6 & 61.01 & 38.06 & 10.25 & 1.66 \\
                                & \daft - IMDB & 68.43 & 64.88 & 64.64 & 62.43 & 61.23& 38.52 & 9.58 & 3.60\\
                                & \daft - SST2 & 74.8 & 72.11 & 68.89 & 66.71 & 63.54 & 38.97 & 9.64 & 2.67 \\ 
                              & \daft - Tweet & 53.72 & 52.43 & 49.77 & 48.37 & 51.15 & 31.50 & 6.11 & 2.23\\ 
                              & \daft - Yelp & 69.22 & 66.87 & 64.13 & 62.47 & 51.87& 32.19 & 8.65 & 0.3 \\ \cmidrule{2-10}
                              & Average & 67.95	& 65.61	& 63.26	& 60.72	& 57.77	& 35.85 & 	8.85 & 2.09\\
   \midrule
    \multirow{6}{4em}{IMDB} & \daft - Amazon & 81.35 & 81.84 & 81.66 & 70.94 & 61.86 & 50.11 & 2.53 & 0.65\\
                            & \daft - Cornell & 77.79 & 77.09 & 80.67 & 66.68 & 57.4 & 49.72 & 2.46 & -0.03\\
                            & \daft - SST2 & 76.10 & 75.28 & 77.99 & 64.86 & 57.45 & 47.36 & -1.33 & -0.977 \\
                              & \daft - Tweet & 59.12 & 60.27 & 60.49 & 58.10 & 53.54 & 45.20 & -0.62 & -0.70\\
                              & \daft - Yelp & 76.93 & 78.72 & 80.11 & 69.47 & 60.73 & 45.82 & -2.52 & 0.247 \\ 
                              \cmidrule{2-10}
                              &Average & 74.26& 74.64	& 76.19	& 66.01	& 58.20	& 47.64	& 0.10	& -0.16\\
   \midrule
    \multirow{6}{4em}{SST2} & \daft - Amazon & 76.35 & 73.84 & 73.54 & 70.05 & 72.62 & 58.52& 11.47 & 2.03 \\
                            & \daft - Cornell & 82.05 & 78.97 & 78.81 & 80.32 & 77.56 & 64.29 & 16.84 & 5.41 \\
                            & \daft - IMDB & 71.55 & 75.20 & 72.81 & 74.51 & 71.62 & 62.14 & 14.46 & 3.85\\
                              & \daft - Tweet & 62.25 & 60.66 & 60.68 & 62.66 & 63.46 & 53.84 & 10.30 & 0.92 \\
                              & \daft - Yelp & 73.42 & 74.51 & 73.2 & 72.18 & 74.17 & 54.1 & 10.89 & 2.75 \\ 
                              \cmidrule{2-10}
                              & Average & 73.13	& 72.64	& 71.81 & 	71.95 & 	71.88 & 	58.58 &	12.79 &	2.99\\
   \midrule
    \multirow{6}{4em}{Tweet} & \daft - Amazon & 60.22 & 53.93 & 55.79 & 41.88 & 42.67 & 26.16 & 1.2 & -2.2 \\
                            & \daft - Cornell & 56.37 & 49.64 & 50.94 & 52.62 & 39.55 & 23.64 & 1.03 & -1.58 \\
                            & \daft - IMDB & 59.72 & 53.76 & 52.58 & 55.26 & 45.22 & 30.19 & 4.69 & 1.59 \\
                              & \daft - SST2 & 62.23 & 55.20 & 57.46 & 54.1 & 41.79 & 26.48 & 2.29 & -1.06 \\
                              & \daft - Yelp & 64.83 & 58.69 & 60.85 & 57.37 & 39.85 & 23.15 & 2.5 & -1.6 \\ 
                              \cmidrule{2-10}
                              & Average & 60.67& 54.25	& 55.53 &	52.25 & 	41.82 & 	25.92	& 2.34	& -0.97 \\
   \midrule
    \multirow{6}{4em}{Yelp} & \daft - Amazon & 79.81 & 84.56 & 71.46 & 85.73 & 49.85 & 26.89 & -2.32 & -0.41 \\
                            & \daft - Cornell & 74.39 & 83.52 & 67.09 & 81.15 & 45.62 & 29.28 & -0.65 & -0.81 \\
                               & \daft - IMDB & 77.45 & 83.51 & 67.84 & 83.75 & 47.97 & 29.62 & 1.02 & 0.29 \\
                                & \daft - SST2 & 75.4 & 80.74 & 66.07 & 80.69 & 45.82 & 26.711 & -0.86 & -1.48 \\
                              & \daft - Tweet & 67.36 & 72.68 & 61.60 & 77.06 & 43.10 & 26.81 & -0.79 & -1.22\\ 
                              \cmidrule{2-10}
                              & Average & 74.88 &	81.00	& 66.81	& 81.68 &	46.47 &	27.86 &	-0.72	& -0.73 \\
    % \texttt{{\char'134}table}& 300 & For tables\\
    % \texttt{{\char'134}table*}& 400& For wider tables\\
    \bottomrule
  \end{tabular}
\end{table*}

\begin{table*}
\small
  \caption{Relative Improvement of \daftsq compared to \ft (Text Similarity). The R, B, X at the end of the \daft names denote the base model of the \daft models, and they denote: Roberta, Bert, Xlnet base models respectively.}
  \label{Tab:FT_base_and_pretuned_appnd_text}
  \begin{tabular}{clrrrrrrrr}
    \toprule
    $\;$ Dataset $\;$ &  \multicolumn{1}{c}{$\;$ \daft Name $\;$}  & $\; n=2 \;$ & $\; n=4 \; $ & $\; n=8 \;$& $ n=16 \;$ & $ n=32 \;$ & $ n=64\;$ & $n=128$ & $n=256$ \\
    \midrule
    \multirow{3}{4em}{MRPC} & \daft -  QQP-R & 34.78	& 32.60 &	32.08&	-6.49&	24.68&	18.15&	20.21&	10.83 \\
                              & \daft - STSB-R &55.83	& 54.79&	61.86&	13.01&	41.26&	32.65&	28.13&	18.20 \\
                              & \daft - QQP-B &36.45	&34.58	&39.62	&-1.60	& 24.19 &	16.03&	12.41&	8.69\\
                              & \daft - STSB-B & 54.27	&52.31	& 59.27	& 12.49	& 37.97	& 27.81 &	23.47	& 12.33\\
                              & \daft - QQP-X & 33.75& 29.95 & 	35.87 &	-2.57	& 24.02 &	16.12 &	15.38 &	6.05\\
                              & \daft - STSB-X & 54.22	& 53.29 &	59.77 &	13.61 &	40.77	& 30.08 &	25.80 &	16.34\\
                             \cmidrule{2-10}
                              & Average & 44.88	& 42.92 &	48.08	& 4.74	& 32.15	& 23.47 &	20.90 &	12.07\\
   \midrule
    \multirow{3}{4em}{QQP} & \daft - MRPC-R & 75.56	& 78.10& 	88.81& 	86.98& 	66.05&	47.51&	16.33	& 4.71 \\
                                & \daft - STSB-R & 81.24 &	80.61 &	99.21 &	97.98 &	77.47 &	57.41	& 19.36&	7.38\\
                                & \daft - MRPC-B & 63.40& 65.81 & 	72.71 &	78.10 &	59.59 &	42.61 &	9.76	& -0.62\\
                                & \daft - STSB-B &79.18	& 82.73 &	89.69 &	94.59 &	69.40	& 51.00 &	14.10 &	2.03\\
                                & \daft - MRPC-X & 61.59 &	65.76 &	72.62 &	75.59 &	56.21	& 42.88 &	8.57 &	-2.41\\
                                & \daft - STSB-X & 76.61& 78.03 & 	83.61 &	88.56 &	63.45 &	49.58 &	15.04 &	1.97\\
                                \cmidrule{2-10}
                                & Average & 72.93& 75.17 &	84.44 &	86.97 &	65.36 &	48.50 &	13.86 &	2.18\\
   \midrule
    \multirow{3}{4em}{STSB} & \daft - MRPC-R & 46.89	& 53.02 &	50.07 &	51.96 &	45.97& 28.01	& 1.26	& 0.64\\
                            & \daft - QQP-R & 51.78	& 50.82	& 57.42	& 63.87	& 56.15	& 36.55	& 4.85	& 1.95 \\ 
                            & \daft - MRPC-B & 42.88& 40.53 &	44.46 &	47.67 &	44.77 &	24.74 &	-3.51 &	-3.24 \\
                            & \daft - QQP-B & 56.55	& 58.26	& 56.90	& 58.49	& 57.51	& 33.36	& 1.83	& 0.89\\
                            & \daft - MRPC-X & 40.73& 39.28 &	35.59 &	41.34 &	41.23 &	22.05	& -5.52 &	-3.76 \\
                            & \daft - QQP-X & 51.34	& 53.98	& 52.69 &	55.15 &	53.66	& 30.73	& 0.37	& -0.74 \\
                            \cmidrule{2-10}
                            & Average & 48.36 &49.31 & 49.52 & 53.08& 49.88 &	29.24 &-0.12	&-0.71 \\
    \bottomrule
  \end{tabular}
\end{table*}

%%%%

%%%%

\begin{table*}
\small
  \caption{Relative Improvement of \daftsq compared to \dafte (Sentiment Analysis). The base model for all the \daft models here is Roberta-base.}
  \label{Tab:FT_pretuned_and_Ens_weight_appnd}
  \begin{tabular}{clrrrrrrrr}
    \toprule
    $\;$ Dataset  $\;$ &  \multicolumn{1}{c}{$\;\;$ \daft Name $\;\;$}  & $\; n=2 \;$ & $\; n=4 \; $ & $\; n=8 \;$& $ n=16 \;$ & $ n=32 \;$ & $ n=64\;$ & $n=128$ & $n=256$  \\
    \midrule
    \multirow{6}{4em}{Amazon} & \daft - Cornell & -1.60 & -1.91 & -1.57 & -1.38 & -1.43 & -4.177 & -2.35 & -1.78 \\
                            & \daft - IMDB & -2.74 & -1.07 & -0.29 & -0.57 & -0.38 & -0.35 & -0.26 & -0.37 \\
                            & \daft - SST2 & -2.85 & -2.43 & -2.89 & -2.3 & -2.06 & -2.80 & -2.30 & -1.50 \\
                              & \daft - Tweet & -12.01 & -6.68 & -6.87 & -7.51 & -6.75 & -5.17 & -3.92 & -3.09 \\
                              & \daft - Yelp & -0.96 & -0.30 & -0.27 & -0.32 & -2.683 & -2.238 & -1.46 & -0.753 \\ \cmidrule{2-10}
                              & Average & -4.03 &	-2.48	& -2.38	&-2.42	& -2.66 &	-2.95 &	-2.06 &	-1.50\\
   \midrule
    \multirow{6}{4em}{Cornell} & \daft - Amazon & -0.71 & -0.03 & -0.04 & -1.81 & -2.14 & -2.34 & -1.27 & -3.06 \\ 
                                & \daft - IMDB & -3.66 & -4.02 & -2.55 & -2.51 & -2.03 & -2.02 & -1.87 & -1.21\\
                                & \daft - SST2 & -0.02 & 0.187 & -0.04 & 0.05 & -0.62 & -1.70 & -1.82 & -2.10 \\
                              & \daft - Tweet & -12.08 & -11.27 & -11.35 & -10.95 & -8.15 & -6.99 & -4.98 & -2.52\\
                              & \daft - Yelp & -3.21 & -2.87 & -2.85 & -2.49 & -7.72 & -6.50 & -2.70 & -4.38 \\ \cmidrule{2-10}
                               & Average& -3.94	& -3.60	& -3.37 &	-3.54 &	-4.13	& -3.91 &	-2.53	& -2.65 \\
   \midrule
    \multirow{6}{4em}{IMDB} & \daft - Amazon & 1.25 & 0.57 & -0.35 & -0.43 & -0.79 & -1.25 & -0.48 & -0.65 \\
                            & \daft - Cornell & -0.73 & -2.05 & -0.89 & -2.92 & -3.53 & -1.51 & -0.54 & -1.32 \\
                            & \daft - SST2 & -1.675 & -3.05 & -2.36 & -3.98 & -3.50 & -3.06 & -4.22 & -2.25 \\
                              & \daft - Tweet & -11.16 & -11.35 & -11.96 & -7.92 & -5.89 & -4.48 & -3.53 & -1.98 \\
                              & \daft - Yelp & -1.21 & -1.25 & -1.20 & -1.29 & -1.49 & -4.07 & -5.39 & -1.05 \\ \cmidrule{2-10}
                              & Average & -2.70	& -3.41 &	-3.35	& -3.31 &	-3.04 &	-2.87	& -2.83 & -1.45\\
   \midrule
    \multirow{6}{4em}{SST2} & \daft - Amazon & -2.89 & -2.64 & -2.94 & -5.91 & -3.70 & -4.71 & -5.33 & -4.65 \\
                            & \daft - Cornell & 0.25 & 0.23 & -0.02 & -0.24 & -0.95 & -1.25 & -0.77 & -1.49 \\
                            & \daft - IMDB & -5.53 & -1.88 & -3.38 & -3.45 & -4.26 & -2.54 & -2.79 & -2.95\\
                              & \daft - Tweet & -10.65 & -10.03 & -10.16 & -10.01 & -8.81 & -7.53 & -6.33 & -5.69 \\
                              & \daft - Yelp & -4.50 & -2.27 & -3.16 & -4.74 & -2.84 & -7.37 & -5.82 & -3.97 \\ \cmidrule{2-10}
                              & Average & -4.67	& -3.32 &	-3.94 &	-4.87 &	-4.11 &	-4.68 &	-4.21 &	-3.75\\
   \midrule
    \multirow{6}{4em}{Tweet} & \daft - Amazon & -4.89 & -5.09 & -4.98 & -11.98 & -3.42 & -3.64 & -2.69 & -1.63 \\
                            & \daft - Cornell & -7.17 & -7.74 & -7.94 & -5.31 & -5.53 & -5.56 & -2.86 & -1.00 \\
                            & \daft - IMDB & -5.19 & -5.21 & -6.93 & -3.68 & -1.69 & -0.57 & 0.67 & 2.18  \\
                            & \daft - SST2 & -3.70 & -4.31& -3.96 & - 4.40 & -4.02 & -3.40 & -1.64 & -0.48 \\ 
                              & \daft - Yelp & -2.15 & -2.17 & -1.89 & -2.36 & -5.33 & -5.94 & -1.44 & -1.02\\ \cmidrule{2-10}
                              & Average & -4.62	& -4.91 &	-5.14 &	-5.54 &	-4.00 &	-3.82 &	-1.59 &	-0.39 \\
   \midrule
    \multirow{6}{4em}{Yelp} & \daft - Amazon & 0.18 & -0.84 & 0.62 & 0.10 & 0.60 & -3.20 & -3.85 & -0.98 \\
                            & \daft - Cornell & -2.83 & -1.40 & -1.93 & -2.37 & -2.25 & -1.37 & -2.21 & -1.37 \\ 
                            & \daft - IMDB & -1.13 & -1.41 & -1.50 & -0.97 & -0.67 & -1.11 & -0.57 & -0.29 \\
                            & \daft - SST2 & -2.27 & -2.90 & -2.53 & -2.62 & -2.11 & -3.33 & -2.42 & -2.04 \\
                              & \daft - Tweet & -6.75 & -7.22 & -5.16 & -4.58 & -3.94 & -3.26 & -2.34 & -1.79 \\ \cmidrule{2-10}
                              & Average & -2.56 &	-2.76 &	-2.10	& -2.09	& -1.68 &	-2.46 &	-2.28 &	-1.29 \\
    % \texttt{{\char'134}table}& 300 & For tables\\
    % \texttt{{\char'134}table*}& 400& For wider tables\\
    \bottomrule
  \end{tabular}
\end{table*}

\begin{table*}
\small
  \caption{Relative Improvement of \daftsq compared to \dafte - LR (Text Similarity). The R, B, X at the end of the \daft names denote the base model of the \daft models, and they denote: Roberta, Bert, Xlnet base models respectively.}
  \label{Tab:FT_pretuned_and_Ens_weight_appnd_text}
  \begin{tabular}{clrrrrrrrr}
    \toprule
    $\;$ Dataset $\;$ &  \multicolumn{1}{c}{$\;$ \daft Name $\;$}   & $\; n=2 \;$ & $\; n=4 \; $ & $\; n=8 \;$& $ n=16 \;$ & $ n=32 \;$ & $ n=64\;$ & $n=128$ & $n=256$  \\
    \midrule
    \multirow{3}{4em}{MRPC} & \daft -  QQP-R &  -11.80 &	-11.25 &	-14.47 &	-13.93	& -8.60	 & -2.85	& 4.15 &	7.20\\
                              & \daft - STSB-R &  1.98	& 3.60 &	4.81	& 4.01 & 	3.56 &	9.08 &	11.00	& 14.33\\
                              & \daft - QQP-B &  -10.71	& -9.92 & -9.59 &	-9.44 &	-8.95 &	-4.59 &	-2.61 &	5.13\\
                              & \daft - STSB &  0.95 &1.94 &	3.14 &	3.53 &	1.15 &	5.10 &	6.97 &	8.66\\
                              & \daft - QQP-X &  -12.47	& -13.02 &	-12.02 &	-10.33 &	-9.08 &	-4.52 &	-0.04 &	2.58\\
                              & \daft - STSB-X & 0.92 &	2.60 &	3.46 &	4.56 &	3.20 &	6.96 &	8.99 &	12.54 \\
                              \cmidrule{2-10}
                              & Average &  -5.19& -4.34 &	-4.11 &	-3.60 &	-3.12 &	1.53 &	4.74 &	8.41\\
   \midrule
    \multirow{3}{4em}{QQP} & \daft - MRPC-R & -1.37 &	-1.01 &	1.90	&-0.05 &	2.44 &	1.73 &	6.68 &	7.89 \\
                            & \daft - STSB-R & 1.82 &	0.39 &	7.51	& 5.83 &	9.49 &	8.55 &	9.46	&10.65\\ 
                            & \daft - MRPC-B & -8.20& -7.84 &	-6.80 &	-4.80 &	-1.55 &	-1.65 &	0.65& 2.40\\ 
                            & \daft - STSB-B & 0.66 &	1.57	& 2.37 &	4.01 &	4.51 &	4.13 &	4.64 &	5.13\\ 
                            & \daft - MRPC-X & -9.22 & -7.87& -6.84 &	-6.14 &	-3.63 &	-1.46 &	-0.43	& 0.56\\ 
                            & \daft - STSB-X & -0.78 &	-1.05 &	-0.91 &	0.79 &	0.83 &	3.16 &	5.50 &	5.07\\ 
                            \cmidrule{2-10}
                            & Average & -2.85	& -2.64 &	-0.46 &	-0.06 &	2.01& 2.41	&4.42 &	5.28 \\
   \midrule
    \multirow{3}{4em}{STSB} & \daft - MRPC-R &-6.23	& -3.43 &	-3.72 &	-2.61 &	-4.53 &	-0.39 &	2.68 &	4.96\\
                            & \daft - QQP-R & -3.11	&-4.82 &	0.99	& 5.03 &	2.13 &	6.25 &	6.32 &	6.32\\ 
                            & \daft - MRPC-B & -8.79 & -11.31 &-7.32 &-5.35 & -5.32	& -2.93 &	-2.15 &	0.91\\ 
                            & \daft - QQP-B & -0.07	& -0.12 &	0.66 &	1.58 &	3.02 &	3.77 &	3.26 &	5.21\\ 
                            & \daft - MRPC-X & -10.16	& -12.10 &	-13.01 &	-9.41 &	-7.63 &	-5.03 &	-4.20 &	0.36\\ 
                            & \daft - QQP-X & -3.39	& -2.82 &	-2.04 &	-0.56 &	0.50 &	1.73 &	1.78 &	3.51\\ 
                            \cmidrule{2-10}
                            & Average &-5.29 &-5.77 &	-4.07	& -1.89 &	-1.97 &	0.57 &	1.28 &	3.55
                            \\
    \bottomrule
  \end{tabular}
\end{table*}

\begin{table*}
\small
  \caption{Relative Improvement of \daftsq compared to \dafte - RF (Text Similarity). The R, B, X at the end of the \daft names denote the base model of the \daft models, and they denote: Roberta, Bert, Xlnet base models respectively.}
  \label{Tab:FT_pretuned_and_Ens_weight_appnd_text_RF}
  \begin{tabular}{clrrrrrrrr}
    \toprule
    $\;$ Dataset $\;$ &  \multicolumn{1}{c}{$\;$ \daft Name $\;$}   & $\; n=2 \;$ & $\; n=4 \; $ & $\; n=8 \;$& $ n=16 \;$ & $ n=32 \;$ & $ n=64\;$ & $n=128$ & $n=256$  \\
        \midrule
    \multirow{3}{4em}{MRPC} & \daft -  QQP-R & 12.91	& -2.02 &	-10.60	 & -14.67 &	-11.38	&-7.66 &	-3.32 &	-0.07 \\
                              & \daft - STSB-R & 30.55 &	14.37&	9.56&	3.12 &	0.40 &	3.67 &	3.04 &	6.58  \\
                              & \daft - QQP-B & 14.31	& -0.56 &	-5.49	& -10.21&	-11.73&	-9.32&	-9.59&	-1.99 \\
                              & \daft - STSB &  29.24	& 12.54 &	7.81 &	2.65 &	-1.94 &	-0.11 &	-0.70 &	1.29\\
                              & \daft - QQP-X & 12.05	& -3.98	& -8.03	& -11.09 &	-11.86 &	-9.25 &	-7.21 &	-4.38\\
                              & \daft - STSB-X & 29.20 &	13.26 &	8.15 &	3.67 &	0.05 &	1.66 &	1.17	& 4.91\\
                              \cmidrule{2-10}
                              & Average & 28.02 &	10.39 &	4.53 &	-0.85 &	-3.05 &	-4.96 &	-7.17 &	-6.78\\
   \midrule
    \multirow{3}{4em}{QQP} & \daft - MRPC-R & 5.92	& 2.20 &	2.44 &	-0.88 &	-1.49 &	-3.20 &	1.52 &	3.59 \\
                            & \daft - STSB-R & 9.35	& 3.64 &	8.08 &	4.95 &	5.29 &	3.29 &	4.17 &	6.23\\ 
                            & \daft - MRPC-B & -1.42 &	-4.85 &	-6.29 &	-5.59 &	-5.32 &	-6.42 &	-4.21 &	-1.69\\ 
                            & \daft - STSB-B & 8.10 &	4.86 &	2.92 &	3.15 &	0.50 &	-0.91 &	-0.42 &	0.93\\ 
                            & \daft - MRPC-X & -2.51 &	-4.88 &	-6.34 &	-6.92 &	-7.32 &	-6.24 &	-5.25 &	-3.46\\ 
                            & \daft - STSB-X & 6.56 &	2.16 &	-0.38 &	-0.04 &	-3.03	& -1.84 &	0.40 &	0.88\\ 
                            \cmidrule{2-10}
                            & Average &  7.39	& 3.24	& 0.54	& -0.83 &	-3.83 &	-4.85 &	-4.83 &	-3.99\\
   \midrule
    \multirow{3}{4em}{STSB} & \daft - MRPC-R & 11.08 &	5.44	& -4.39	 &-5.82	& -8.44 &	-6.09 &	-2.41 &	-0.63\\
                            & \daft - QQP-R & 14.77 &	3.92 &	0.29 &	1.56 &	-2.06 &	0.18 &	1.05 &	0.66\\ 
                            & \daft - MRPC-B & 8.05 &	-3.17 &	-7.96 &	-8.48 &	-9.20 &	-8.49 &	-7.00 &	-4.47\\ 
                            & \daft - QQP-B & 18.38	& 9.05	& -0.04	& -1.78	& -1.21	& -2.17 &	-1.86 &	-0.39\\ 
                            & \daft - MRPC-X & 6.42	& -4.03 &	-13.61 &	-12.40 &	-11.42 &	-10.46 &	-8.94 &	-4.99\\ 
                            & \daft - QQP-X & 14.44	& 6.10	& -2.72	& -3.85	 & -3.62	& -4.09	 & -3.26 &	-2.00\\ 
                            \cmidrule{2-10}
                            & Average & 18.46	& 9.18	& -0.69	& -3.30	 & -4.10	& -5.72	& -4.95 &	-5.33
                            \\
    \bottomrule
  \end{tabular}
\end{table*}

\vfill
% \section{Appendix}
% You may include other additional sections here.

\end{document}